\documentclass[journal=jacsat,manuscript=article]{achemso}

\usepackage[version=3]{mhchem} 

\newcommand{\name}{NAG2G}

\usepackage{multirow}
\usepackage{algorithm}
\usepackage{algorithmic}
\usepackage{graphicx} 
\usepackage{booktabs} 
\usepackage{pifont}
\usepackage{bbding}
\usepackage{fontawesome}
\usepackage{amsmath}
\usepackage{amsfonts}
\usepackage[T1]{fontenc}
\usepackage[sort&compress,numbers,super]{natbib}

\author{Lin Yao}
\affiliation[DP Technology]
{DP Technology, Beijing, China}

\author{Wentao Guo}
\affiliation[University of California, Davis]
{Department of Chemistry, University of California, Davis, California, 95616, USA}
\alsoaffiliation{Department of Statistics, University of California, Davis, California, 95616, USA}
\alsoaffiliation[DP Technology]{DP Technology, Beijing, China}
\alsoaffiliation{equal contributions}

\author{Zhen Wang}
\affiliation[DP Technology]
{DP Technology, Beijing, China}
\alsoaffiliation{equal contributions}

\author{Shang Xiang}
\affiliation[DP Technology]
{DP Technology, Beijing, China}

\author{Wentan Liu}
\affiliation[DP Technology]
{DP Technology, Beijing, China}

\author{Guolin Ke}
\email{kegl@dp.tech}
\affiliation[DP Technology]
{DP Technology, Beijing, China}

\title[An \textsf{achemso} demo]
 {Node-Aligned Graph-to-Graph (NAG2G): Elevating Template-Free Deep Learning Approaches in Single-Step Retrosynthesis}

\abbreviations{IR,NMR,UV}

\begin{document}

\begin{tocentry}

		\centering
		\includegraphics[width=0.9\textwidth]{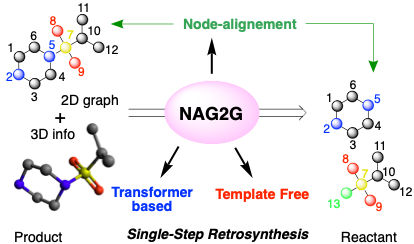}
		\label{fig:TOC}

\end{tocentry}

\begin{abstract}
Single-step retrosynthesis (SSR) in organic chemistry is increasingly benefiting from deep learning (DL) techniques in computer-aided synthesis design. While template-free DL models are flexible and promising for retrosynthesis prediction, they often ignore vital 2D molecular information and struggle with atom alignment for node generation, resulting in lower performance compared to the template-based and semi-template-based methods. To address these issues, we introduce Node-Aligned Graph-to-Graph (NAG2G), a transformer-based template-free DL model. NAG2G combines 2D molecular graphs and 3D conformations to retain comprehensive molecular details and incorporates product-reactant atom mapping through node alignment which determines the order of the node-by-node graph outputs process in an auto-regressive manner. Through rigorous benchmarking and detailed case studies, we have demonstrated that NAG2G stands out with its remarkable predictive accuracy on the expansive datasets of USPTO-50k and USPTO-FULL. Moreover, the model's practical utility is underscored by its successful prediction of synthesis pathways for multiple drug candidate molecules. This not only proves NAG2G's robustness but also its potential to revolutionize the prediction of complex chemical synthesis processes for future synthetic route design tasks. 
\end{abstract}

\section{Introduction}

The single-step retrosynthesis (SSR)~\cite{Corey1995-ud} is an essential operation in organic chemistry, involving the reversed synthesis of a target product or intermediate in a single step. To achieve automatic multi-step synthesis route design, SSR plays a critical role in building the blocks for each separated stage. Typically, the design of retrosynthesis strategies demands a thorough understanding and knowledge of organic chemistry principles, such as reaction mechanisms and reactive sites. With the emergence of computer-aided synthetic planning tools, researchers are now harnessing deep learning (DL) techniques to address this task, recognizing their immense potential.

Various DL architectures have been developed and refined to suit the purpose of learning reactions for SSR tasks.~\cite{zhongrecent} Even though there are notable variations in their network structures and data representation formats, they mainly fall into two primary groups: template-dependent and template-independent. In the following section, we will provide a concise overview of recent DL-based methods, highlighting their model designs, as well as their strengths and weaknesses. 

\begin{figure}
		\centering		\includegraphics[width=1.0\textwidth]{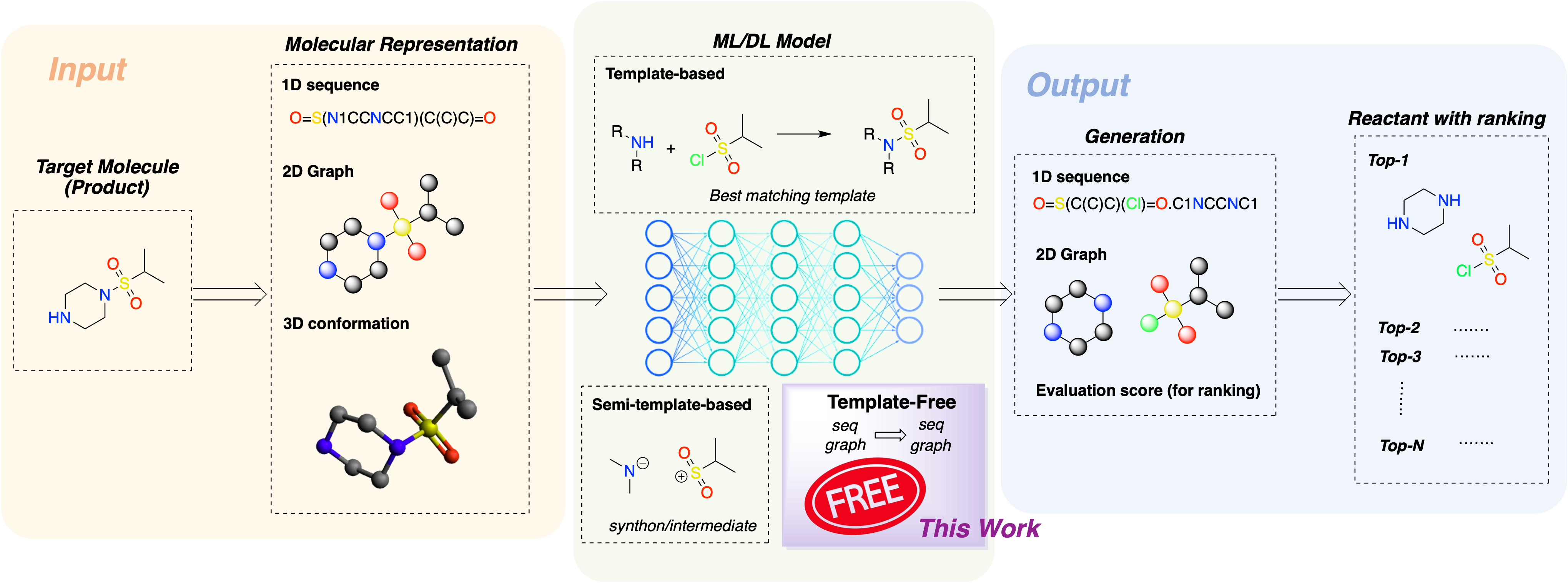}
		\caption{Overview of computer-aided SSR workflow based on template-based, semi-template-based and template-free design. }
		\label{fig:intro}
\end{figure}

\textbf{Template or non-template?} The chemical intuition of organic synthetic chemists is accumulated from the knowledge of reaction rules. Naturally, a dictionary or so-called templates of existing reactions (such as an organic synthesis textbook) will serve as a bible for SSR design. Therefore, the initial generation of retrosynthesis tools was trained to search for the most likely reaction templates in the library that could be used for generating the desired product. For instance, the program \textit{Synthia} (previously named as \textit{Chematica})~\cite{2018chematica} employs over 80,000 rules crafted by synthetic experts to determine the appropriate reaction step based on a huge decision tree. Deep learning strategies, like \textit{RetroSim}~\cite{coley2017computer} and \textit{NeuralSym}~\cite{segler2017neural}, use traditional molecular similarity metrics, such as fingerprints and Tanimoto similarity, to look for the templates that match well with products. Other contemporary approaches include \textit{LocalRetro}~\cite{localretro}, \textit{GLN}~\cite{gln} and \textit{RetroComposer}~\cite{Yan_2022}. The limitations of template-based methods are inherent to the library they rely on. The library may not cover all potential reactions, and there is a risk of incorrect associations between intricate products and template structures. To avoid being overly dependent on dictionaries, semi-template approaches emerged. This method breaks down the SSR prediction process into two stages -- synthons~\cite{koca2012synthon} or intermediate detection followed by reactant generation. Both steps are critical as the preprocessing and identification of synthon are directly associated with the reactant prediction. The advantage of the two-stage methodology includes synthons understanding, searching capabilities expansion, and reaction scheme exploration. Meantime, errors can be easily passed from the first step to the second. As the technique of semi-template approaches grows, several models include \textit{G2G}~\cite{graph2graph}, \textit{RetroXpert}~\cite{retroxpert}, \textit{RetroPrime}~\cite{retroprime}, \textit{GraphRetro}~\cite{graphretro} \textit{SemiRetro}~\cite{gao2022semiretro}, \textit{G2Retro}~\cite{chen2023g2retro}, \textit{Graph2Edits}~\cite{zhong2023retrosynthesis} have emerged, highlighting the compatibility of graph-to-graph models with molecular topology edits, which we will discuss later.

Can a deep learning model learn chemistry without any prerequisite knowledge (including dictionary, template, synthons, intermediates, and editing strategies) given by scientists? The answer is YES. Template-free models, such as pioneering \textit{seq2seq}~\cite{retro_seq2seq}, \textit{SCROP}~\cite{SCROP}, \textit{Tied Transformer}~\cite{tied_transformer}, \textit{Augmented Transformer}~\cite{retro_at}, and \textit{RetroDCVAE}~\cite{retrodcvae} all consider retrosynthesis as a prediction problem. The foundational idea posits that molecules can be analyzed in a manner akin to natural language processing (NLP) tasks. In this framework, product molecules are broken down into tokens based on their one-dimensional (1D) string representations, like the Simplified Molecular Input Line Entry System (SMILES). This tokenization allows us to treat the transformation of products into reactants, drawing parallels between chemical reactions and language translation processes. Recent studies have achieved marked improvements by applying the advanced NLP \textit{Transformer}~\cite{transformer} model, which employs a multi-head attention mechanism. This mechanism enables the model to assign varying degrees of importance to different segments of input data, enhancing its ability to manage the message-passing process between each pair of atoms within a molecule and between pairs of product and reactant.

\textbf{Choices of molecular representation} To help computers think like chemists, it is crucial to translate reaction information, specifically, molecular-level reactants and products, into \textit{in silico} "language", or so-called molecular representation. One popular approach of DL-based SSR models~\cite{SCROP,tied_transformer,retro_at,retrodcvae,wan2022retroformer} is to employ 1D sequences, such as SMILES. Despite its simplicity, the 1D sequence-based model exhibits several limitations: 1) The sequence disregards the extensive molecular topological information~\cite{graph2smiles,get,GTA}; 2) Legal SMILES follow intricate syntax rules, which magnify the difficulty of valid SMILES generation; 3) Effectively utilizing atom mapping information between products and reactants is challenging for 1D representation. Without alignment, model performance may decline due to lost atom correlations between products and reactants. 4) Due to the fact that a single molecule can have multiple SMILES representations when generating multiple candidate reactants for a product, it is possible to generate multiple reactant SMILES representing the same reaction, which may reduce the diversity of the candidates~\cite{retro_at}.

To overcome the limitations of 1D sequences, models involving 2D molecular graphs, which encompass atom (node) and bond (edge) topology, have been proposed for molecular representation in SSR tasks.~\cite{MEGAN,graph2graph,chen2022mathsfg2retro,liu2022mars,chen2023g2retro,zhong2023retrosynthesis}. 2D graphs encapsulate a wealth of information about the atomic environment, such as neighboring atoms and their connections. The graph topology offers an optimal solution for two-step tasks involving the modification of synthons and intermediates under the context of semi-template models. In order to effectively utilize the natural mapping information between atoms in products and reactants, prior approaches have employed a repeated graph edit strategy~\cite{MEGAN}, wherein the input graph of the product is iteratively modified by taking edit actions (such as adding nodes, removing nodes, updating nodes or edges) until it reaches an end -- the reactant. For example, semi-template models such as \textit{G2Retro}~\cite{chen2022mathsfg2retro}, \textit{MARS}~\cite{liu2022mars}, \textit{Graph2Edits}~\cite{zhong2023retrosynthesis} employ graph edit strategies~\cite{MEGAN} for graph generation. However, the 2D graph edition requires a delicate arrangement of edit actions and editing types forehead. 
Despite the advanced generation processes reducing computational expenses, the iterative action-prediction cycles, which require graph input and output at each editing step, continue to add to the computational burden.

\textbf{Boundaries and pushing boundaries.} Models have improved by combining different methods with the latest machine learning designs, as we've discussed before. Even though there are still some challenges, we see great potential in the new Transformer-based models that can make predictions without using templates.~\cite{retro_seq2seq,lin2020automatic} Despite template-dependent models taking the upper hand, we've observed that template-free methods not only demonstrate remarkable performance with neat and concise structures but are also capable of capturing the nuances of chemical reasoning themselves. To leverage more information to the model, molecular representation, and attention mechanism (consideration of long-range dependencies between atoms) adaption are required. For example, \textit{GET}~\cite{get} merges both graphs and SMILES encoders. \textit{GTA}~\cite{GTA} integrates topological data into attention bias. Specifically, \textit{Retroformer}~\cite{wan2022retroformer}, using 1D sequence for the encoder, incorporates product-reactant atom alignments for better results.~\cite{zhong2022root} Graph-based template-free methods, including G2GT~\cite{g2gt}, have advanced in exploiting graph topology. Nevertheless, they have yet to leverage node-alignment strategies for enhanced performance.

To leverage the benefits of a template-free approach and address the above limitations, we developed \name{} that utilizes both 2D graph and 3D coordinates, with improved efficiency of graph generation and node alignment according to proper atom mappings as shown in figure ~\ref{fig:intro}. Moreover, we implement an auto-regressive approach that generates graphs node-by-node according to the aligned order, drawing inspiration from language generation techniques. \name{} is trained using two widely recognized datasets, USPTO-50k~\cite{uspto50k} and USPTO-Full~\cite{gln,retro_at} with augmented data, showing great capacity compared to existing models. Additionally, our model demonstrates its proficiency in tackling real-world problems by iteratively generating step-by-step synthesis pathways for drug candidates. To gain deeper insights into the significance of each component within our methodology, we conducted ablation studies, systematically omitting certain parts of the model to evaluate their impact on performance.


\section{Method}

\subsection{Model construction}

\begin{figure}
		\centering		\includegraphics[width=0.7\textwidth]{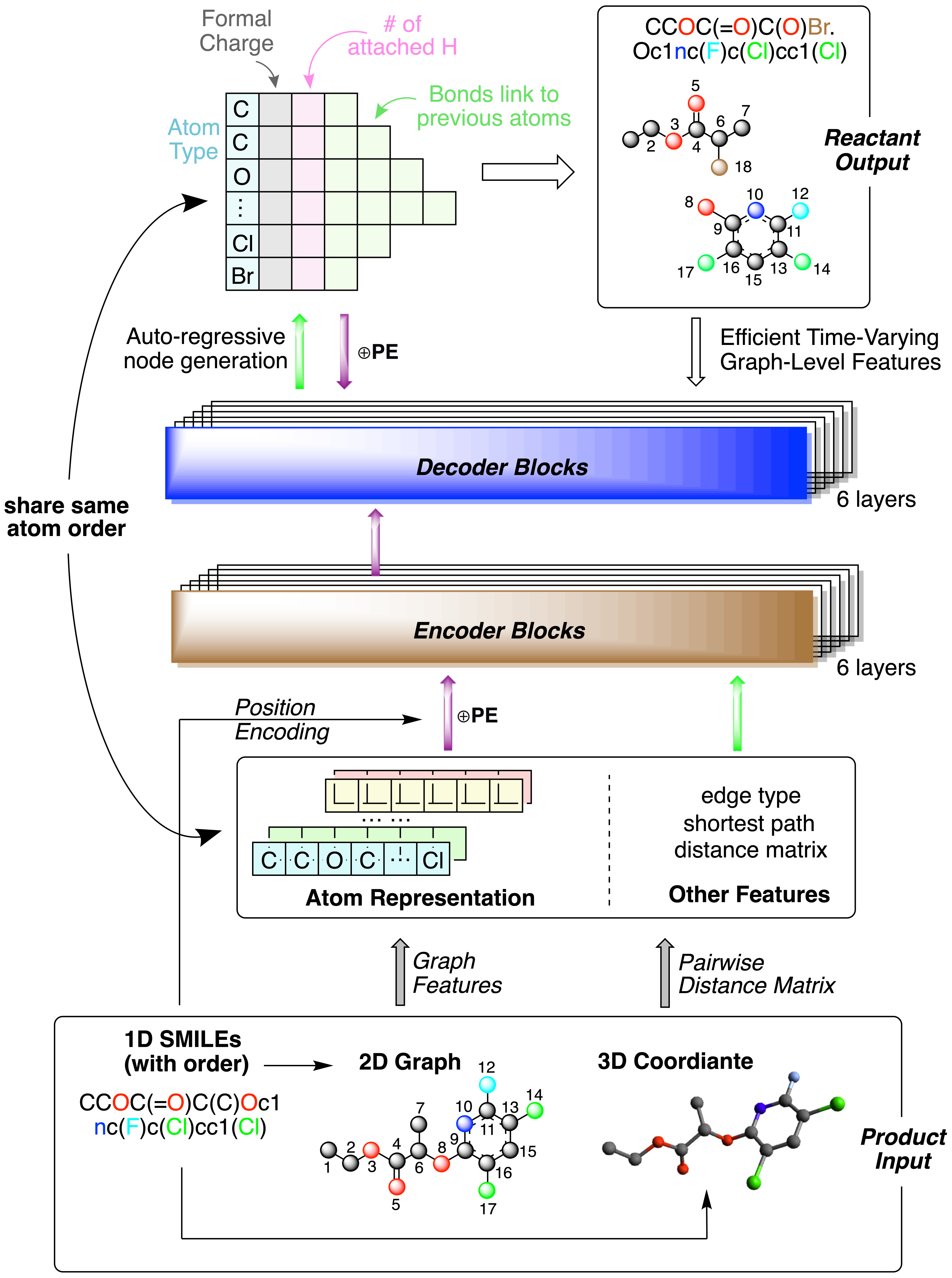}
		\caption{The network architecture of \name{}.}
		\label{fig:model}
\end{figure}

\textbf{Encoders}, the components of a neural network that processes and compresses input product into a compact representation, play the critical role of learning molecular representation in the \name{} Transformer-based encoder-decoder architecture. Competent models, such as \textit{Graphormer}~\cite{graphormer} and \textit{Uni-Mol}~\cite{zhou2023unimol} have demonstrated the efficiency of encoder representation learning strategies. Thus, we adopt the encoder from \textit{Uni-Mol+} \cite{lu2023highly} which incorporates both 2D graph and 3D conformation for molecular representation as shown in 
Figure~\ref{fig:model}. The 1D positional encoding is also taken into account, serving as the node order encoder. Formally, we can denote the process of the encoder as the following equation (1):

\begin{equation}
    \mathbf{O}^{\text{enc}} = f_{\text{enc}}(\mathbf{X}, \mathbf{P}^{\text{enc}}, \mathbf{E}, \mathbf{R}; \mathbf{\theta}^{\text{enc}}),
\end{equation}

In the proposed formulation: $\mathbf{X}$ denotes the atom features; $\mathbf{P}^{\text{enc}}$ represents the 1D positional encoding, which is supplementary to the atomic embeddings; $\mathbf{E}$ signifies the edge features inherent to the 2D graph structure; $\mathbf{R}$ corresponds to the atomic coordinates in the 3D conformation; $\mathbf{\theta}^{\text{enc}}$ encapsulates the encoder's learnable parameters, and $\mathbf{O}^{\text{enc}}$ is the derived molecular representation result from the encoder. 

\textbf{Decoders} primarily operates to generate the reactant graph node-by-node through an auto-regressive approach. At the $i$-th time step, which also corresponds to the $i$-th generated node (atom), the decoder receives three distinct inputs: 

1) The encoder's output, including keys and values that help in the interaction between the encoder and decoder.

2) The decoder outputs from prior steps (from $1$ to $i-1$), which is typical of auto-regressive models in that the prediction of a new value is based on its preceding values. During the iterative process, 1D positional encoding is added, which is essential for \name{} to align atom order between encoder (product) inputs and decoder (reactant) outputs. 

3) The graph-level features of the current output graph, such as node degrees and shortest paths between nodes. Incorporating these graph-level features directly into the model presents an efficiency challenge, as the graph features vary across time steps. To address this issue, we propose an efficient method for integrating these graph-level features. 

Given the above inputs, the decoder generates a new node at the $i$-th time step auto-regressively, starting from atomic type, and then associated formal charge, the number of connected hydrogen atoms, and finally its edges (types of bond) linked to prior nodes (atoms). The information for each node is produced sequentially given its above predictions. For instance, the formal charge is predicted based on the prior atomic type prediction. The process is denoted as:

\begin{equation}
\begin{aligned}
    t_i &= f_{\text{dec}}(\mathbf{P}^{\text{dec}}_{1:i}, \mathbf{N}_{1:i-1}, \mathbf{G}_{1:i-1}, \mathbf{O}^{\text{enc}}; \mathbf{\theta}^{\text{dec}}), \\
    c_i &= f_{\text{dec}}(t_i, \mathbf{P}^{\text{dec}}_{1:i}, \mathbf{N}_{1:i-1}, \mathbf{G}_{1:i-1}, \mathbf{O}^{\text{enc}}; \mathbf{\theta}^{\text{dec}}), \\ 
    h_i &= f_{\text{dec}}(c_i, t_i, \mathbf{P}^{\text{dec}}_{1:i}, \mathbf{N}_{1:i-1}, \mathbf{G}_{1:i-1}, \mathbf{O}^{\text{enc}}; \mathbf{\theta}^{\text{dec}}), \\ 
    e_{i,1} &= f_{\text{dec}}(h_i, c_i, t_i, \mathbf{P}^{\text{dec}}_{1:i}, \mathbf{N}_{1:i-1}, \mathbf{G}_{1:i-1}, \mathbf{O}^{\text{enc}}; \mathbf{\theta}^{\text{dec}}), \\ 
    & ... \\
    e_{i,k} &= f_{\text{dec}}(e_{i,k-1}, ..., e_{i,1}, h_i, c_i, t_i, \mathbf{P}^{\text{dec}}_{1:i}, \mathbf{N}_{1:i-1}, \mathbf{G}_{1:i-1}, \mathbf{O}^{\text{enc}}; \mathbf{\theta}^{\text{dec}}), \\ 
\end{aligned}
\end{equation}

where $\mathbf{N}_{1:i-1}$ represents the set of nodes generated from the previous $i-1$ time steps, $\mathbf{P}^{\text{dec}}_{1:i}$ denotes the 1D positional encoding of the current $i$ nodes, $\mathbf{G}_{1:i-1}$ represents the graph feature extracted from previous outputs, and $\mathbf{\theta}^{\text{dec}}$ denotes the parameters of the decoder. The atomic type, associated formal charge, and the number of connected hydrogen atoms for the $i$-th node are represented by $t_i$, $c_i$, and $h_i$, respectively. The $d$-th edge, denoted by $e_{i,d}=(j, b)$, connects the $i$-th node and the $j$-th node with the bond type $b$. To define an edge's generative order, edges linked to nodes with larger 1D positions are prioritized. Generation of $c_i$ and $h_i$ is skipped if a node has zero charges or no linked hydrogen atoms to reduce generative steps. The overview of \name{}'s architecture is depicted in Figure~\ref{fig:model}.

\subsubsection{Node alignment and data augmentation}

Molecular graphs, unlike sentences, lack an inherent sequence, as atoms within a molecule do not have a natural order until assigned. In order to circumvent the need to consider the order of node in graph generation, several methods transform the graph generation task into indirect approaches, such as graph edit action prediction or SMILES prediction tasks, as discussed earlier. Alternatively, some methods utilize a one-in-all scheme, generating the entire graph output in a single step. Although this scheme avoids considering generation order, it lacks flexibility and is unsuitable for multi-solution tasks such as retrosynthesis. To adopt a more flexible, auto-regressive method, the output node order must be determined. A simple solution is using the canonical SMILES atom order; however, this fixed order restricts output data augmentation and limits the model's performance. Consequently, devising a robust and flexible strategy to tackle the unordered nature of graph nodes remains a formidable challenge in graph generation tasks. The unordered nodes not only challenge the graph generation but also the encoder input data augmentation. As input graphs inherently lack sequence, graph data augmentation must rely on alternative strategies such as omitting certain node or edge information. However, this approach may be unsuitable for retrosynthesis, as the omission of critical information, such as different reaction sites, could result in vastly different outputs. Utilizing a reactant from the training set under these circumstances may introduce inaccuracies. Therefore, a more appropriate encoder input data augmentation strategy might be based on node order, ensuring a more accurate and reliable outcome.

\begin{figure}
		\centering
            \includegraphics[width=0.8\textwidth]{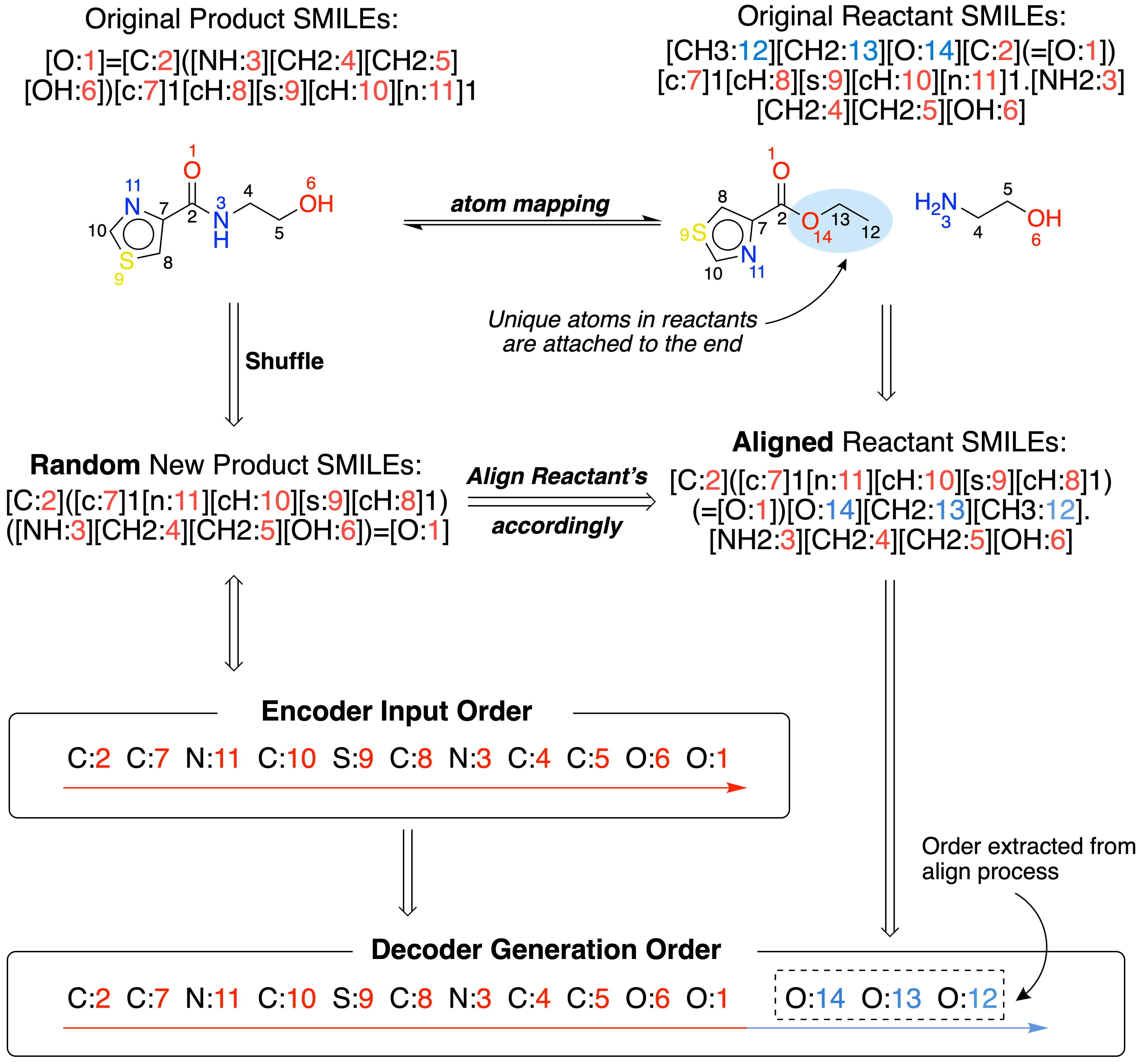}
		\caption{An example to illustrate the process of data augmentation and product-reactant alignment. The red numbers indicate the atoms present in both the product and reactants, while the blue ones represent the atoms found only in the reactants.}
		\label{fig:align_new}
\end{figure}

\begin{figure}
		\centering
		\includegraphics[width=1.0\textwidth]{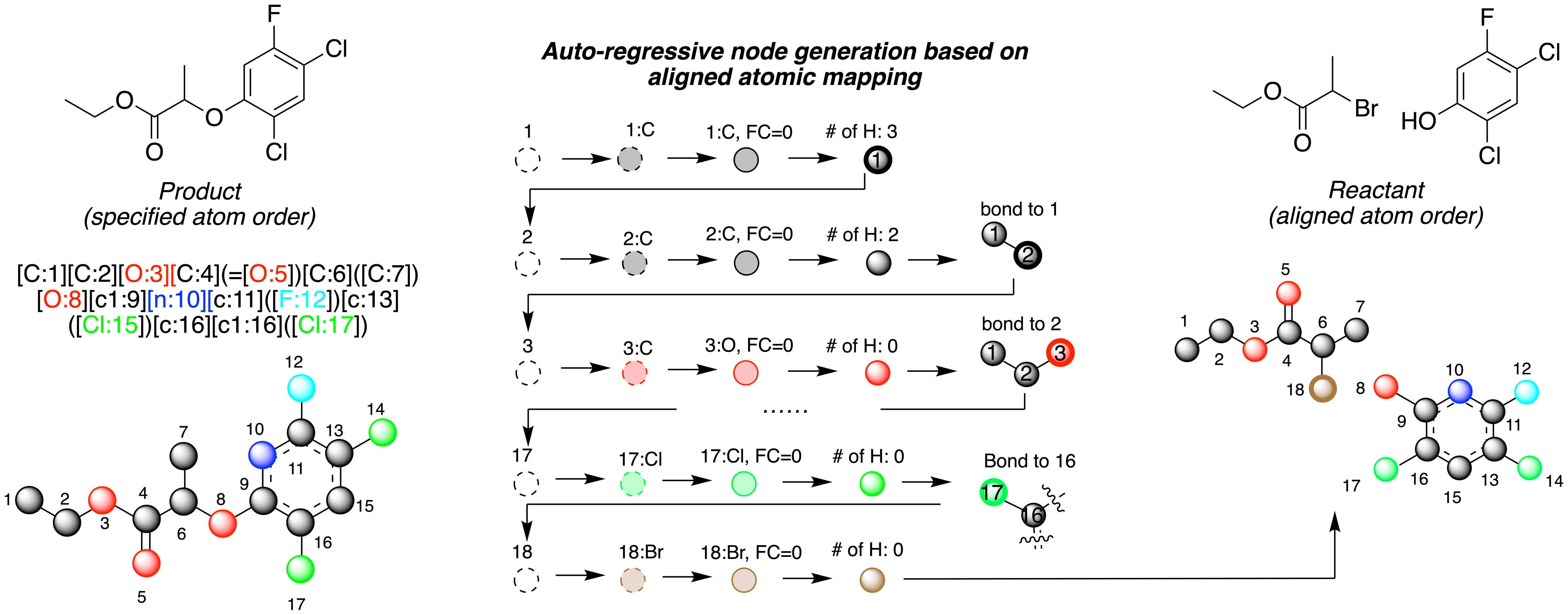}
		\caption{The illustration of the node-aligned graph-to-graph generation. }
		\label{fig:graph_gen}
\end{figure}

To address the challenges in input and output data augmentation and enable a flexible node-by-node auto-regressive generation, we propose a novel method based on product-reactant node alignment. Our method begins with the random generation of the product's SMILES sequence by RDKit\cite{rdkit}, as shown in Figure~\ref{fig:align_new}. By following the new order in the SMILES sequence, we obtain the data-augmented input graph's node sequence order. Subsequently, the graph node order is marked using position embedding. For the product graph with a determined order, we establish a unique and unambiguous rule that corresponds to the reactant node order for node-by-node output, as demonstrated in Figure~\ref{fig:align_new} and Figure~\ref{fig:graph_gen}. In the reactant, atoms can be classified into two types: those shared with the product and those exclusive to the reactant. The assignment of atomic order should consider both of these aspects. Firstly, in generating the order, we stipulate that the shared atoms' order in the reactant should precede the order of non-shared atoms. To ensure that for a specific ordered product input, there exists a unique corresponding ordered reactant, the order of shared atoms in the reactant should follow the order in the product. Subsequently, the reactant SMILES is aligned with the product SMILES using RDKit to obtain the most similar SMILES. Finally, the order of non-shared atoms is extracted from the aligned reactant SMILES, ensuring the uniqueness of the non-shared atoms' order. 
This approach utilizes the product-reactant alignment information by ensuring the node generation order mirrors the input graph's order in training process, and allows for consistent and equivariant data augmentation in both input and output, enhancing the overall robustness and accuracy of the generation process. By utilizing \name{}, we provide a robust and adaptable auto-regressive generation procedure that can effectively handle the complexities of molecular graphs and enhance the performance of graph-to-graph generation models, and offers a concise, profound, and persuasive solution for input-output data augmentation, ensuring logical and efficient node-by-node auto-regressive generation.

\subsection{Efficient time-varying graph-level features} 

\begin{figure}
		\centering
            \includegraphics[width=0.5\textwidth]{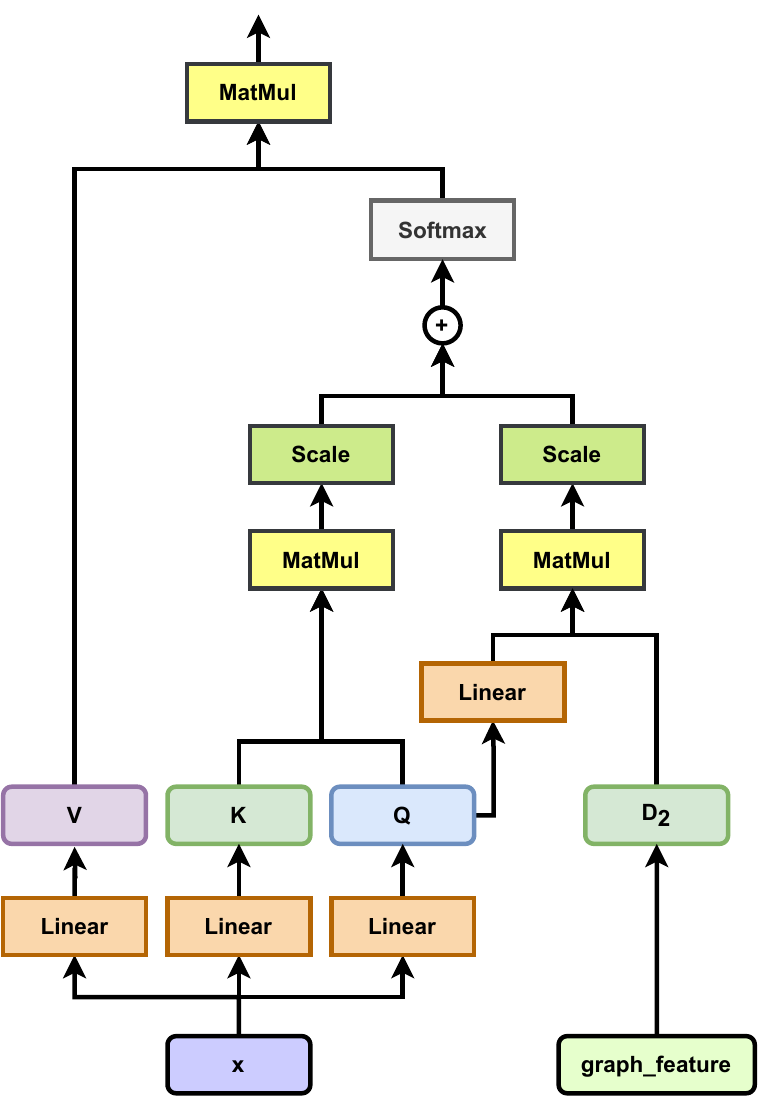}
		\caption{The illustration of decoder attention mechanism.}
		\label{fig:attn}
\end{figure}

\label{sec:graph_feature}
During the generation process through decoders, the implementation of teacher forcing during training allows for the true output from a previous time step to be used as input for the current step, rather than the model's own prediction. This technique not only aligns the model's learning with the correct sequence of outputs but also enables parallel processing of data at various time steps. The interaction between the current and previous time steps is addressed within the decoder's attention layer. To avoid the inadvertent incorporation of future information, the attention matrix in a transformer model is masked with an upper triangular matrix. This ensures that a given output at a specific time step can only be influenced by preceding elements in the sequence, preserving the autoregressive property where the prediction for each step is conditioned only on the known past information. Formally, we denote this process as:
\begin{equation}
\text{Attention}(\mathbf{Q}, \mathbf{K}, \mathbf{V}) = \text{softmax}\left(\frac{\mathbf{Q}\mathbf{K}^T}{\sqrt{d_h}} + \mathbf{M}\right)\mathbf{V},
\end{equation}

where $\mathbf{Q}, \mathbf{K}, \mathbf{V} \in \mathbf{R}^{n \times d_h}$ represent the query, key, and value matrices, respectively. $d_h$ is the dimension of one head. And $n$ is the number of time steps. $\mathbf{M}$ is an additive mask matrix that ensures only the relevant information from the current and previous time steps is considered during the attention computation. For the sake of simplicity, we present the calculation for only one head. The multi-head attention process executes the above single-head calculation in parallel. 
During the calculation of one head, the computational complexity is $\mathcal{O}(n \times n \times d_h)$, and the peak memory consumption is $\mathcal{O}(n \times n)$. 

As previously mentioned, graph-level features vary across time steps, and their direct utilization poses an efficiency challenge during model training. Specifically, to maintain the time-varying graph features, a matrix with shape $n \times n \times d_h$ is required\footnote{Here, we consider node-wise graph features, such as node degrees. Pair-wise graph features, such as the shortest path, will consume significantly more memory.}. These time-varying graph features are then employed as additive positional encodings. As a result, the attention layer can be represented as:
\begin{equation}
\text{Attention}(\mathbf{Q}, \mathbf{K}, \mathbf{V}, \mathbf{D}) = \text{softmax}\left(\frac{\mathbf{Q}(\mathbf{K} + \mathbf{D})^T}{\sqrt{d_h}} + \mathbf{M}\right)(\mathbf{V} + \mathbf{D}),\\
\end{equation}
where $\mathbf{D} \in \mathbb{R}^{n \times n \times d_h}$ denotes the time-varying graph features, and the shape of $\mathbf{Q}, \mathbf{K}, \mathbf{V}$ is reshaped to $n \times 1 \times d_h$ for broadcasting. In this process, although the computational complexity remains unchanged, the peak memory consumption increases to $\mathcal{O}(n \times n \times d_h)$. Considering that $d_h$ is typically $32$ or even larger, this significant increase in peak memory consumption is considered impractical for real-world applications.

To reduce the cost, we first remove $\mathbf{D}$ from $\mathbf{V}$+$\mathbf{D}$ term. Then, the cost is bottlenecked at $\mathbf{Q}\mathbf{D}^T$ for $\mathbf{Q}(\mathbf{K} + \mathbf{D})^T = \mathbf{Q}\mathbf{K}^T + \mathbf{Q}\mathbf{D}^T$ term. Thus, we can reduce the size of the last dimension by substituting $\mathbf{D}$. Thus, the attention can be transformed to:
\begin{equation}
\text{Attention}(\mathbf{Q}, \mathbf{K}, \mathbf{V}, \mathbf{D_2}) = \text{softmax}\left(\frac{\mathbf{Q}\mathbf{K}^T}{\sqrt{d_h}} + \frac{\mathbf{Q}\mathbf{U}\mathbf{D_2}^T}{\sqrt{d_{h2}}} + \mathbf{M}\right)\mathbf{V},\\
\end{equation}
where $\mathbf{U} \in \mathbb{R}^{d_h \times d_{h2}}$ is employed to reduce the dimension of $\mathbf{Q}$, and $\mathbf{D}_2 \in \mathbb{R}^{n \times n \times d_{h2}}$ represents the time-varying graph features with a much smaller dimension $d_{h2}$. With this configuration, the peak memory is reduced to $\mathcal{O}(n \times n \times d_{h2})$. Figure~\ref{fig:attn} illustrates the design of a self-attention layer for time-varying graph features.

\subsection{Data preparation}

\begin{figure}
		\centering		\includegraphics[width=0.8\textwidth]{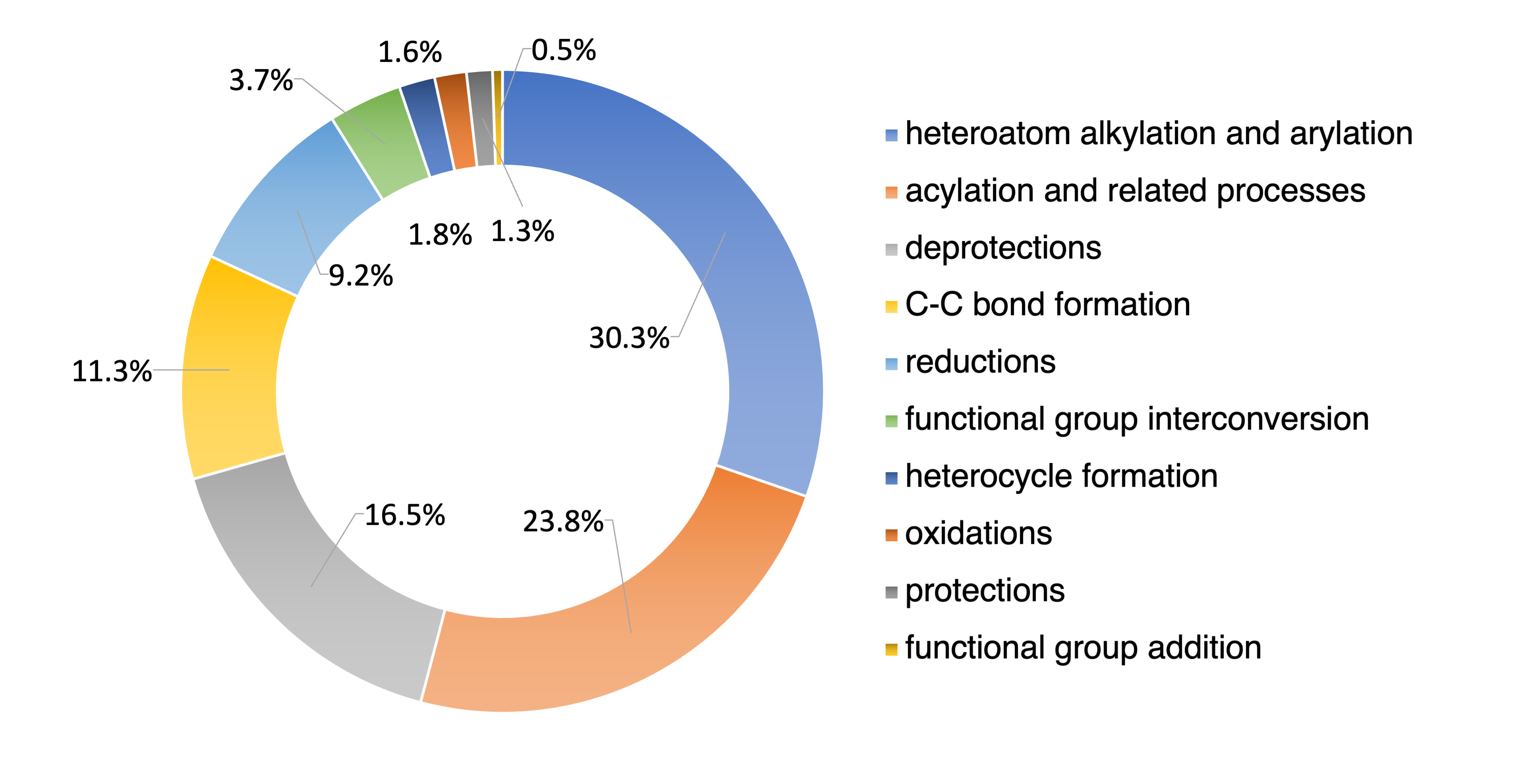}
		\caption{Distribution of 10 types pf reactions in USPTO-50k dataset. The reaction's legends are ranked based on their fraction from largest to smallest.}
		\label{50K}
\end{figure}

The \name{} is trained and tested on two broadly acknowledged datasets, USPTO-50k~\cite{uspto50k} and USPTO-Full~\cite{gln,retro_at}. USPTO-50k comprises 50,016 atom-mapped reactions, categorized into 10 reaction classes. The USPTO-50K dataset was split into 40,008, 5,001, and 5,007 reactions for the training, validation, and test sets, respectively. We also used the filtered USPTO-Full dataset with approximately 1 million atom-mapped reactions as described by Tetko et al.~\cite{retro_at} instead of the original USPTO-Full dataset.~\cite{gln}. After filtering out incorrect reactions, which leads to an approximate 4\% size reduction, training, validation, and test sets contain approximately 769,000, 96,000, and 96,000 reactions. The distribution of reaction classes in the training, valid, and test set of USPTO-50k are the same, displayed in Figure~\ref{50K}. Consistent with previous works, we did not benchmark the USPTO-Full results with the aid of reaction class information.

\section{Results}

\textbf{\name{} Setup} The setup of \name{} contains a 6-layer encoder and a 6-layer decoder. The input embedding dimension was set to 768, and the number of attention heads was set to $24$. We employed the Adam optimizer~\cite{adam} with $(\beta_1, \beta_2)$ = $(0.9, 0.999)$, and linear warmup and decay with a peak learning rate of 2.5e-4. The training process took a total of 12,000 steps with a batch size of 16, requiring 6 hours to finish on a single NVIDIA A100 GPU. For the training on the USPTO-Full dataset, \name{} ran 48,000 training steps with a batch size of 64, taking approximately 30 hours to complete on eight NVIDIA A100 GPUs.

\subsection{Results on USPTO dataset} 
To evaluate the performance of \name{} during inference, we utilize the commonly adopted beam search method for top candidate predictions. We configure the beam size at 10, using a length penalty of 0 and a temperature of 1. Notably, data augmentation is not applied during the inference phase. Additionally, \name{} relies on RDChiral~\cite{rdchiral} to assign the atomic chirality of reactants, drawing from the product's stereochemistry. 

The definition of prediction accuracy follows the approach proposed by Liu et al.~\cite{retro_seq2seq}, which considers a prediction to be accurate only if it completely identifies all reactants for a specific chemical reaction. We measure the top-$k$ accuracy of predictions, defined as the proportion of test cases in which the correct answer appears among the top $k$ candidates of the beam search results.

 \textbf{USPTO-50k}. In table~\ref{tbl:result_50k}, \name{} demonstrates superior performance on the USPTO-50k dataset compared to recent baseline approaches, including template-based, semi-template-based, and template-free models.~\ref{tbl:result_50k}. No supplementary techniques were implemented to aid the inference procedure in NAG2G to ensure a fair comparison. Within the template-free domain, \name{} markedly surpasses all benchmarks across every metric. Even though certain baselines leverage extra data or methods to bolster their results (indicated by $\ast$), \name{} continues to excel significantly without any such enhancements. Despite the additional use of pre-defined rules in template-based and semi-template-based methods, \name{} outperforms them without prerequisite information. This marks the debut of \name{} surpassing both template-based and semi-template-based approaches, that earlier template-free benchmarks never achieved. Besides, \name{} gives SOTA results based on MaxFrag metric proposed by \textit{Augmented Transformer}~\cite{retro_at}, see SI for details. Detailed results include testing of each reaction class on the USPTO-50k dataset when trained with the class known is summarized in SI as well.

\begin{table}[H]
 \centering
 \caption{Top-$k$ accuracy for retrosynthesis prediction on USPTO-50k.
 The best performance is in \textbf{bold}, and the best results in each method type are \underline{underlined}. Models denoted by an asterisk ($\ast$) employed supplementary datasets for training or incorporated techniques to enhance the accuracy during inference. In order to maintain a fair comparison, we also present their results without the implementation of these additional techniques. 
}
\vspace{-0.05in}
 \begin{tabular}{lcccccccccccccc}
 \toprule
  & \multicolumn{14}{c}{\textbf{Top-$k$ Accuracy (\%)}} \\
  \cline{2-15} 
 \multicolumn{1}{c}{} & \multicolumn{9}{c}{\textbf{USPTO-50k}} & \textbf{}\\
  \cline{2-10} \cline{12-15} 

 \multicolumn{1}{c}{} & \multicolumn{4}{c}{\textbf{Reaction Class Known}} & \textbf{} & \multicolumn{4}{c}{\textbf{Reaction Class Unknown}} & \textbf{} \\
 \cline{2-5} \cline{7-10} \cline{12-15} 
 \textbf{Model} & 1 & 3 & 5 & 10 & & 1 & 3 & 5 & 10 \\
 \midrule
 \textbf{Template-Based} & & & & & & & & & \\
 \midrule
 RetroSim \cite{coley2017computer}& 52.9 & 73.8 & 81.2 & 88.1 & & 37.3 & 54.7 & 63.3 & 74.1 \\
 NeuralSym \cite{segler2017neural}& 55.3 & 76.0 & 81.4 & 85.1 & & 44.4 & 65.3 & 72.4 & 78.9 \\
 GLN \cite{gln} & 64.2 & 79.1 & 85.2 & 90.0 & & 52.5 & 69.0 & 75.6 & 83.7  \\
 MHNreact \cite{mhnreact} & - & - & - & - & &50.5 & 73.9 & 81.0 & \underline{87.9} \\

 RetroComposer \cite{Yan_2022} & \underline{65.9} & \underline{85.8} & \underline{89.5} & \underline{91.5} & & \underline{54.5} & \textbf{\underline{77.2}} & \underline{83.2} & 87.7 \\
  \midrule
 \textbf{Semi-Template-Based} & & & & & & & & & \\
 \midrule
 G2G \cite{graph2graph} & 61.0 & 81.3 & 86.0 & 88.7 & & 48.9 & 67.6 & 72.5 & 75.5 \\
 RetroXpert \cite{retroxpert} & 62.1 & 75.8 & 78.5 & 80.9 & & 50.4 & 61.1 & 62.3 & 63.4 \\
 RetroPrime \cite{retroprime} & 64.8 & 81.6 & 85.0 & 86.9 & & 51.4 & 70.8 & 74.0 & 76.1 \\ 
 GraphRetro \cite{graphretro} & 63.9 & 81.5 & 85.2 & 88.1 & & 53.7 & 68.3 & 72.2 & 75.5 \\
 SemiRetro \cite{gao2022semiretro} & 65.8 & 85.7 & 89.8 & 92.8 & & \underline{54.9} & 75.3 & 80.4 & 84.1 \\
 G2Retro \cite{chen2022mathsfg2retro} & 63.6 & 83.6 & 88.4 & 91.5 && 54.1 & 74.1 & 81.2 & 86.7 \\
 MARS \cite{liu2022mars} & \underline{66.2} & \underline{85.8} & \underline{90.2} & \underline{92.9} & & 54.6 & \underline{76.4} & \underline{83.3} & \underline{88.5} \\
 \midrule
 \textbf{Template-Free} & & & & & & & & & \\
 \midrule

 LV-Transformer \cite{lvtransformer} & - & - & - & - && 40.5 & 65.1 & 72.8 & 79.4 \\

 SCROP \cite{SCROP} & 59.0 & 74.8 & 78.1 & 81.1 & & 43.7 & 60.0 & 65.2 & 68.7 \\
 GET \cite{get} & 57.4 & 71.3 & 74.8 & 77.4 && 44.9 & 58.8 & 62.4 & 65.9 \\
 Tied Transformer \cite{tied_transformer} & - & - & - & - & & 47.1 & 67.1 & 73.1 & 76.3 \\
 MEGAN \cite{MEGAN} & 60.7 & 82.0 & 87.5 & 91.6 & & 48.1 & 70.7 & 78.4 & 86.1 \\
 Aug. Transformer \cite{retro_at} & - & - & - & - & & 48.3 & - & 73.4 & 77.4 \\
 Aug. Transformer $\ast$ \cite{retro_at} & - & - & - & - & & 53.5 & 69.4 & 81 & 85.7 \\
 GTA \cite{GTA} & - & - & - & - & & 51.1 & 67.6 & 74.8 & 81.6 \\
 Graph2SMILES \cite{graph2smiles} & - & - & - & - & & {52.9} & 66.5 & 70.0 & 72.9 \\
 RetroDCVAE \cite{retrodcvae}  & - & - & - & - & & 53.1 & 68.1 & 71.6 & 74.3 \\
 DualTF \cite{sun2021energybased} & 65.7 & 81.9 & 84.7 & 85.9 & & 53.6 & 70.7 & 74.6 & 77.0 \\
 Retroformer \cite{wan2022retroformer} & 64.0 & 82.5 & 86.7 & 90.2 & & 53.2 & 71.1 & 76.6 & 82.1 \\
 G2GT \cite{g2gt} &-&-&-&-&& 48.0 & 57.0 & 64.0 & 64.5 \\
 G2GT $\ast$ \cite{g2gt} &-&-&-&-&& 54.1 & 69.9 & 74.5 & 77.7 \\

  \name{} (ours) &\textbf{\underline{67.2}} & \textbf{\underline{86.4}} & \textbf{\underline{90.5}} & \textbf{\underline{93.8}} && \textbf{\underline{55.1}} & \underline{76.9} & \textbf{\underline{83.4}} & \textbf{\underline{89.9}} \\
 \bottomrule
 \end{tabular}
 \label{tbl:result_50k}
 \end{table}

\textbf{USPTO-FULL}. Table~\ref{tbl:result_full} presents the performance metrics of various models evaluated on the USPTO-Full dataset. As the dataset size increases, the performance of all models declines due to the heightened complexity of the task. Notably, while template-based methods have shown impressive results on the USPTO-50k dataset, their performance falters considerably on the larger USPTO-Full dataset. This trend indicates that the reliance on template-based methods from pre-established rules becomes a limitation when confronting larger and more complex datasets. In contrast, though weakened as well, template-free methods demonstrate a more versatile and adaptive capability, particularly more suited for expansive datasets. Still, it is evident that \name{} consistently outperforms preceding baselines across all evaluative criteria.

\begin{table}[H]
 \centering
 \caption{Top-$k$ accuracy for retrosynthesis prediction on the USPTO-Full dataset. Models denoted by an asterisk ($\ast$) used supplementary datasets for training or incorporated techniques to improve accuracy during inference. For models denoted by a circle ($\circ$), the invalid reactions are excluded from the test set, following the setting of Augmented Transformer~\cite{retro_at}. To align our methods with the previous baselines, we adopted the approach from Augmented Transformer\cite{retro_at}, assuming that the methods failed on the removed test data, as evidenced by the results of our methods without a circle ($\circ$).
}
 \vspace{-0.05in}
 \begin{tabular}{lcccccc}
 \toprule
 \multicolumn{2}{c}{\textbf{Model}} & & \multicolumn{4}{c}{\textbf{Top-$k$ Accuracy (\%)}} \\
\cline{1-2} \cline{4-7} 
Model Type & Methods && 1 & 3 & 5 & 10 \\
 \midrule
 Template-Based  & RetroSim \cite{coley2017computer} & & 32.8 & - & - & 56.1 \\
                 & NeuralSym \cite{segler2017neural} & & 35.8 & - & - & 60.8 \\
                 & GLN \cite{gln} && 39.3 & - & -& 63.7 \\

 \midrule
 Semi-Template-Based & RetroPrime \cite{retroprime} && 44.1 & 59.1 & 62.8 & 68.5 \\ 
 \midrule
 Template-Free & MEGAN \cite{MEGAN} && 33.6 &-&-& 63.9 \\
 & \name{} (ours) && \textbf{47.7} & \textbf{62.0} & \textbf{66.6} & \textbf{71.0} \\
& Aug. Transformer ${\ast\circ}$ \cite{retro_at} & & 46.2 & - & - & 73.3 \\
 & G2GT ${\ast\circ}$ \cite{GTA} & & 49.3 && 68.9 & 72.7 \\
 & \name{} (ours)$\circ$ && \textbf{49.7} & \textbf{64.6} & \textbf{69.3} & \textbf{74.0} \\
 \bottomrule
 \end{tabular}
 \label{tbl:result_full}
 \vspace{-5pt}
\end{table}

\subsection{Result interpretation (ablation study)}

To identify the importance of each component that has been designed for \name{}, we performed ablation studies by studying the impact of its removal. This aids in understanding the \name{} structure and how node alignment, data augmentation, and the incorporation of time-varying graph features benefit the model.

 \begin{table}[H]
 \centering
 \caption{Ablation study, on USPTO-50k with reaction class unknown.}
 \vspace{-0.05in}
 \begin{tabular}{cccccccc}
 \toprule
 \multicolumn{3}{c}{\textbf{Strategies}} && \multicolumn{4}{c}{\textbf{Top-$k$ Accuracy (\%)}} \\
 \cline{1-3} \cline{5-8} 
 Node Alignment & Data Augmentation & Graph Features && 1 & 3 & 5 & 10 \\
 \midrule
 \ding{51} & \ding{51} & \ding{51} && \textbf{55.1} & \textbf{76.9} & \textbf{83.4} & \textbf{89.9} \\
 \ding{51} & \ding{51} & \ding{53}  && 54.1 & 75.9 & 82.6 & 88.8 \\
 \ding{51} & \ding{53}  & \ding{51} && 49.2 & 69.2 & 75.3 & 80.4 \\
 \ding{53}  & \ding{51} & \ding{51} && 46.1 & 47.6 & 48.5 & 49.9 \\
 \ding{53}  & \ding{53}  & \ding{51} && 40.3 & 54.9 & 58.9 & 62.6 \\
 \bottomrule
 \end{tabular}
 \label{tbl:abl}
 \end{table}

Table~\ref{tbl:abl} presents a comprehensive quantitative breakdown of the impact of each strategy on the model's performance on the USPTO-50k dataset, focusing specifically on Top-$k$ accuracy percentages. When all three techniques—node alignment, data augmentation, and graph features—are utilized, the model hits a peak Top-$1$ accuracy of $55.1\%$ and maintains high performance across Top-$3$ ($76.9\%$), Top-$5$ ($83.4\%$), and Top-$10$ ($89.9\%$), emphasizing the synergistic effect of this combination. Eliminating graph features results in a marginal decrease in each Top-$k$ accuracy metric. However, incorporating these graph features contributes to an approximately $1\%$ improvement in the metrics under challenging scenarios where further progress is difficult to achieve.
 Omitting data augmentation leads to a more pronounced decline, with the top-$1$ accuracy decreasing by $5.9\%$ and the top-$10$ accuracy experiencing a reduction of $9.5\%$, underscoring its role in enhancing model robustness and generalizability to unseen data. The performance deteriorates dramatically without node alignment, with the accuracy dropping by $50\%$ for Top-$10$. This highlights the importance of node alignment in capturing the structural information of the graphs and enabling the model to make more accurate predictions. It is evident that the difference between top-$1$ and top-$10$ accuracy significantly reduces from 34.8\% (when both data augmentation and node alignment are employed) to merely 3.8\% (when data augmentation is used without node alignment). This illustrates that using data augmentation without node alignment leads to a lower diversity of candidate predictions compared to when both techniques are employed. However, it is worth noting that the gap remains at 22.3\% when neither node alignment nor data augmentation is utilized, which is not particularly high. Consequently, this highlights that node alignment and data augmentation are complementary techniques that, when jointly employed, can enhance the performance metrics. In conclusion, our ablation study shows that node alignment, data augmentation, and graph features are all crucial components of our model, with each strategy playing a vital role in enhancing prediction accuracy. The combination of these strategies yields the best overall performance, emphasizing the importance of incorporating them in future work on graph-based reaction prediction models.

\begin{table}[h]
\centering
\caption{Top-$k$ validity of the generated molecules on USPTO-50k with reaction class unknown.}
\begin{tabular}{lcccc}
 \toprule
 \multirow{2}{*}{\textbf{Model}} & \multicolumn{4}{c}{\textbf{Top-$k$ Validity (\%)}} \\
 \cline{2-5} 
 & 1 & 3 & 5 & 10 \\
 \midrule
 \name{} (ours) & \textbf{99.7} & \textbf{98.6} & \textbf{97.1} & \textbf{92.9} \\
 \name{} w/o charge & 89.9 & 90.2 & 86.1 & 75.9 \\
 \name{} w/o hydrogen & 89.6 & 88.4 & 87.6 & 83.4 \\
 \name{} w/o charge or hydrogen & 80.8 & 82.5 & 81.5 & 77.6 \\
 \midrule
 GET \cite{get} & 97.8 & 86.6 & 80.5 & 70.7 \\
 Graph2SMILES \cite{graph2smiles} & 99.4 & 90.9 & 84.9 & 74.9 \\
 RetroPrime \cite{retroprime} & 98.9 & 98.2 & \textbf{97.1} & 92.5 \\
 \bottomrule
\end{tabular}
\label{tab:validity}
\end{table}
The other Table ~\ref{tab:validity} illustrates the Top-$k$ validity of various models on the USPTO-50k dataset, focusing on the auto-regressive node generation approach adopted by the NAG2G model. When all atomic features (type, formal charge, num of hydrogens) are utilized, \name{} delivers a Top-$1$ validity of $99.7\%$, with strong performances also observed in Top-$3$ ($98.6\%$), Top-$5$ ($97.1\%$), and Top-$10$ ($92.9\%$) predictions. The exclusion of certain molecular properties during node generation reveals the importance of each feature when generating SMILEs. Omitting just the formal charge results in a Top-$1$ validity of $89.9\%$, while disregarding the hydrogens brings it to $89.6\%$. The most pronounced decrease is observed when both these features are excluded, dropping the Top-$1$ validity to $80.8\%$. The results with respect to the accuracy are also studied, see supporting information for details. When juxtaposed with other models like \textit{GET}, \textit{Graph2SMILES}, and \textit{RetroPrime}, \name{} consistently outperforms. 

Furthermore, our ablation studies on the encoder reveal that the performance of our model is robust and not overly dependent on its encoder configuration. For detailed findings, please refer to the supplementary information provided.

\subsection{Case studies}

\begin{figure}
		\centering		\includegraphics[width=0.9\textwidth]{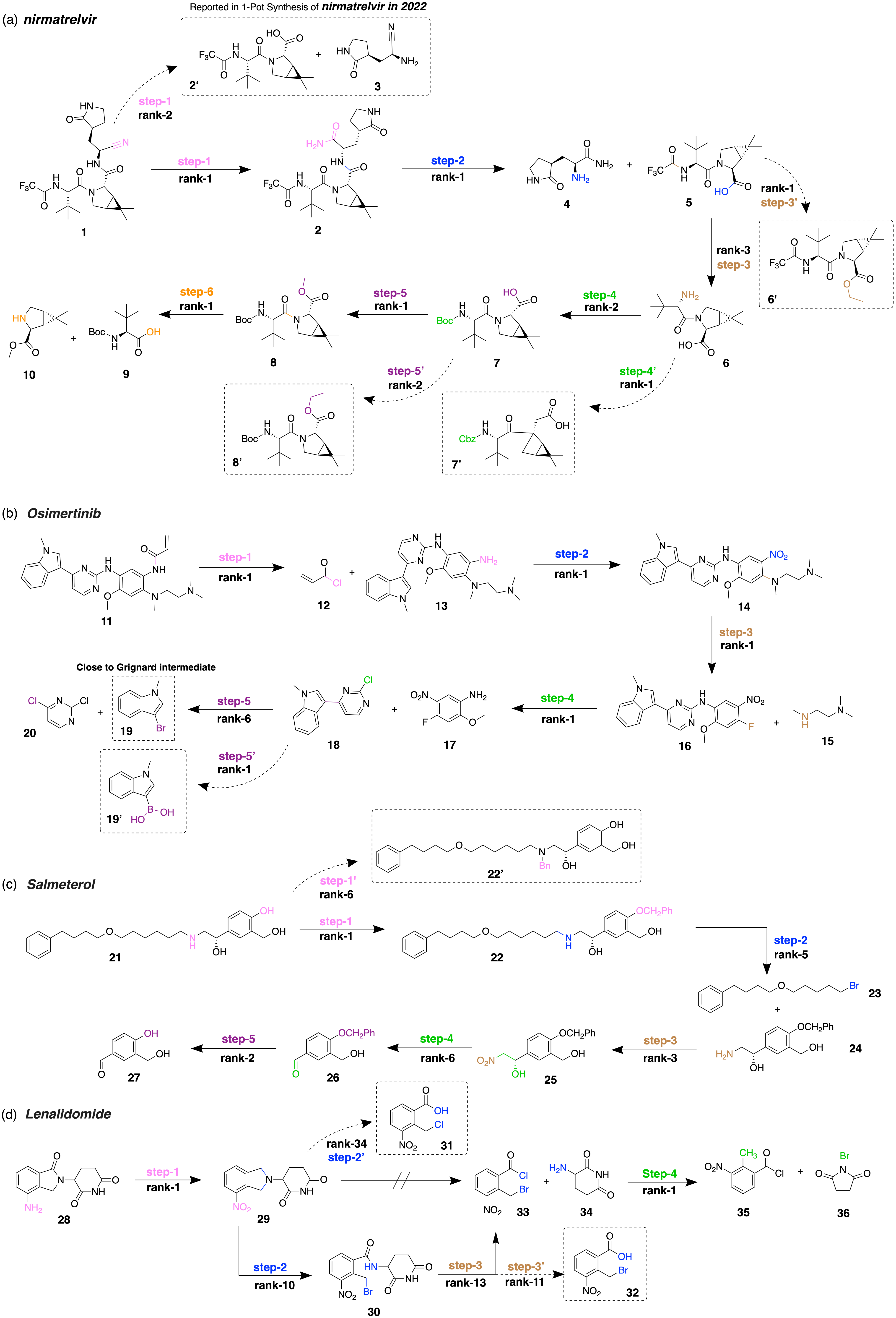}
		\caption{The synthesis route of four drug molecules and the predicted ranks given by \name{}. }
		\label{case}
\end{figure}

To assess the capability of designing synthetic routes for organic molecules, we picked four drug candidates as the target product and performed sequentially SSR using \name{} trained with USPTO-50k dataset. The performance of template-free \name{} successfully outperforms the previous model~\cite{zhong2023retrosynthesis} according to its inference results as shown in Figure~\ref{case}(a). All six synthetic steps documented in the original literature for Nirmatrelvir~\cite{hammond2022oral} are accurately predicted by \name{}, achieving all predictions within the top three ranks. The initial step involving the dehydration of the amide to form the cyano group was ranked first, while the rank predicted condensation reaction aligns with a recent advanced one-pot synthesis strategy for Nirmatrelvir~\cite{caravez20221}. The exact step-2 and step-6 condensations are pulled out by \name{} with both answers standing out among other candidates. In the third step, the trifluoroacetylation of the amine in \textbf{6} was predicted as the third-ranked reaction by our model, an improvement over its sixth-ranked prediction in \textit{Graph2Edits}. For step-4 and step-5, our model's first- and second-ranked predictions effectively serve the protective function using different, yet common, reagents.\cite{owen2021oral} The second test case, Osimertinib, known by its research name AZD9291, gained approval in 2014 as a clinical treatment for patients with non-small cell lung cancer (NSCLC).~\cite{finlay2014discovery} As shown in figure~\ref{case}(b), \name{} successfully delineates a five-step synthesis as described in the literature, tracing the synthetic pathway from commercially available pyrimidines to the final product AZD9291. The initial step of the reverse synthesis is an acylation reaction, ranking the first in order of likelihood, followed by another rank-1 reduction of the nitro group. Subsequently, it correctly identified two consecutive nucleophilic aromatic substitution steps as the top choices. In the final step, the model's highest probability prediction was a Suzuki coupling, which was also the rank-2 reaction inferred by \textit{Graph2Edits} model~\cite{zhong2023retrosynthesis}. Although the original strategy involving a Grignard reaction was not predicted, the rank-7 result suggests the alternative pathway. For the third case, we selected Salmeterol, a long-acting bronchodilator, which has been tested in the former template-based model by Coley~\cite{coley2017computer}, \textit{LocalRetro}\cite{localretro} and root-aligned strategy by Zhong\cite{zhong2022root}. With the presence of amine and phenol functional groups in compound \textbf{21}, \name{} recommends protecting the reactive sites, ranking these steps as first and second.~\cite{hett1994enantioselective,guo2011enantioselective} The subsequent reaction is the production of compound \textbf{22} through a Williamson ether-type reaction, identified as the rank-5 choice. Moreover, the rank-6 for the fourth step successfully predicts the transformation of \textbf{26} into \textbf{25} \textit{via} an asymmetric Henry reaction, as described in Guo's synthesis~\cite{guo2011enantioselective}. It is noteworthy that while the original synthesis protects two hydroxyl groups in compound \textbf{27} simultaneously using 2,2-Dimethoxypropane, \name{} opts to protect only the more reactive phenol group, which aligns with the requirements of the complete synthetic scheme.

The last example, lenalidomide, which is also involved in the work of \textit{LocalRetro}~\cite{localretro} and \textit{Graph2Edit}~\cite{zhong2023retrosynthesis} turns out to be a more challenging task~\cite{ponomaryov2015scalable}. \name{} accurately predicted the first and last steps involving nitro reduction and NBS (N-Bromosuccinimide) substitution as the top-$1$ reactions. For the cyclization step involving the formation of two C-N bonds, \name{} proposed a precursor, compound \textbf{31}, with a chlorine substituent instead of bromine. Nonetheless, \name{} is capable of suggesting a step-wise ring closure mechanism, mirroring the results reported by \textit{LocalRetro}.

\subsection{Error analysis}

In the sequence-to-sequence (seq2seq) model~\cite{retro_seq2seq} context, the generation of reactants in the SMILES format can result in three outcomes: 1) SMILES strings generated by RDKit are invalid, corresponding to a non-chemically feasible structure. 2) The SMILES strings are chemically valid but do not represent suitable reactants capable of producing the desired product under given reaction conditions. 3) The SMILES strings represent reactive compounds that can lead to the products as common reactions, even though they may not match the ground truth reactants exactly. In evaluating the first type of error for our model on the USPTO-50k dataset, we focused on validity, which gauges the percentage of valid SMILES strings generated among the top-$k$ predictions. Our model boasts a top-$1$ validity of 99.7\%, outperforming other advanced models ~\cite{get,graph2smiles,retroprime}. This superior validity is due to the autoregressive generation process as we have discussed in the ablation study section. If the model randomly omits predictions for charge and hydrogen attachments, the top-$1$ validity significantly decreases to 80.8\%. The identification of the second and third types of errors involves expert evaluation by organic chemists. To impartially review the inference results without pre-assigned reaction classes, we selected three representatives to examine the variety of reactants that can be predicted by our model, referred to here as \name{}. In all cases, the ground truth falls within the top 10 predictions, alongside a variety of other reaction types, showcasing the model's extensive predictive range. The rich diversity of suggested synthetic solutions enhances retrosynthetic route design by providing a wide array of chemical reactions to consider, thus offering alternatives for subsequent route selection and evaluation. For a detailed summary table and analysis, please refer to supplementary information. \\
Another limitation stemming from the origins of the USPTO dataset is the lack of detailed reaction information, such as conditions, yields, and selectivity. With this information, the NAG2G model could provide more accurate and reliable rankings for its predictions. Additionally, single-step prediction models, including ours, might overlook interactions between consecutive steps. For example, in case 3, the selective phenolic benzylation, while an efficient protecting step, could significantly impact the subsequent asymmetric Henry reaction. This scenario illustrates that effective single-step predictions require not just accuracy, but also a thorough evaluation of route complexity and yield trade-offs. In response to these challenges, we are actively developing advanced scoring methods to facilitate multi-step process decisions based on the results provided by NAG2G.

\section{Conclusion}

In this study, we present the Node-Aligned Graph-to-Graph (\name{}) model — a graph-based SSR method free from templates. This model employs the Transformer encoder-decoder framework, generating reactant molecule graphs in an auto-regressive fashion. Testing on well-established datasets, USPTO-50k and USPTO-Full, indicates that \name{} offers competitive performance against prevailing SOTA models. Ablation studies shed light on the contributions of various components, underlining the potential of our approach. The replication of case studies and error analysis highlights the superior performance of \name{} in specific SSR tasks, suggesting that further advancements could enhance its capabilities even more.

While many SSR models, detailed in our introduction, show promise, \name{} marks a notable stride in applying deep learning to single-step retrosynthesis—especially when considering a template-independent approach. Our methodology opens the door that intricate neural networks aren't the sole route to excellence; meticulous model design, combined with precise task definitions, can yield competitive results. Our current design is tailored for single-step retrosynthesis predictions, where input and output graphs bear close resemblance. For broader graph-to-graph generation tasks, especially with considerable input-output disparities, refinements may be needed. Moving forward, our ultimate goal is to evolve this method, delving into multi-step synthesis planning for more intricate chemical synthesis scenarios. \looseness=-1

\begin{suppinfo}
The NAG2G  is available at https://github.com/dptech-corp/NAG2G \\
Experimental results related to NAG2G’s performance, auxiliary ablation studies, and prediction examples. (PDF)
\end{suppinfo}

\subsection{Author contribution}
W.G. and Z.W. contributed equally to this work.

\begin{acknowledgement}
Lin Yao and Wentao Guo thank the support from Tuantuan during the project. 
\end{acknowledgement}


\newpage
\begin{suppinfo}

Experimental results related to \name{}'s performance, auxiliary ablation studies.

\section{Related work on graph generation}

Graph generation can be broadly categorized into two classes: global models and sequential models. Global models generate the entire graph structure simultaneously, as opposed to sequentially. These models often employ a pairwise matrix, such as an adjacency matrix, to represent the graph and learn to generate the matrix. Commonly used global graph generation techniques include autoencoder-based models~\cite{kipf2016variational,simonovsky2018graphvae,flam2020graph,ma2018constrained,grover2019graphite,guo2020interpretable,li2020dirichlet}, Generative Adversarial Networks (GANs)~\cite{de2018molgan,yang2019conditional,yang2020learn}, and flow-based generative models~\cite{liu2019graph,madhawa2019graphnvp}.

In contrast, sequential models generate graphs incrementally. Many previous works fall into this category, such as MolRNN~\cite{li2018multi}, GraphRNN~\cite{you2018graphrnn}, MolecularRNN~\cite{popova2019molecularrnn}, Bacciu et al.'s works~\cite{bacciu2019graph,bacciu2020edge}, GraphGen~\cite{goyal2020graphgen}, MolMP~\cite{li2018multi}, GRAN~\cite{liao2019efficient}, GRAM~\cite{kawai2019scalable}, AGE~\cite{fan2020attention}, DeepGMG~\cite{li2018learning}, and BiGG~\cite{dai2020scalable}. Sequential models are simple and effective, but suffer from the need to specify a generation order for nodes.

The proposed \name{} leverages the node alignment strategy to determine node generation order, thus addressing the challenge in sequential generation. It is important to note that the proposed \name{} model is primarily designed for single-step retrosynthesis prediction, as the input and output graphs exhibit small differences, allowing for node alignment. However, for general graph generation problems, the proposed method may not perform optimally.

\section{More experimental results} 

\subsection{Ablation study on encoders}
In NAG2G, we directly employ the existing model backbones from previous works as encoders. To investigate the performance improvement attributed to the encoders, we conduct an ablation study. Table~\ref{tbl:add_abl1} demonstrates that, apart from the default Uni-Mol+ \cite{lu2023highly}, we also evaluate the performance of Graphormer \cite{graphormer}. The results indicate that the overall performance is comparable. This observation suggests that the choice of encoders does not significantly impact the final performance of NAG2G.

 \begin{table}[ht]
 \centering
 \caption{The performance with different encoders on USPTO-50k with reaction class unknown.}
 \vspace{-0.05in}
 \begin{tabular}{cccccc}
 \toprule
&& \multicolumn{4}{c}{\textbf{Top-$k$ Accuracy (\%)}} \\
\cline{3-6} 
 \textbf{Encoder Type} && 1 & 3 & 5 & 10 \\
 \midrule
 Default (Uni-Mol+) \cite{lu2023highly} && \textbf{55.1} & 76.9 & \textbf{83.4} & \textbf{89.9} \\

 Graphormer \cite{graphormer} && 54.3 & \textbf{77.0} &\textbf{83.4} & 89.0 \\
 \bottomrule
 \end{tabular}
 \label{tbl:add_abl1}
 \end{table}

\subsection{Accuracy and validity of the generated molecules}

In order to accurately represent a molecule, it is crucial to take into account not only the types of atoms and bonds but also the formal charges of atoms and the number of attached hydrogen atoms. Although software tools such as RDKit have the potential to provide this information, they often fail to do so or deliver inaccurate data. To tackle this problem, \name{} generates these features end-to-end, alongside atom types and bonds. The impact of incorporating this additional information is assessed through two ablation studies. 

The first study we have shown in the manuscript evaluates the validity of the generated molecules, as depicted in Table~\ref{tab:validity}. Moreover, the second study examines the overall performance, which is presented in Table~\ref{tbl:add_abl2}. Owing to the increased validity, the model is capable of generating a greater number of valid results, thereby significantly improving the overall performance.

\begin{table}[h]
 \centering
 \caption{The performance of \name{} with/without generating additional atomic features, on USPTO-50k with reaction class unknown.}
 \vspace{-0.05in}
 \begin{tabular}{ccccccc}
 \toprule
 &&& \multicolumn{4}{c}{\textbf{Top-$k$ Accuracy (\%)}} \\
 \cline{4-7}
 \textbf{Charges} & \textbf{Hydrogen Atoms} && 1 & 3 & 5 & 10 \\
 \midrule
 \ding{51} & \ding{51} && \textbf{55.1} & \textbf{76.9} & \textbf{83.4} & \textbf{89.9} \\
 \ding{51} & \ding{53} && 49.0 & 68.9 & 74.7 & 80.8 \\
 \ding{53} & \ding{51} && 48.1 & 68.2 & 73.9 & 79.4 \\
 \ding{53} & \ding{53} && 42.6 & 60.8 & 65.9 & 70.9 \\

 \bottomrule
 \end{tabular}
 \label{tbl:add_abl2}
 \end{table}

\subsection{Accuracy of the primary reagent (MaxFrag)}

The MaxFrag metric, previously proposed in the Aug. Transformer \cite{retro_at}, is utilized to assess the accuracy of the primary (largest) reactant prediction. This metric was introduced due to its significance in the classical procedure, where focusing solely on the main compound transformations provides the minimal information necessary to derive an efficient retrosynthesis route. As demonstrated in Table~\ref{tab:MaxFrag}, our proposed model, \name{}, outperforms the Aug. Transformer, signifying an improved ability to accurately predict the primary reagent based on a given product.

\begin{table}[h]
 \centering
\caption{The top-$k$ Accuracy of MaxFrag on USPTO-50k with reaction class unknown.}
\begin{tabular}{ccccc}
\toprule
\multirow{2}{*}{\textbf{Model}} & \multicolumn{4}{c}{\textbf{Top-$k$ MaxFrag Accuarcy (\%)}} \\
\cline{2-5} 
 & 1 & 2 & 5 & 10 \\
 \midrule
 Aug. Transformer \cite{retro_at} & 58.0 & 73.4 & 84.8 & 89.1 \\
\name{} (ours) & \textbf{59.7} & \textbf{73.6} & \textbf{86.3} & \textbf{91.9} \\
\bottomrule
\end{tabular}
\label{tab:MaxFrag}
\end{table}

\subsection{Accuracy of each reaction class}
In Table~\ref{tab:accu_single_cat}, we exhibit the top-$k$ accuracy results of \name{} on the USPTO-50k dataset, wherein the reaction classes were undisclosed during the training process, and the results are stratified according to the ground truth reaction classes.

\begin{table}[h]
\centering
\caption{The top-$k$ accuracy of each reaction class on USPTO-50k dataset, when trained with reaction class unknown.}
\begin{tabular}{lcccccc}
\toprule
\multicolumn{2}{c}{\textbf{Reaction Class}} & \multicolumn{1}{c}{\textbf{Reaction Fraction(\%)}} & \multicolumn{4}{c}{\textbf{Top-$k$ Accuracy (\%)}} \\
\cline{4-7}
 & && 1 & 3 & 5 & 10 \\
\midrule
heteroatom alkylation and arylation & & 30.3 & 54.4 & 75.5 & 82.3 & 90.5 \\
acylation and related processes & & 23.8 & 67.3 & 87.5 & 91.1 & 94.5 \\
deprotections & & 16.5 & 51.2 & 81.1 & 86.9 & 92.1 \\
C-C bond formation & & 11.3 & 40.0 & 61.6 & 71.8 & 81.7 \\
reductions & & 9.2 & 55.4 & 74.5 & 81.8 & 87.7 \\
functional group interconversion & & 3.7 & 34.8 & 53.3 & 64.1 & 76.1 \\
heterocycle formation & & 1.8 & 44.0 & 66.0 & 74.7 & 82.4 \\
oxidations & & 1.6 & 69.5 & 81.7 & 91.5 & 96.3 \\
protections & & 1.3 & 67.6 & 85.3 & 89.7 & 92.6 \\
functional group addition & & 0.5 & 87.0 & 87.0 & 87.0 & 91.3 \\
\bottomrule
\end{tabular}
\label{tab:accu_single_cat}
\end{table}

\subsection{Prediction example}
We present three examples of \name{} prediction results on the USPTO-50k test dataset without given reaction classes, as shown in Figure~\ref{fig:result_one}, Figure~\ref{fig:result_two}, and Figure~\ref{fig:result_three}. These examples demonstrate the capability of \name{} to predict plausible reactants, even when an exact match with the ground truth reactants is achieved at a lower ranking. The model is able to generate multiple predictions that can result in the desired product, highlighting its potential for application in real-world scenarios.

\newpage

\begin{figure}[htbp]
\centering
\begin{minipage}{0.32\linewidth}
 \centering
 \includegraphics[width=0.9\linewidth]{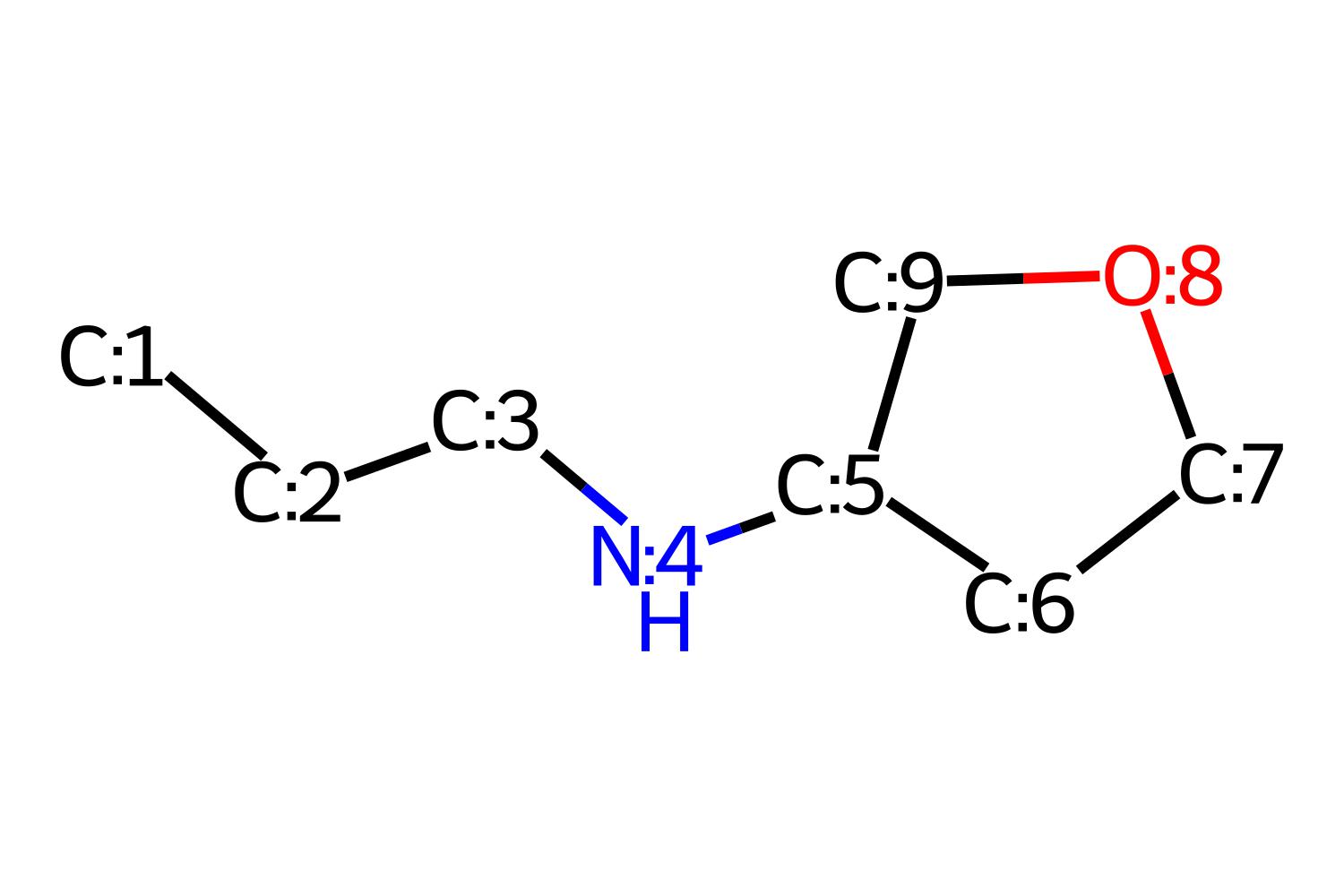}
 \centerline{Production}
\end{minipage}
\vspace{3pt}
\begin{minipage}{0.32\linewidth}
 \centering
 \includegraphics[width=0.9\linewidth]{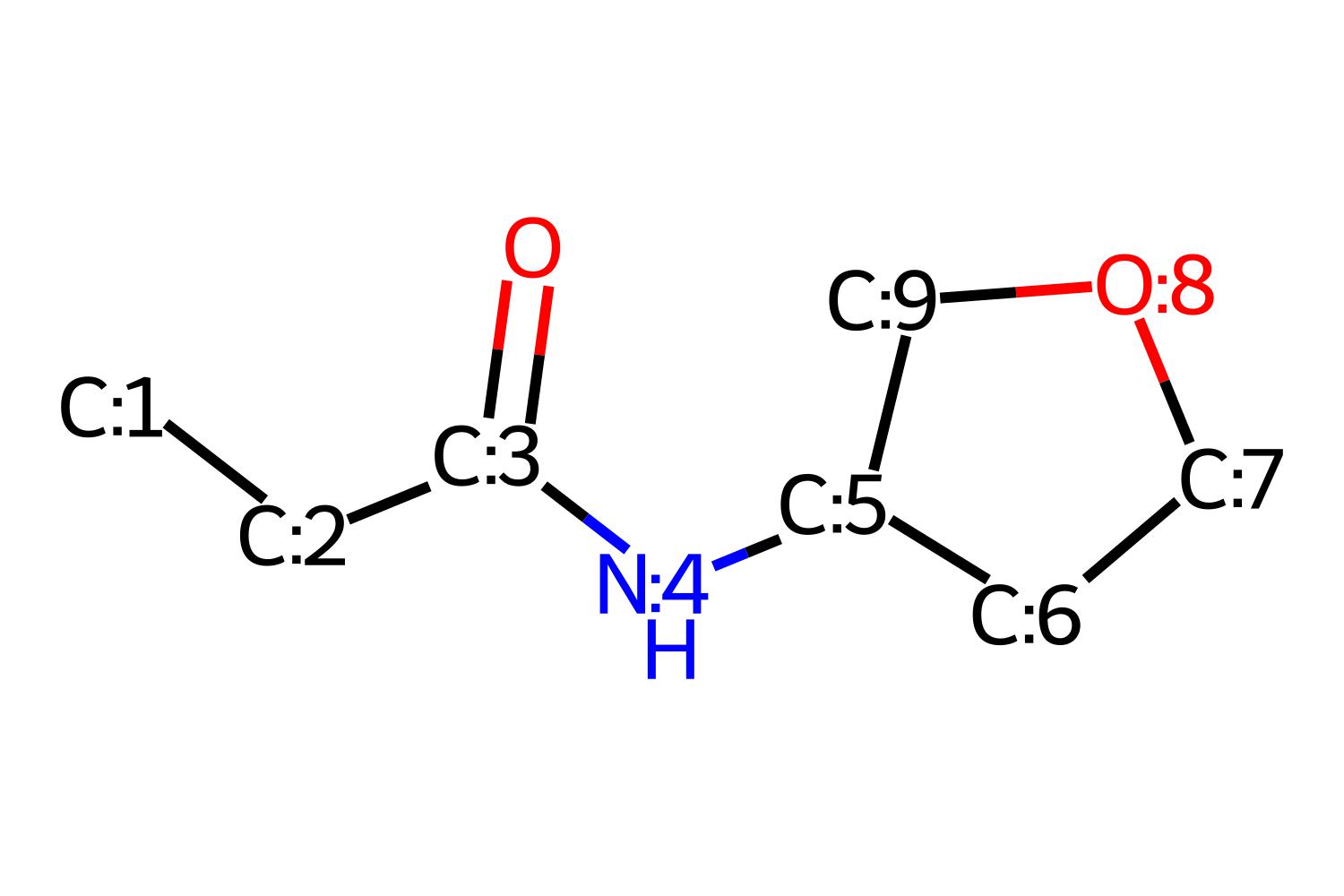}
 \centerline{Ground Truth Reactants}
\end{minipage}
\vspace{3pt}
\begin{minipage}{0.32\linewidth}
 \centering
 \includegraphics[width=0.9\linewidth]{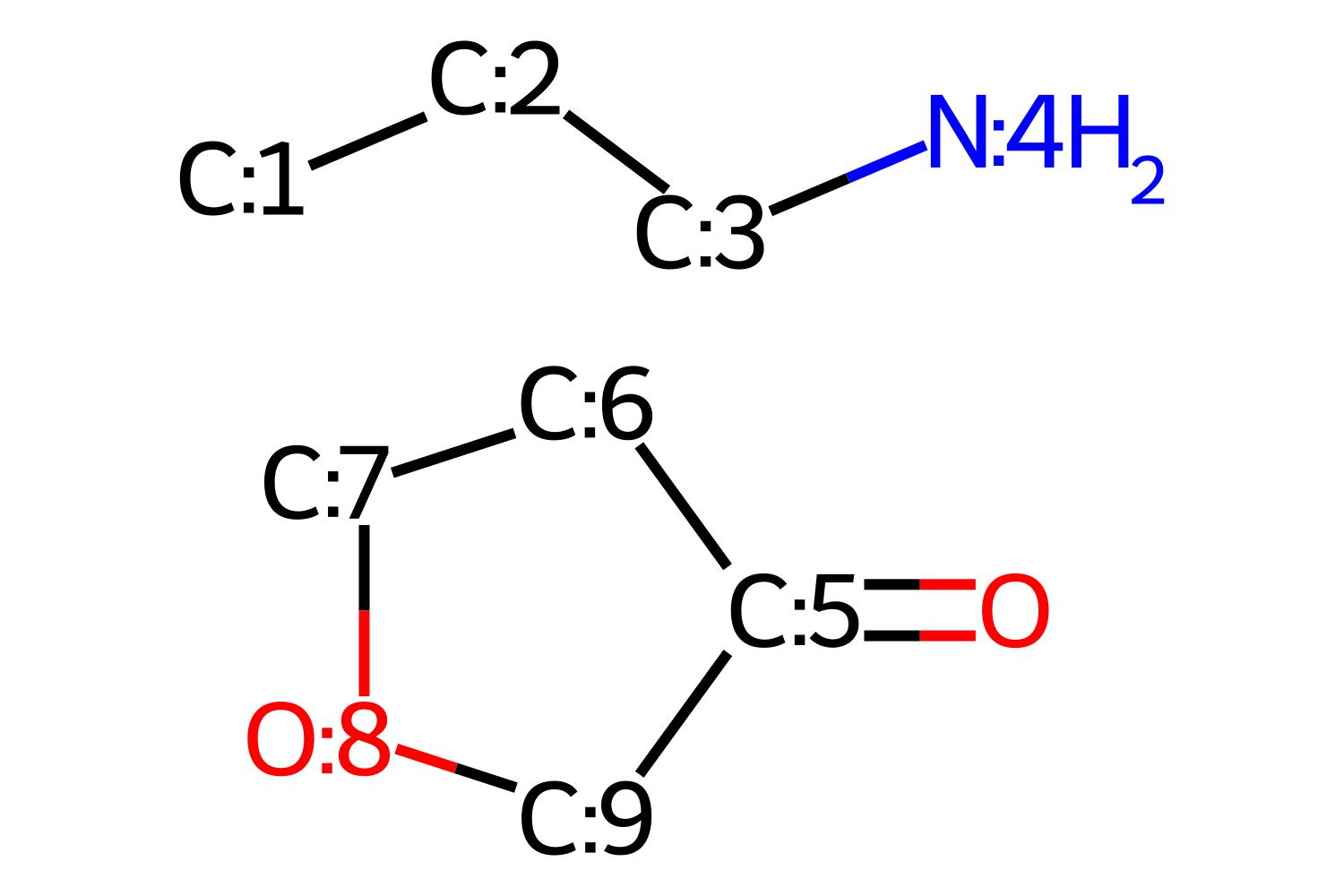}
 \centerline{Top 1 Prediction Reactants}
\end{minipage}
\vspace{3pt}
\begin{minipage}{0.32\linewidth}
 \centering
 \includegraphics[width=0.9\linewidth]{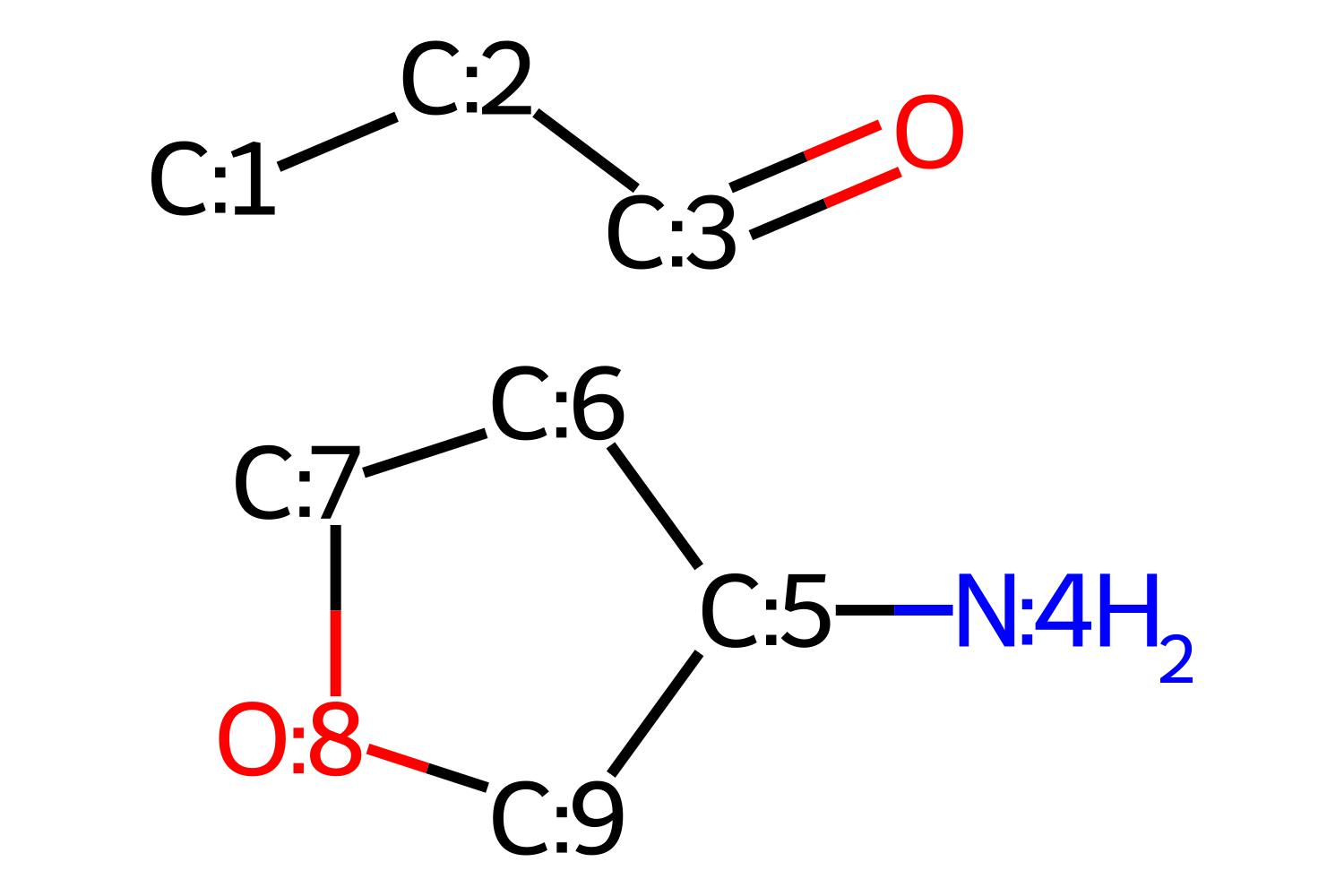}
 \centerline{Top 2 Prediction Reactants}
\end{minipage}
\vspace{3pt}
\begin{minipage}{0.32\linewidth}
 \centering
 \includegraphics[width=0.9\linewidth]{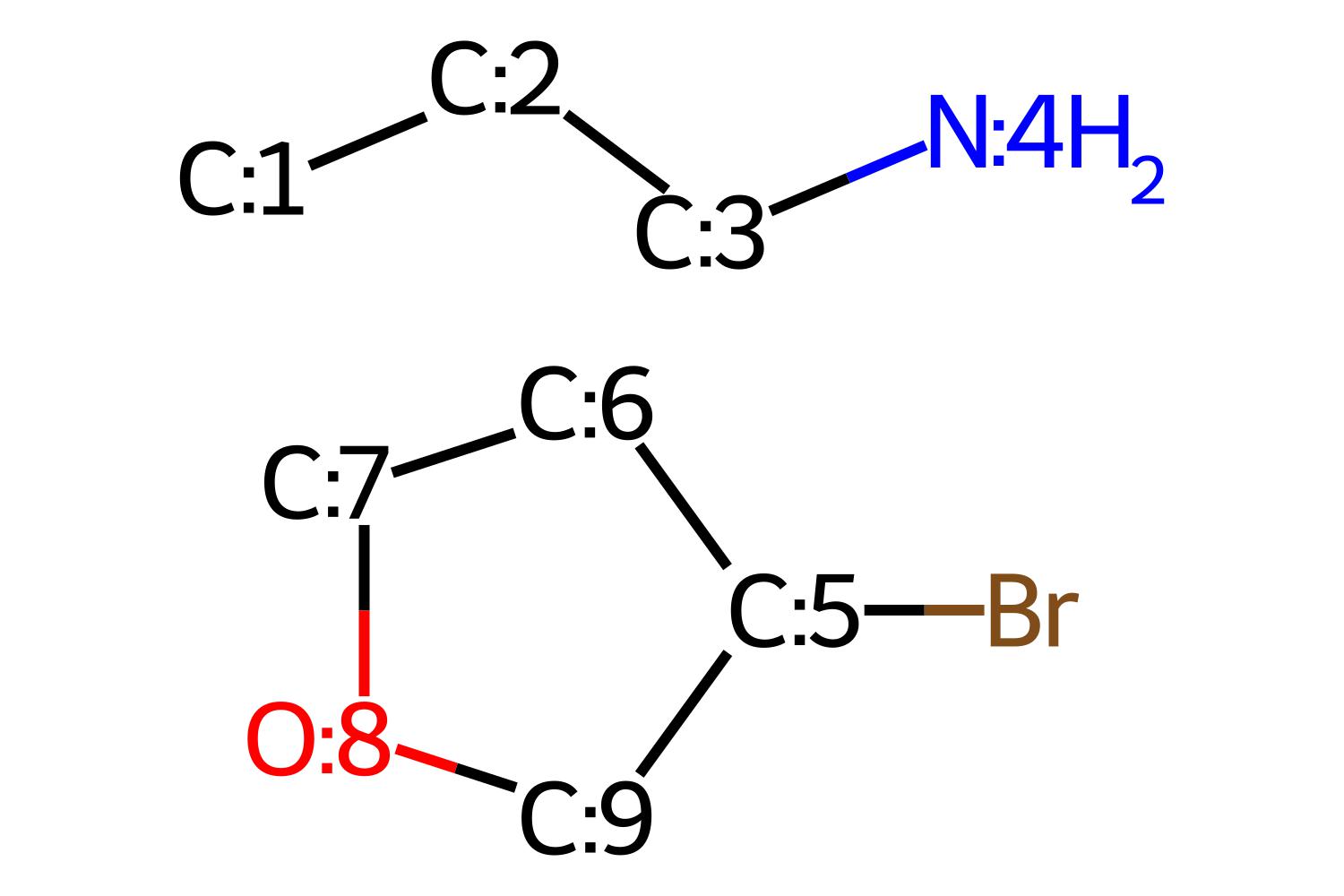}
 \centerline{Top 3 Prediction Reactants}
\end{minipage}
\vspace{3pt}
\begin{minipage}{0.32\linewidth}
 \centering
 \includegraphics[width=0.9\linewidth]{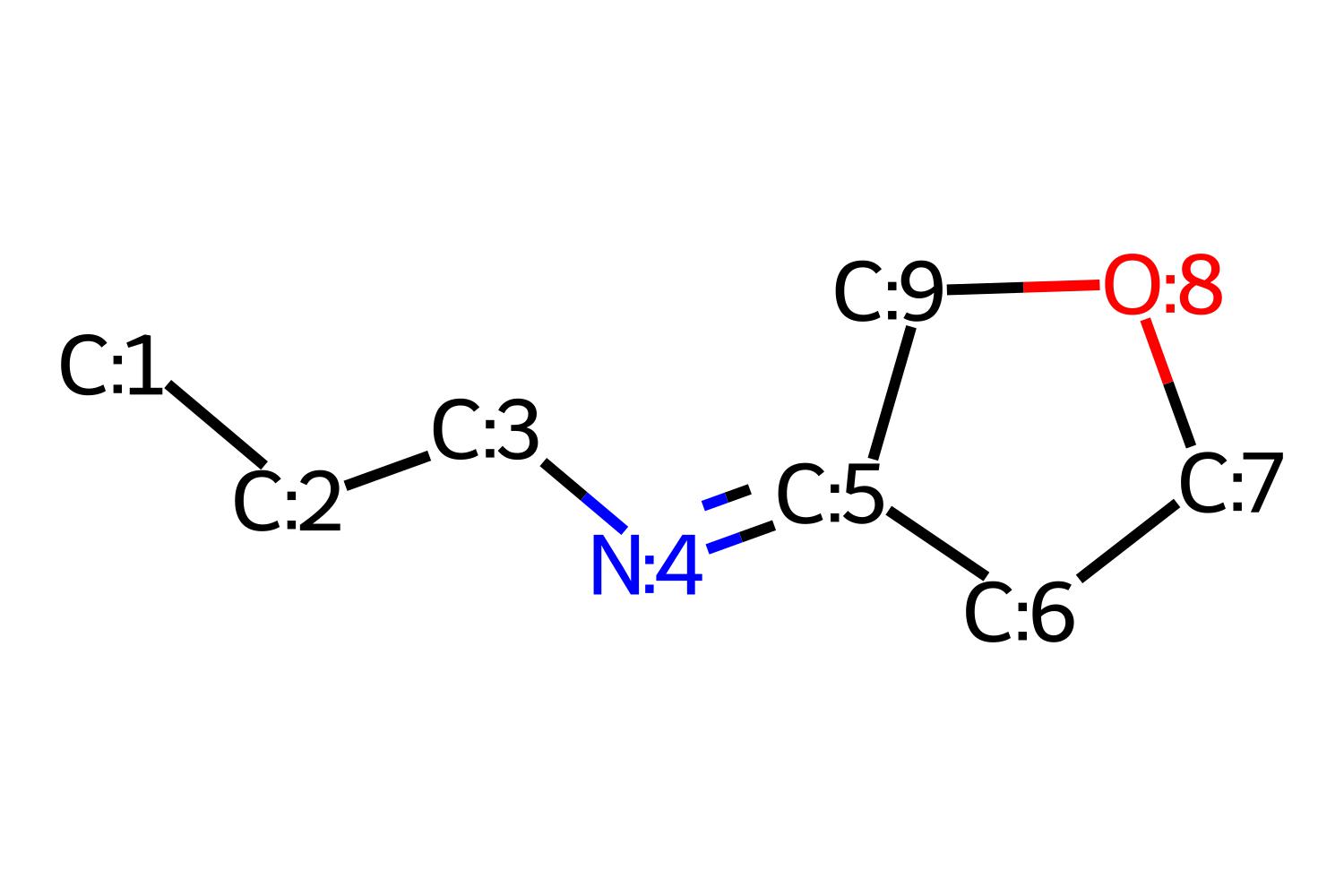}
 \centerline{Top 4 Prediction Reactants}
\end{minipage}
\vspace{3pt}
\begin{minipage}{0.32\linewidth}
 \centering
 \includegraphics[width=0.9\linewidth]{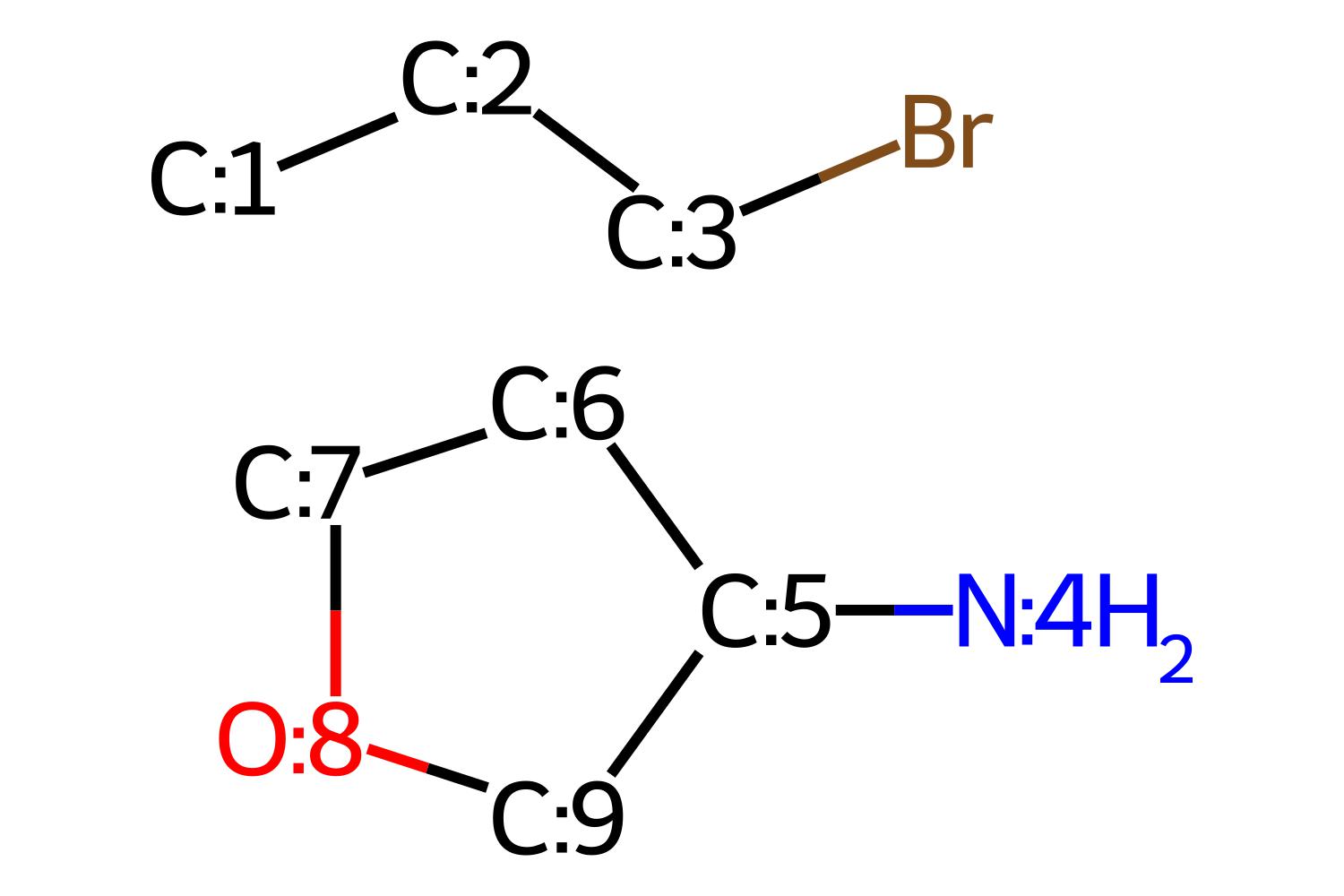}
 \centerline{Top 5 Prediction Reactants} 
\end{minipage}
\vspace{3pt}
\begin{minipage}{0.32\linewidth}
 \centering
 \includegraphics[width=0.9\linewidth]{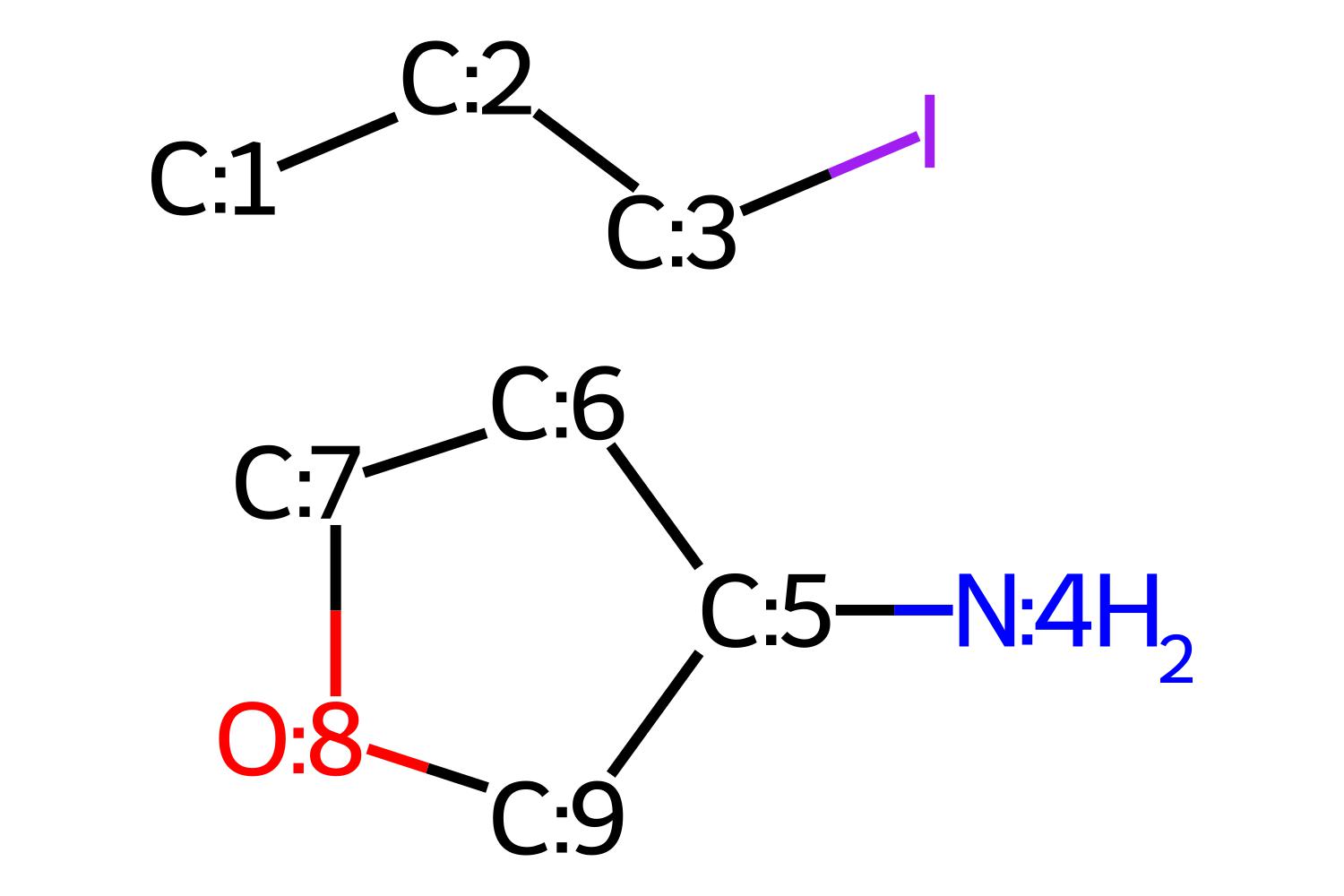}
 \centerline{Top 6 Prediction Reactants} 
\end{minipage}
\vspace{3pt}
\begin{minipage}{0.32\linewidth}
 \centering
 \includegraphics[width=0.9\linewidth]{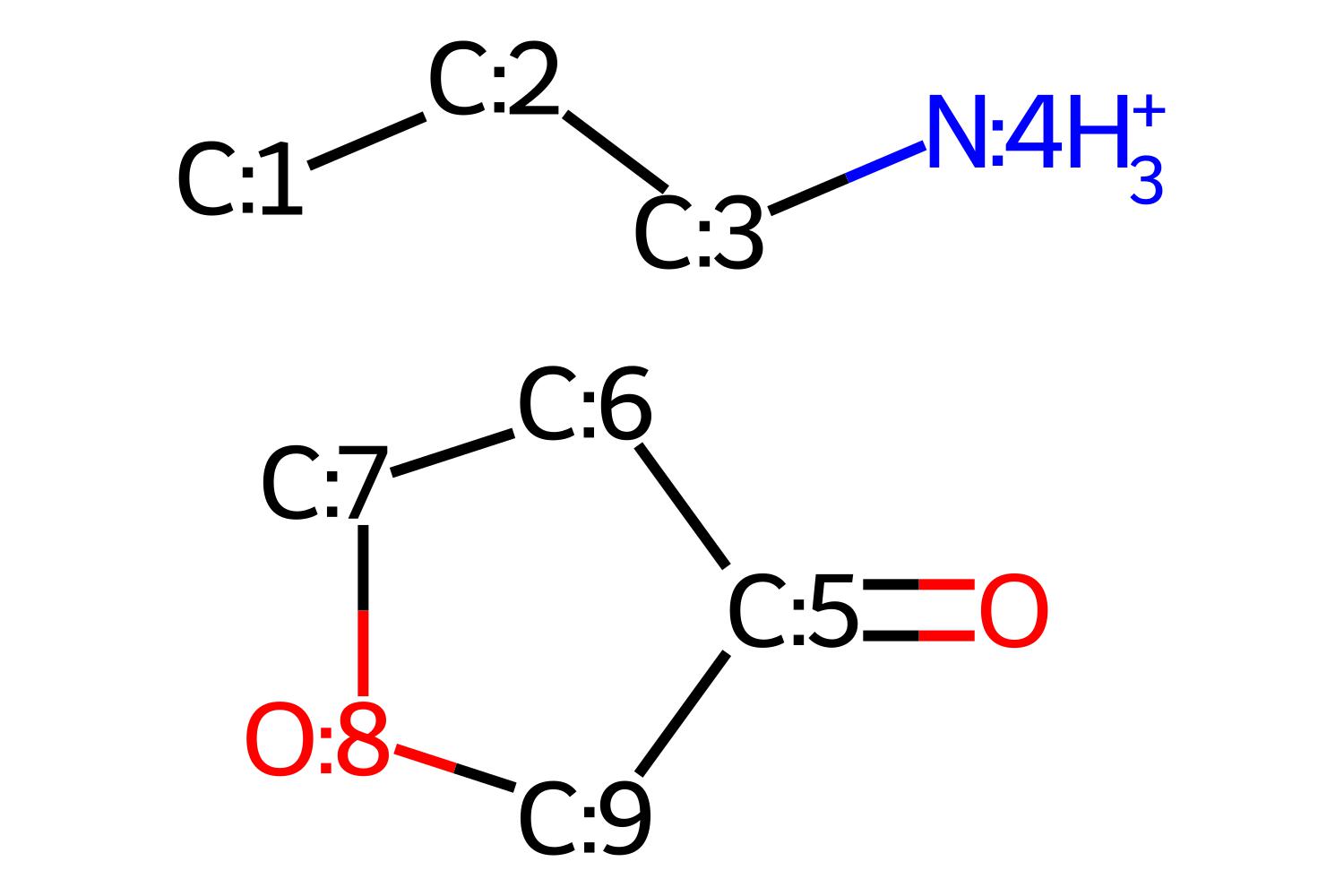}
 \centerline{Top 7 Prediction Reactants} 
\end{minipage}
\vspace{3pt}
\begin{minipage}{0.32\linewidth}
 \centering
 \includegraphics[width=0.9\linewidth]{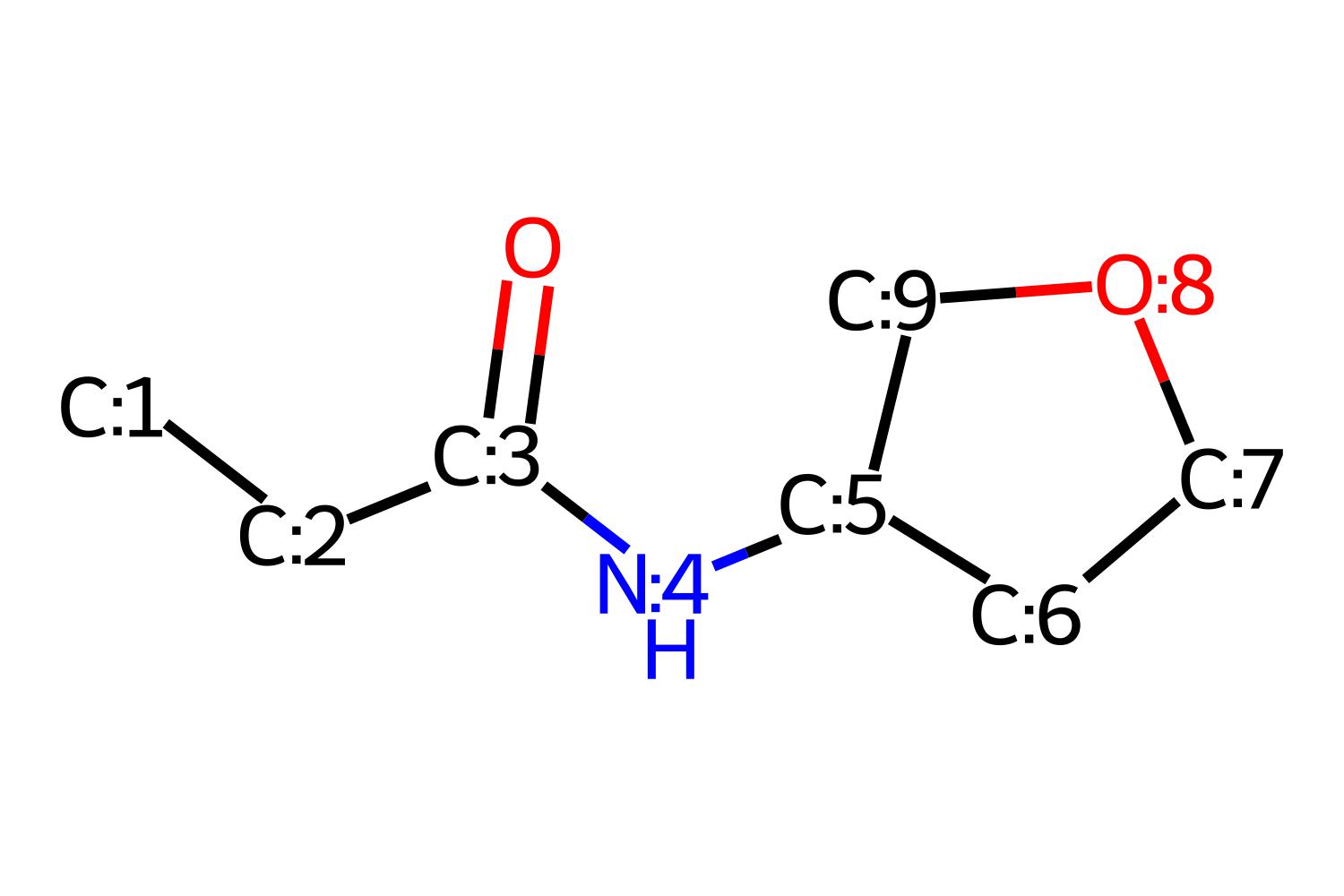}
 \centerline{Top 8 Prediction Reactants} 
\end{minipage}
\vspace{3pt}
\begin{minipage}{0.32\linewidth}
 \centering
 \includegraphics[width=0.9\linewidth]{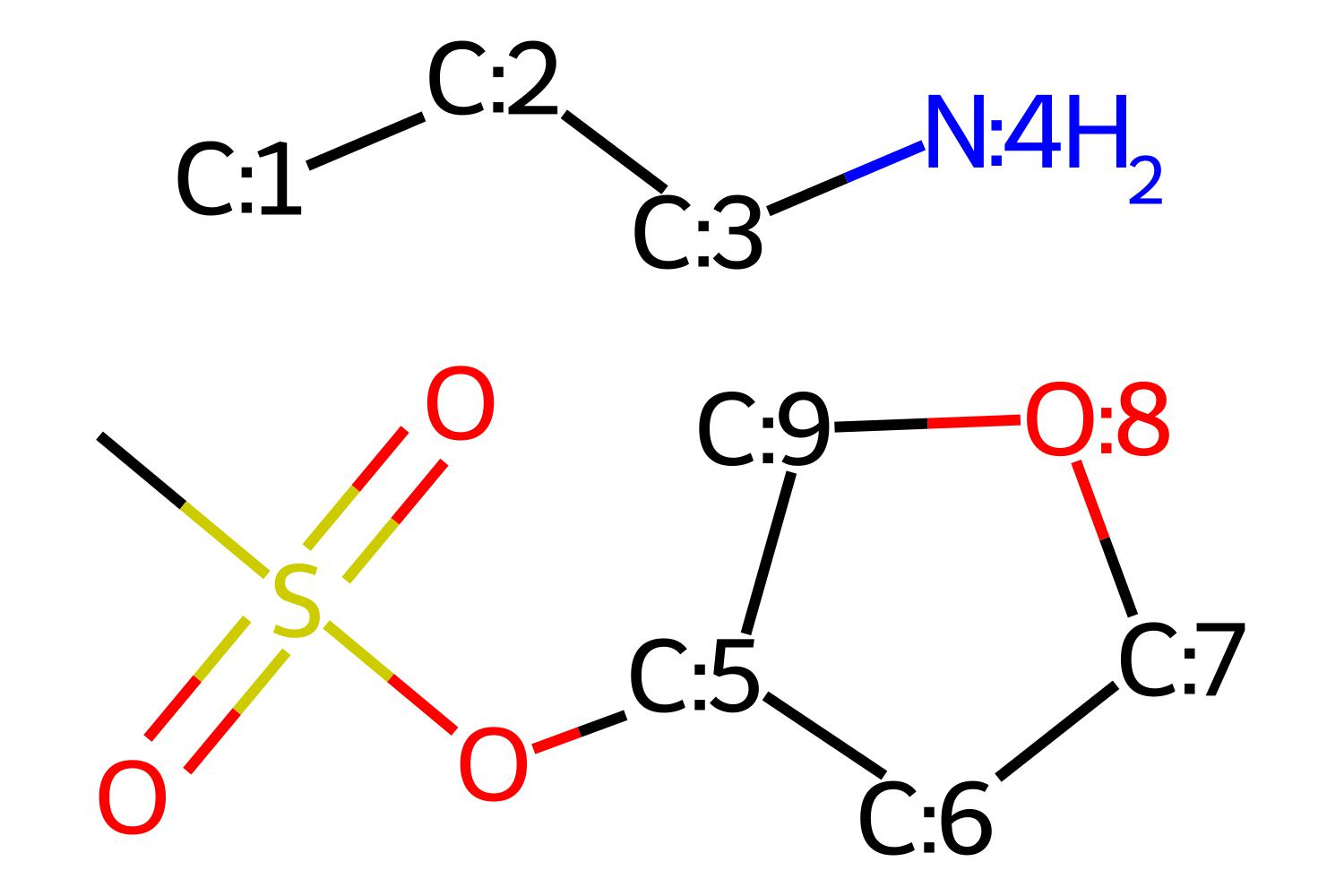}
 \centerline{Top 9 Prediction Reactants} 
\end{minipage}
\vspace{3pt}
\begin{minipage}{0.32\linewidth}
 \centering
 \includegraphics[width=0.9\linewidth]{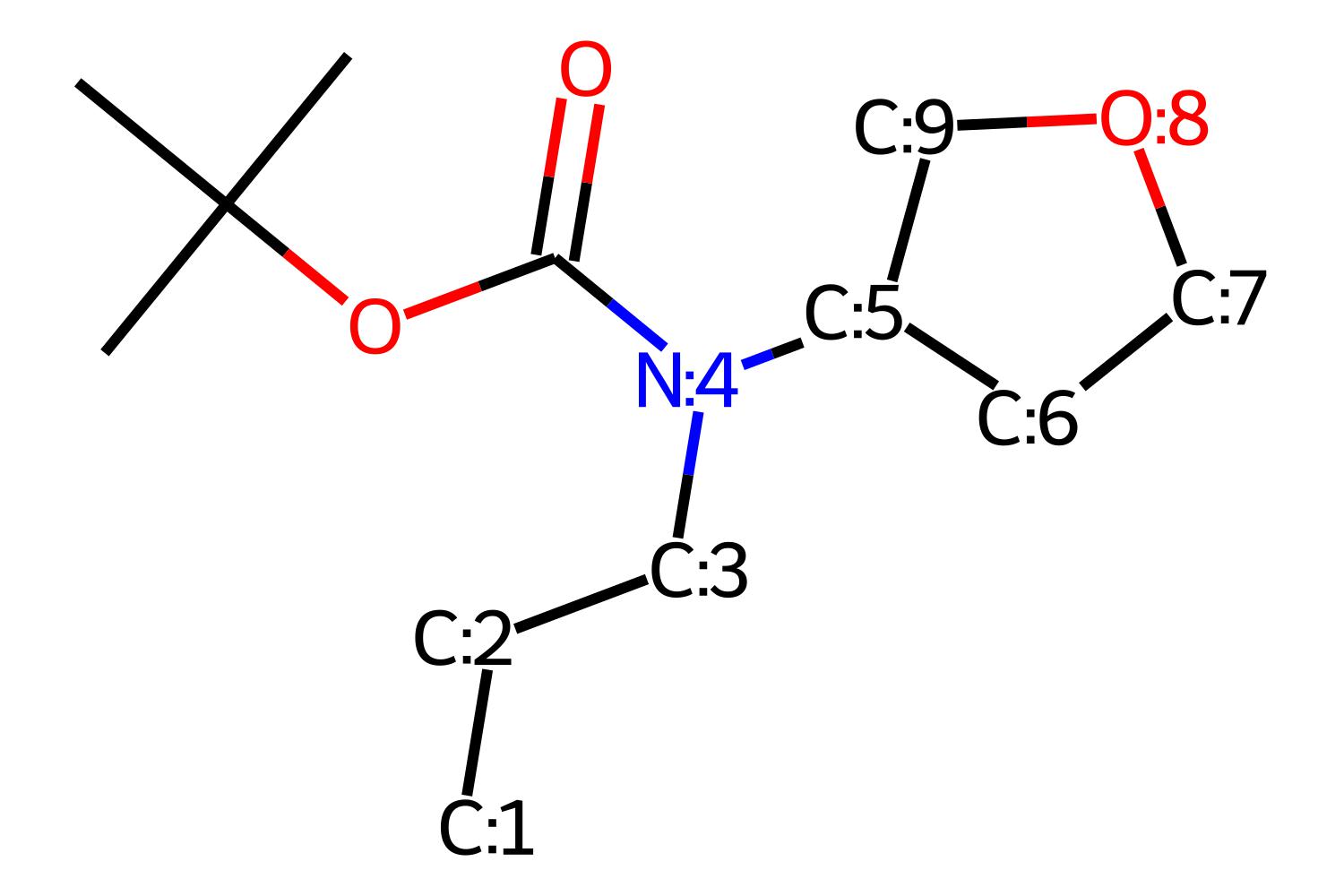}
 \centerline{Top 10 Prediction Reactants} 
\end{minipage}
\caption{This is example 1 of \name{} prediction on the USPTO-50k test dataset with the reaction class unknown. Although only the eighth predicted reactants precisely correspond to the ground truth reactants, all ten predicted reactions are chemically valid within reaction mechanisms. Specifically, the first, second, fourth, seventh, and eighth reactions can be classified as reduction reactions, while the third, fifth, sixth, and ninth reactions belong to the heteroatom alkylation and arylation reaction type. The tenth reaction is categorized as a deprotection reaction.}
\label{fig:result_one}
\end{figure}

\begin{figure}[htbp]
\centering
\begin{minipage}{0.32\linewidth}
 \centering
 \includegraphics[width=0.9\linewidth]{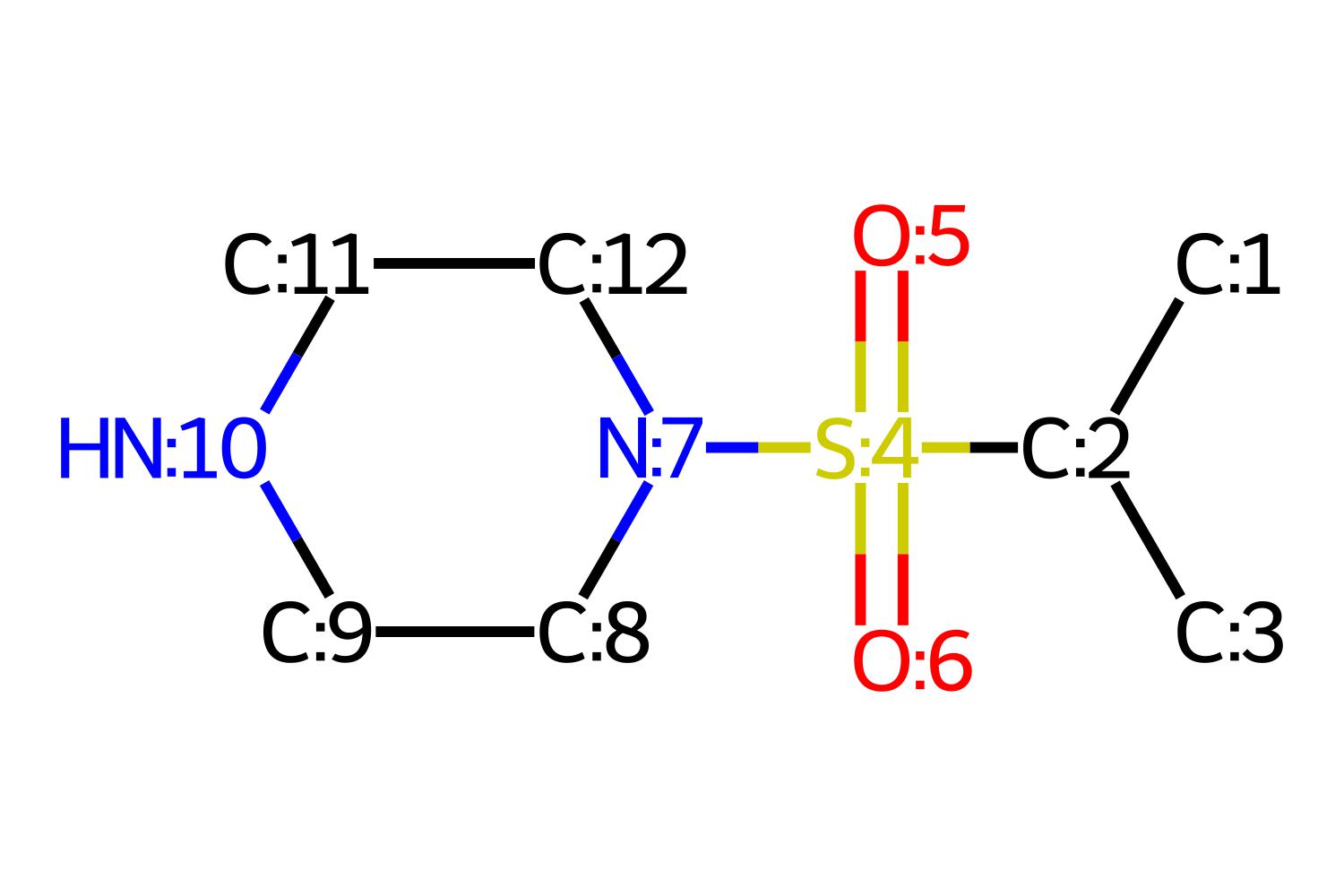}
 \centerline{Production}
\end{minipage}
\vspace{3pt}
\begin{minipage}{0.32\linewidth}
 \centering
 \includegraphics[width=0.9\linewidth]{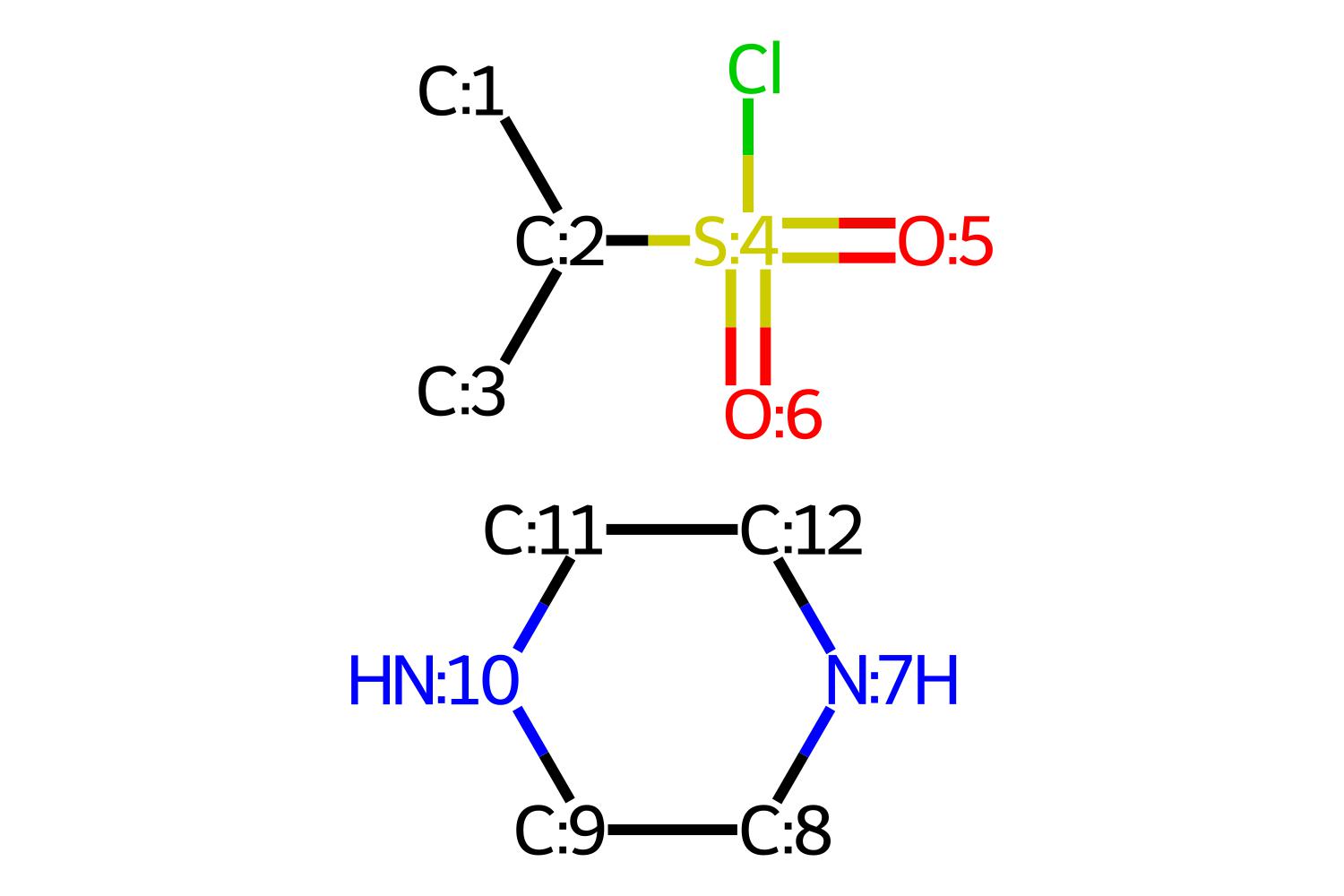}
 \centerline{Ground Truth Reactants}
\end{minipage}
\vspace{3pt}
\begin{minipage}{0.32\linewidth}
 \centering
 \includegraphics[width=0.9\linewidth]{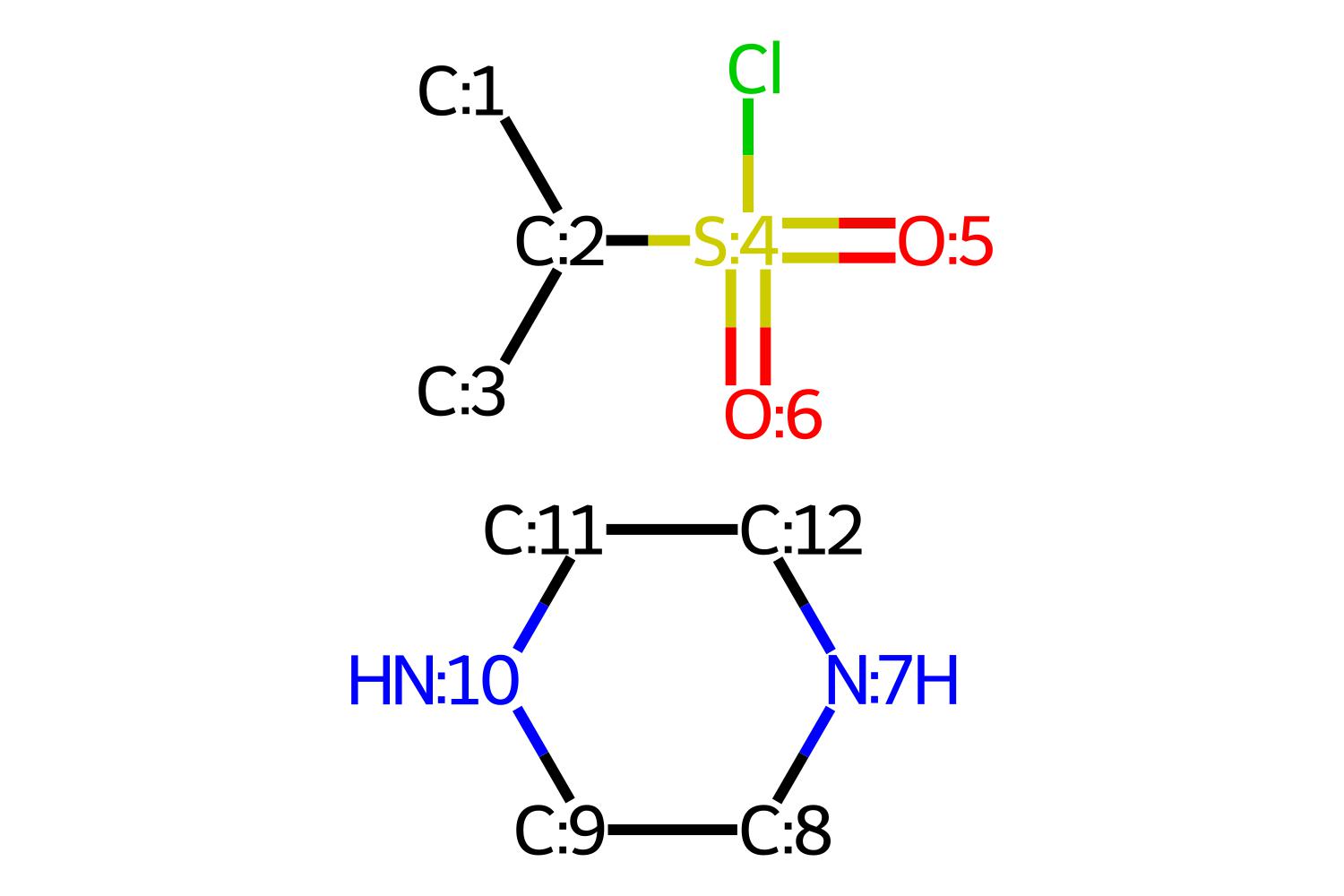}
 \centerline{Top 1 Prediction Reactants}
\end{minipage}
\vspace{3pt}
\begin{minipage}{0.32\linewidth}
 \centering
 \includegraphics[width=0.9\linewidth]{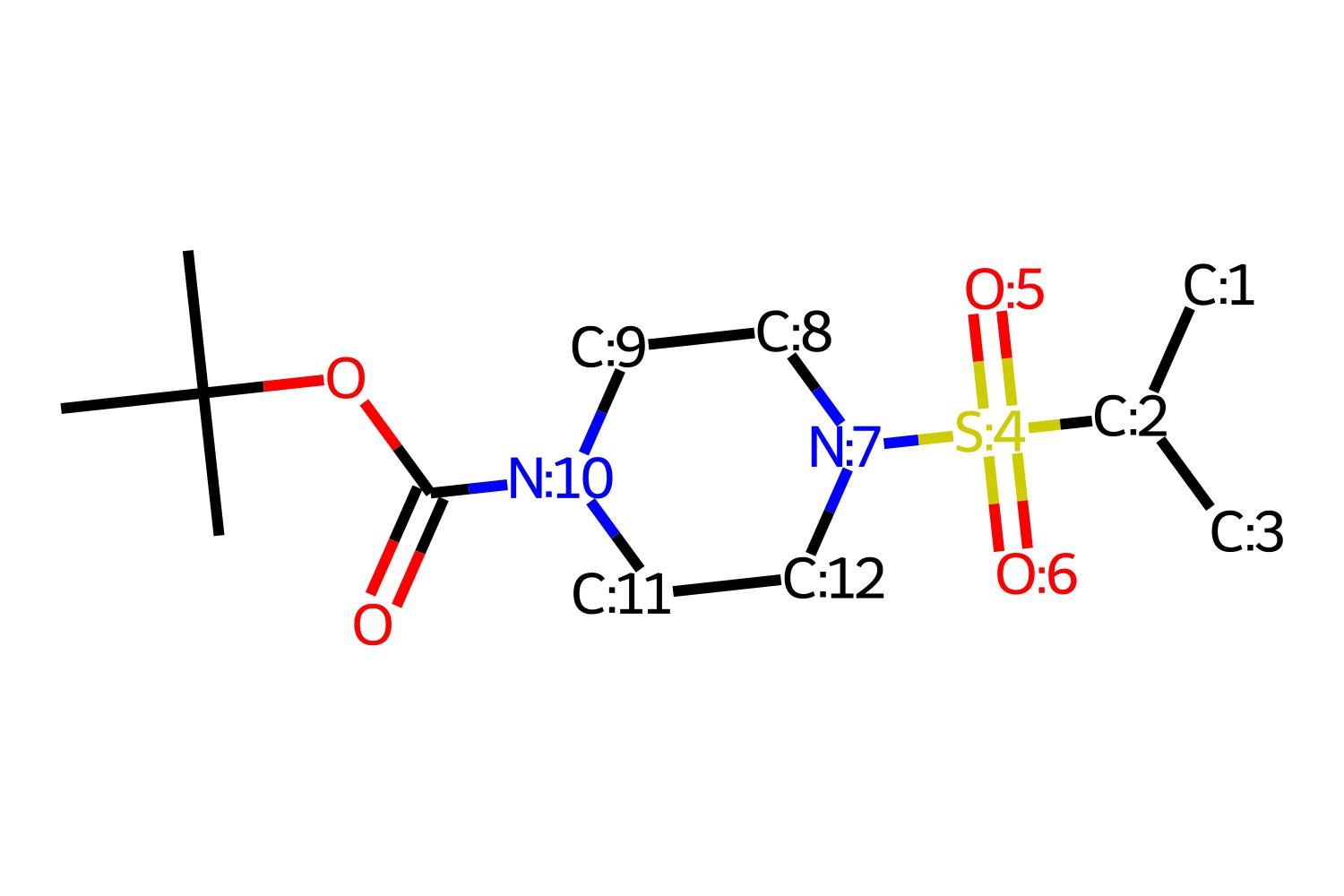}
 \centerline{Top 2 Prediction Reactants}
\end{minipage}
\vspace{3pt}
\begin{minipage}{0.32\linewidth}
 \centering
 \includegraphics[width=0.9\linewidth]{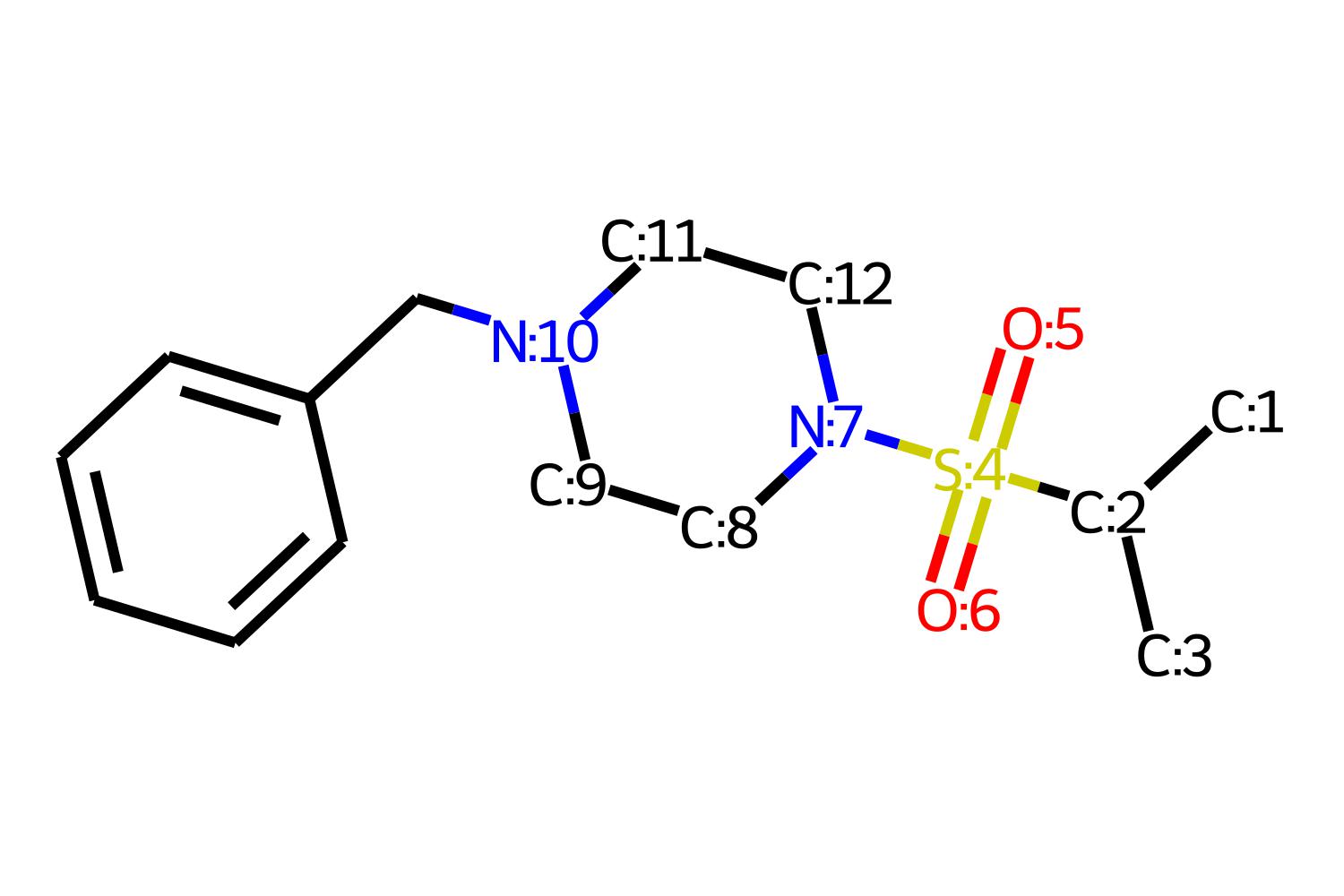}
 \centerline{Top 3 Prediction Reactants}
\end{minipage}
\vspace{3pt}
\begin{minipage}{0.32\linewidth}
 \centering
 \includegraphics[width=0.9\linewidth]{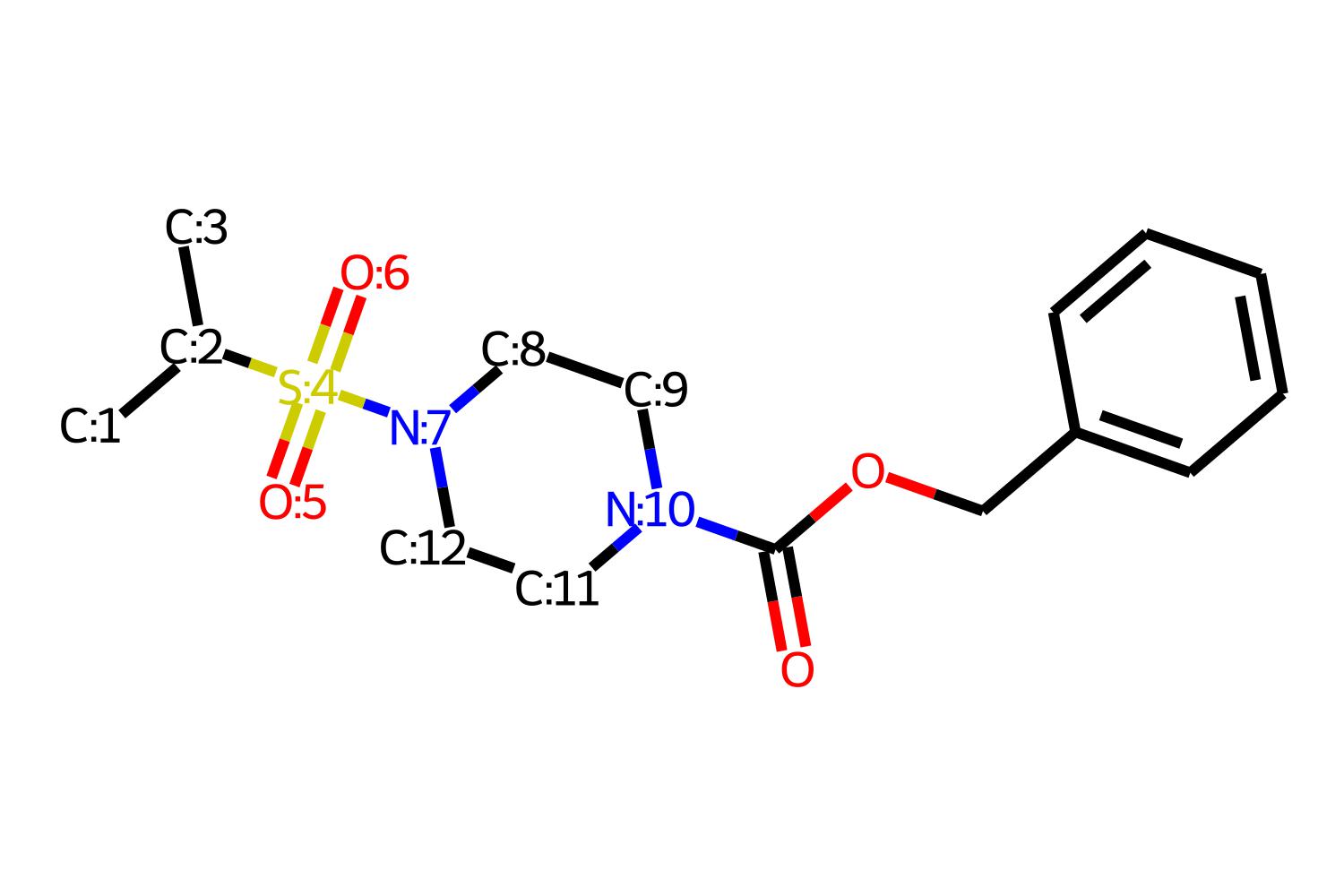}
 \centerline{Top 4 Prediction Reactants}
\end{minipage}
\vspace{3pt}
\begin{minipage}{0.32\linewidth}
 \centering
 \includegraphics[width=0.9\linewidth]{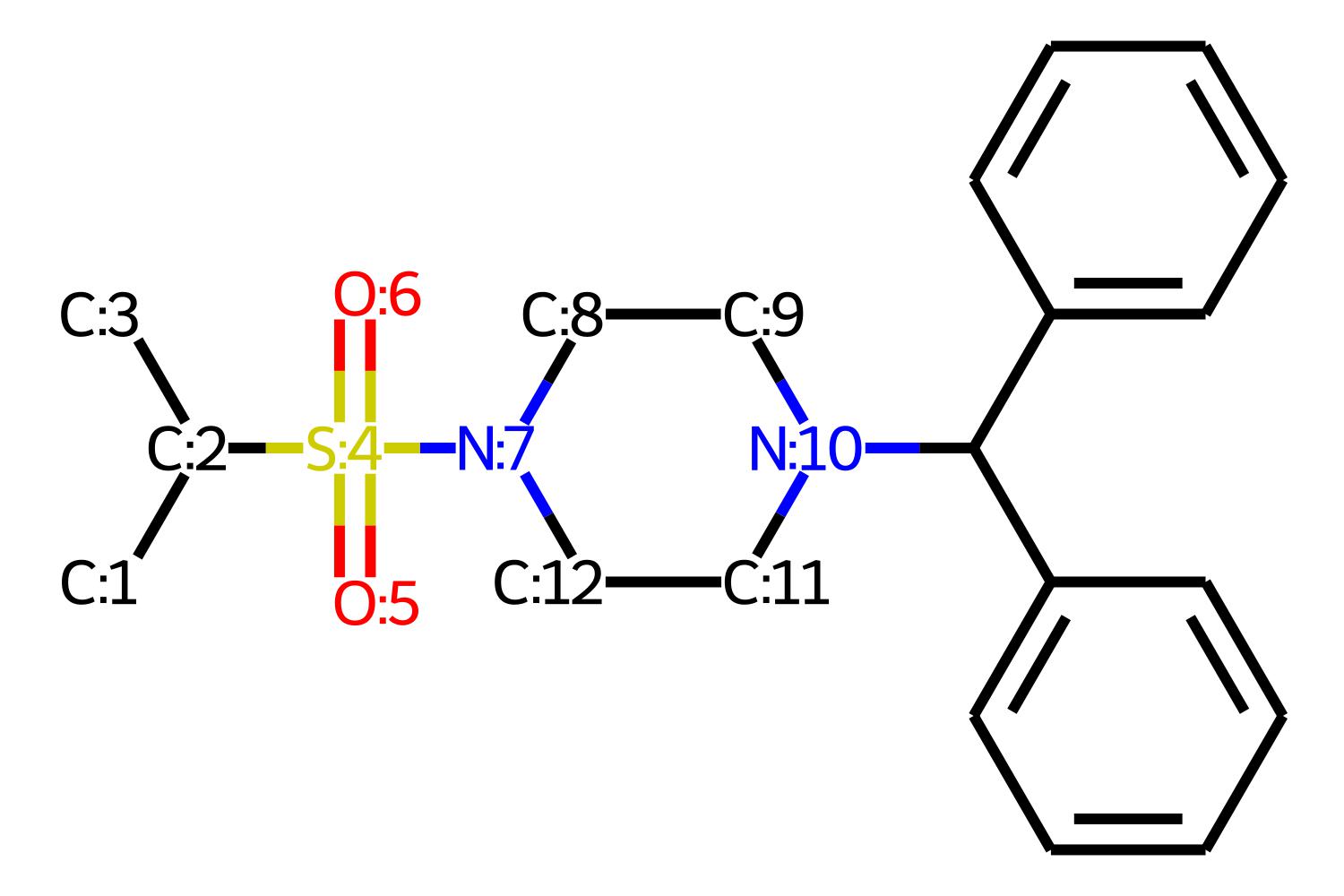}
 \centerline{Top 5 Prediction Reactants} 
\end{minipage}
\vspace{3pt}
\begin{minipage}{0.32\linewidth}
 \centering
 \includegraphics[width=0.9\linewidth]{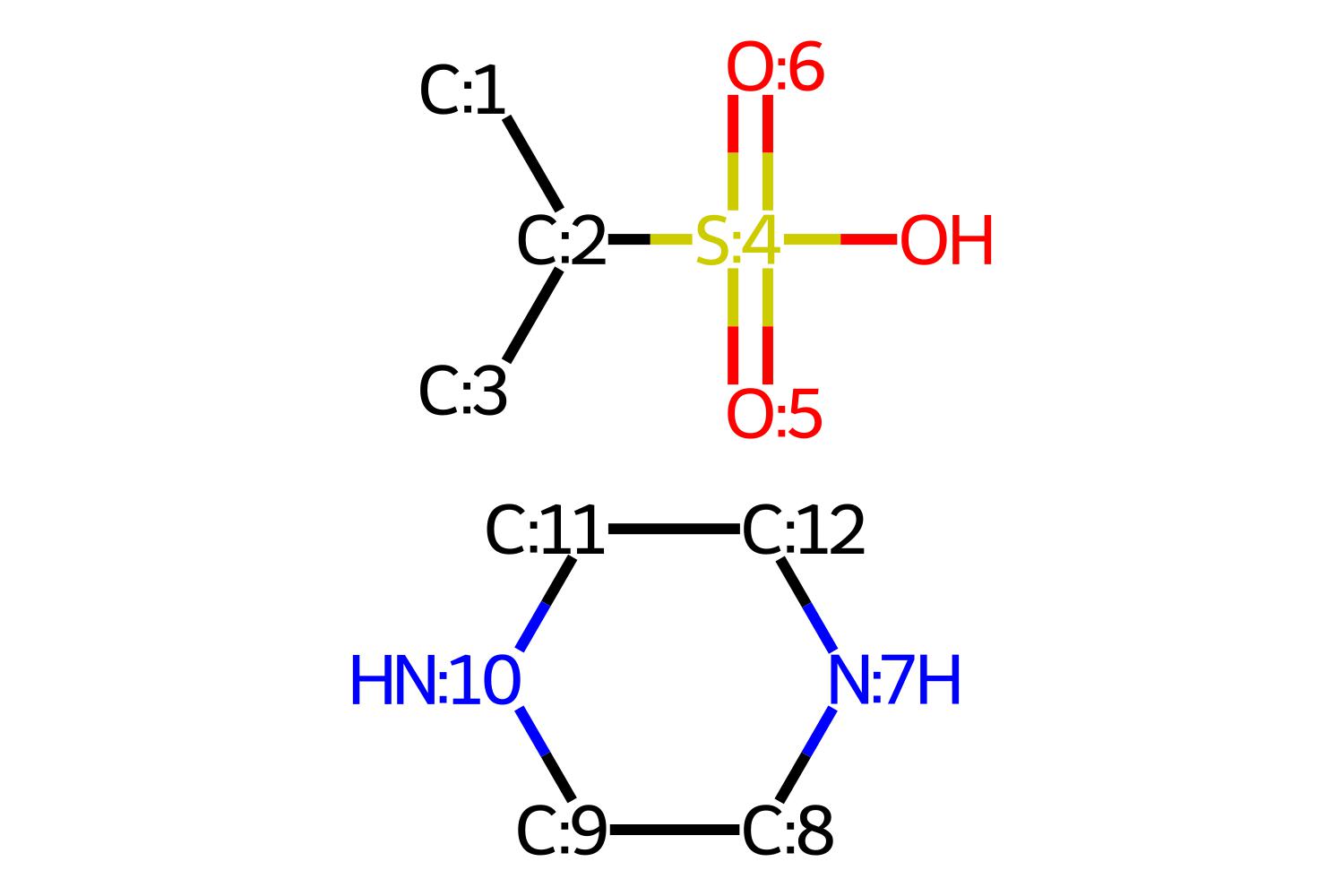}
 \centerline{Top 6 Prediction Reactants} 
\end{minipage}
\vspace{3pt}
\begin{minipage}{0.32\linewidth}
 \centering
 \includegraphics[width=0.9\linewidth]{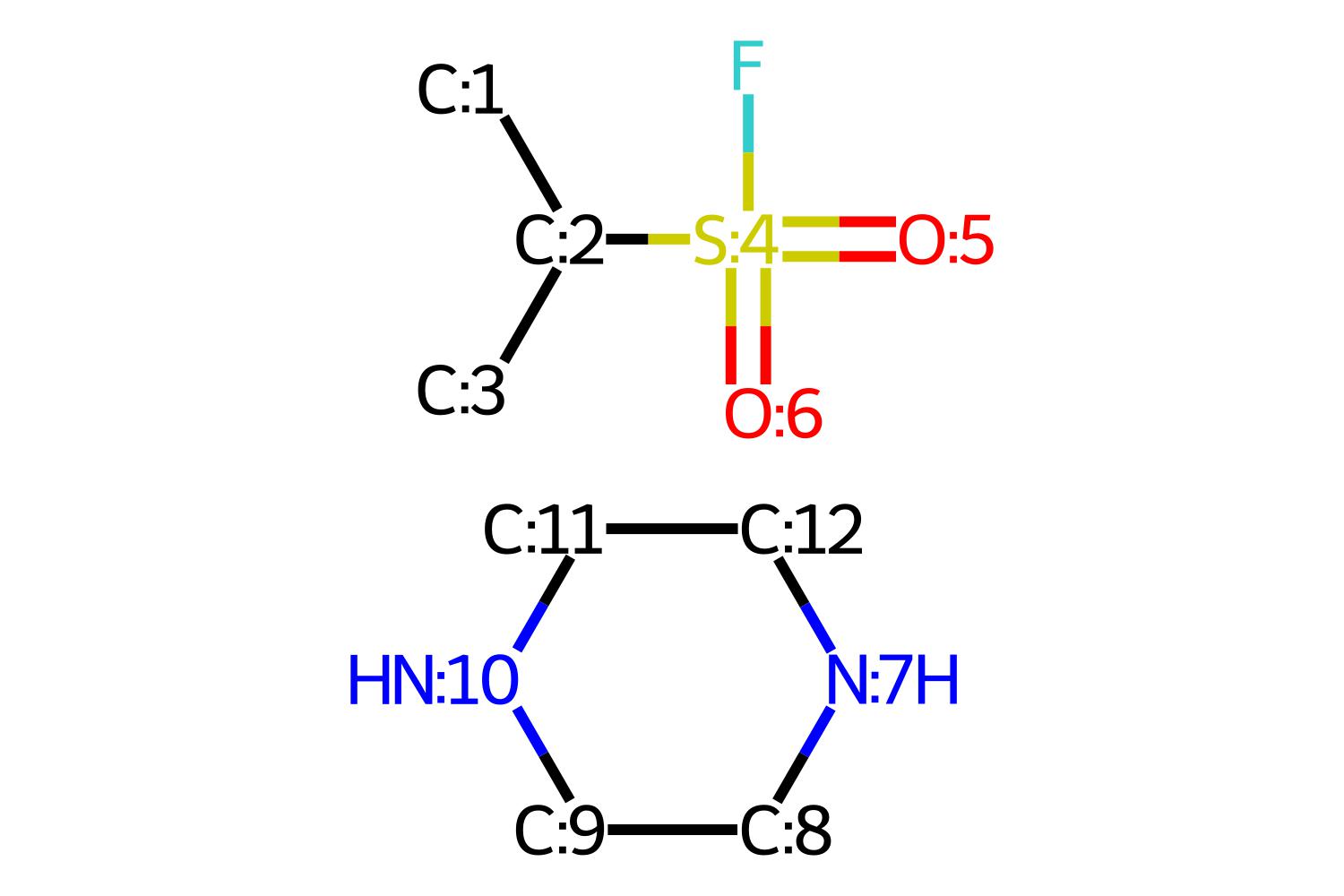}
 \centerline{Top 7 Prediction Reactants} 
\end{minipage}
\vspace{3pt}
\begin{minipage}{0.32\linewidth}
 \centering
 \includegraphics[width=1\linewidth]{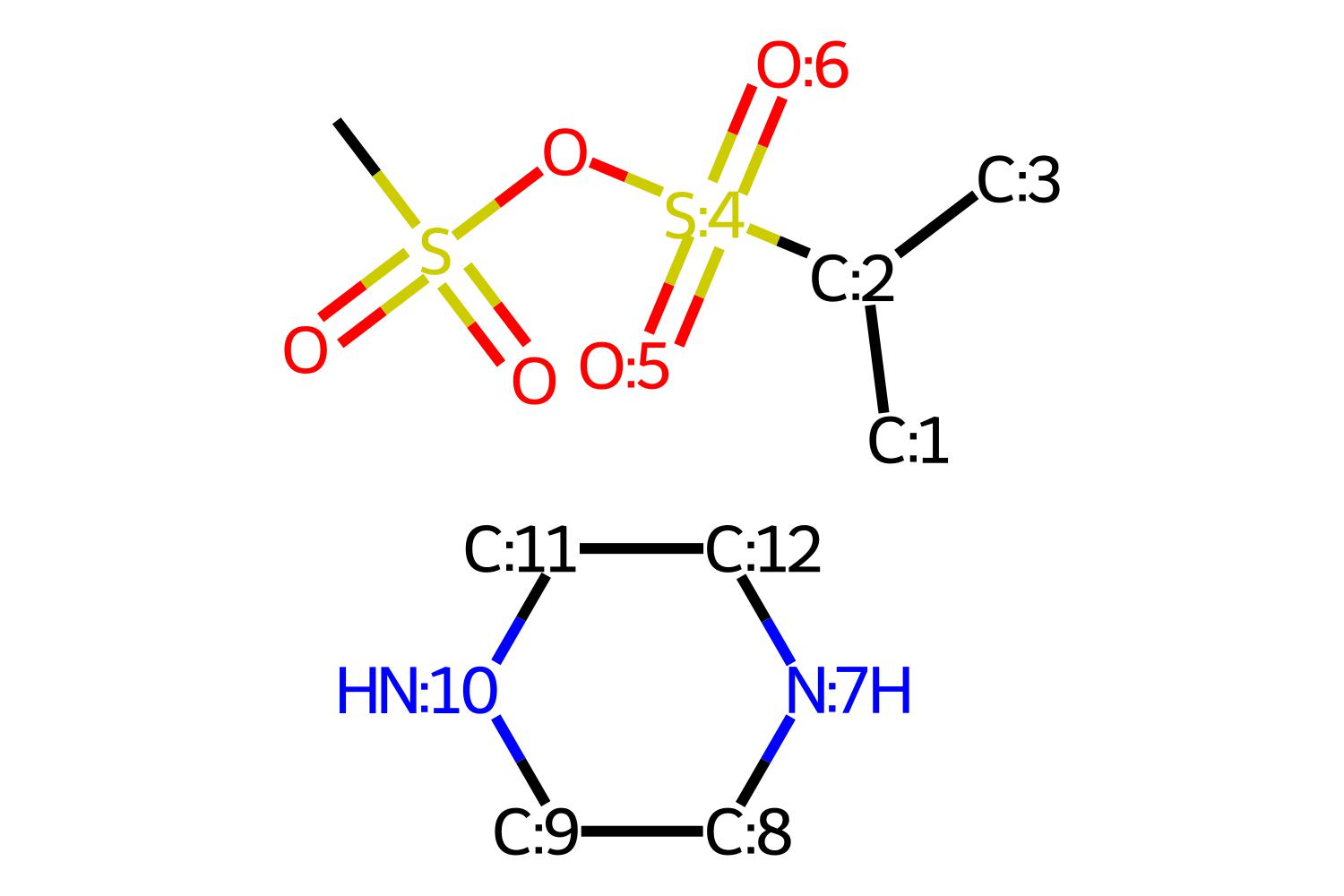}
 \centerline{Top 8 Prediction Reactants} 
\end{minipage}
\vspace{3pt}
\begin{minipage}{0.32\linewidth}
 \centering
 \includegraphics[width=1\linewidth]{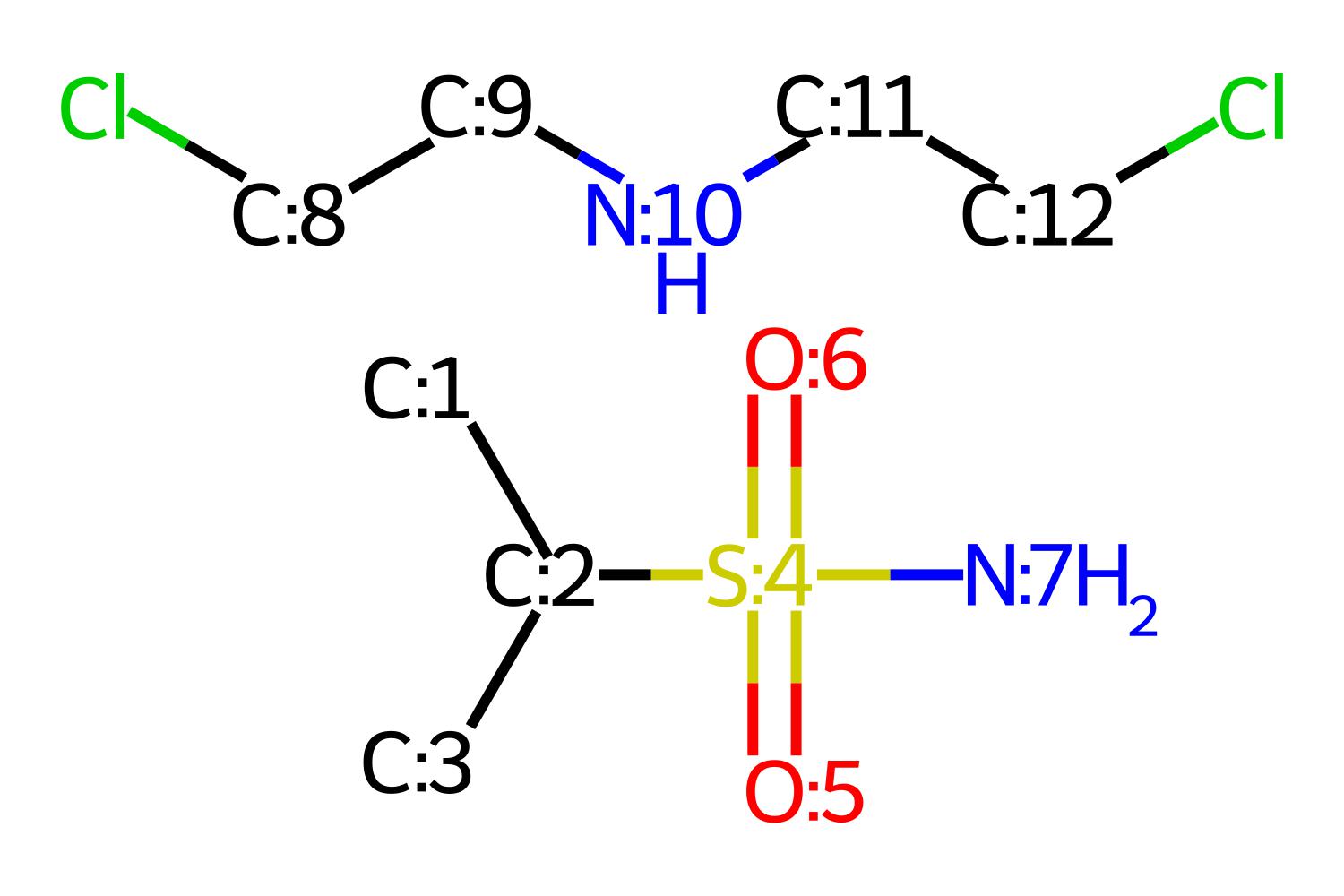}
 \centerline{Top 9 Prediction Reactants} 
\end{minipage}
\vspace{3pt}
\begin{minipage}{0.32\linewidth}
 \centering
 \includegraphics[width=1\linewidth]{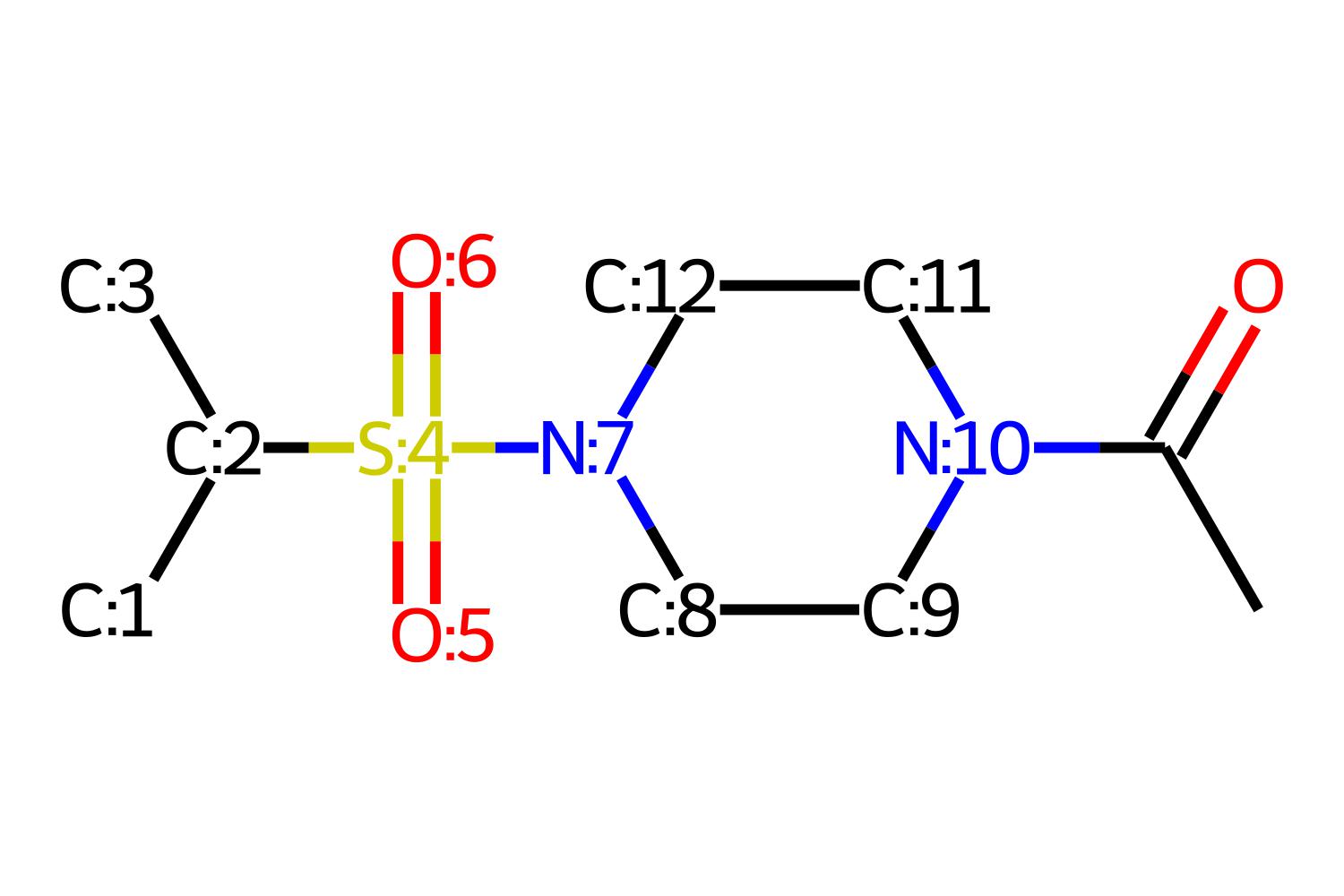}
 \centerline{Top 10 Prediction Reactants} 
\end{minipage}
\caption{This is example 2 of \name{} prediction on the USPTO-50k test dataset with the reaction class unknown. All the reactions are chemically valid in terms of mechanisms except for the sixth reaction. The first predicted reactants exact match the ground truth. The first, seventh, eighth, and ninth reactions fall under the category of heteroatom alkylation and arylation reaction, while the second, third, fourth, fifth and tenth are classified as deprotections reactions. It is worth noting that the sixth predicted reactants can be easily transformed to the ground truth by adding one extra step.}
\label{fig:result_two}
\end{figure}

\begin{figure}[htbp]
\centering
\begin{minipage}{0.32\linewidth}
 \centering
 \includegraphics[width=0.9\linewidth]{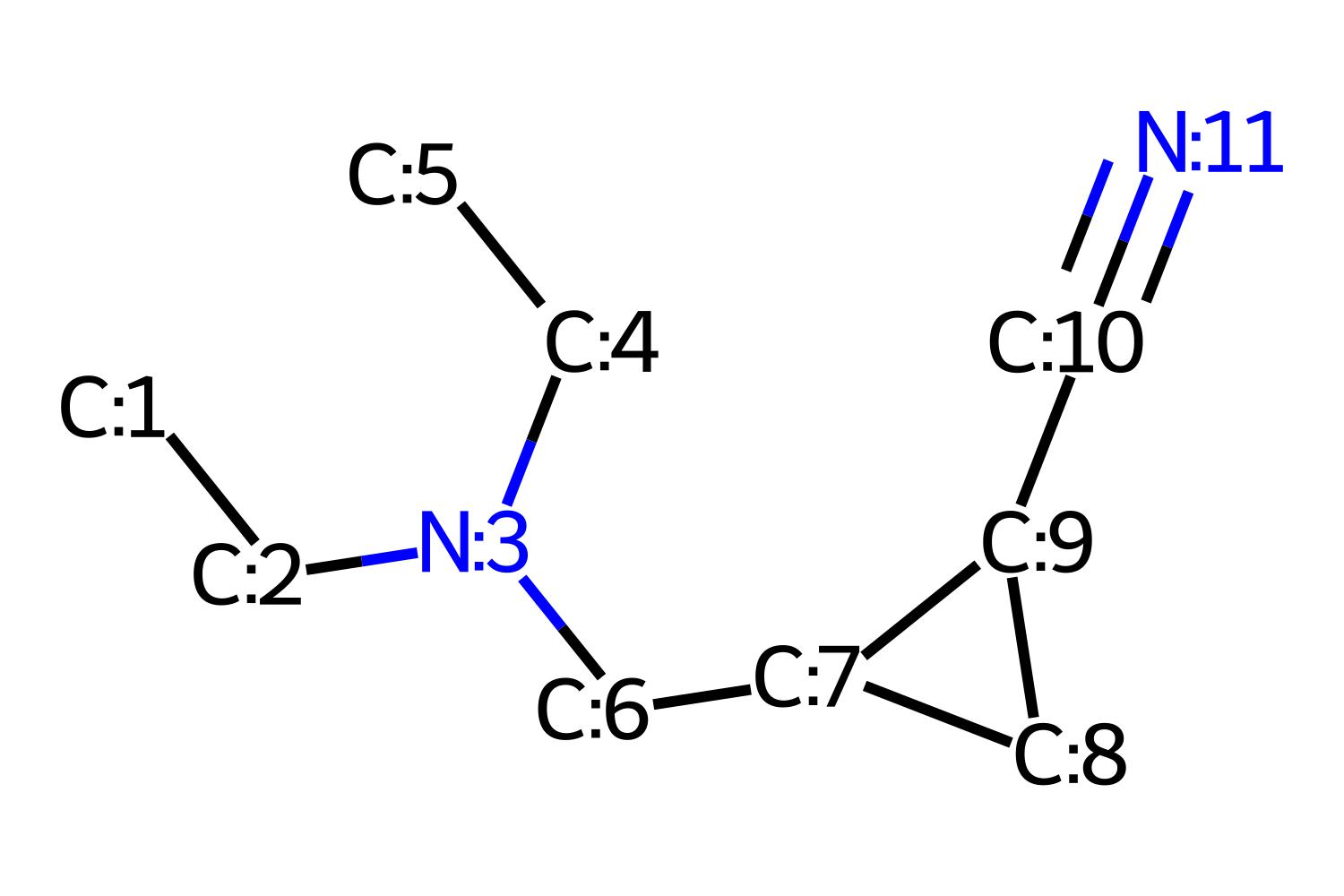}
 \centerline{Production}
\end{minipage}
\vspace{3pt}
\begin{minipage}{0.32\linewidth}
 \centering
 \includegraphics[width=0.9\linewidth]{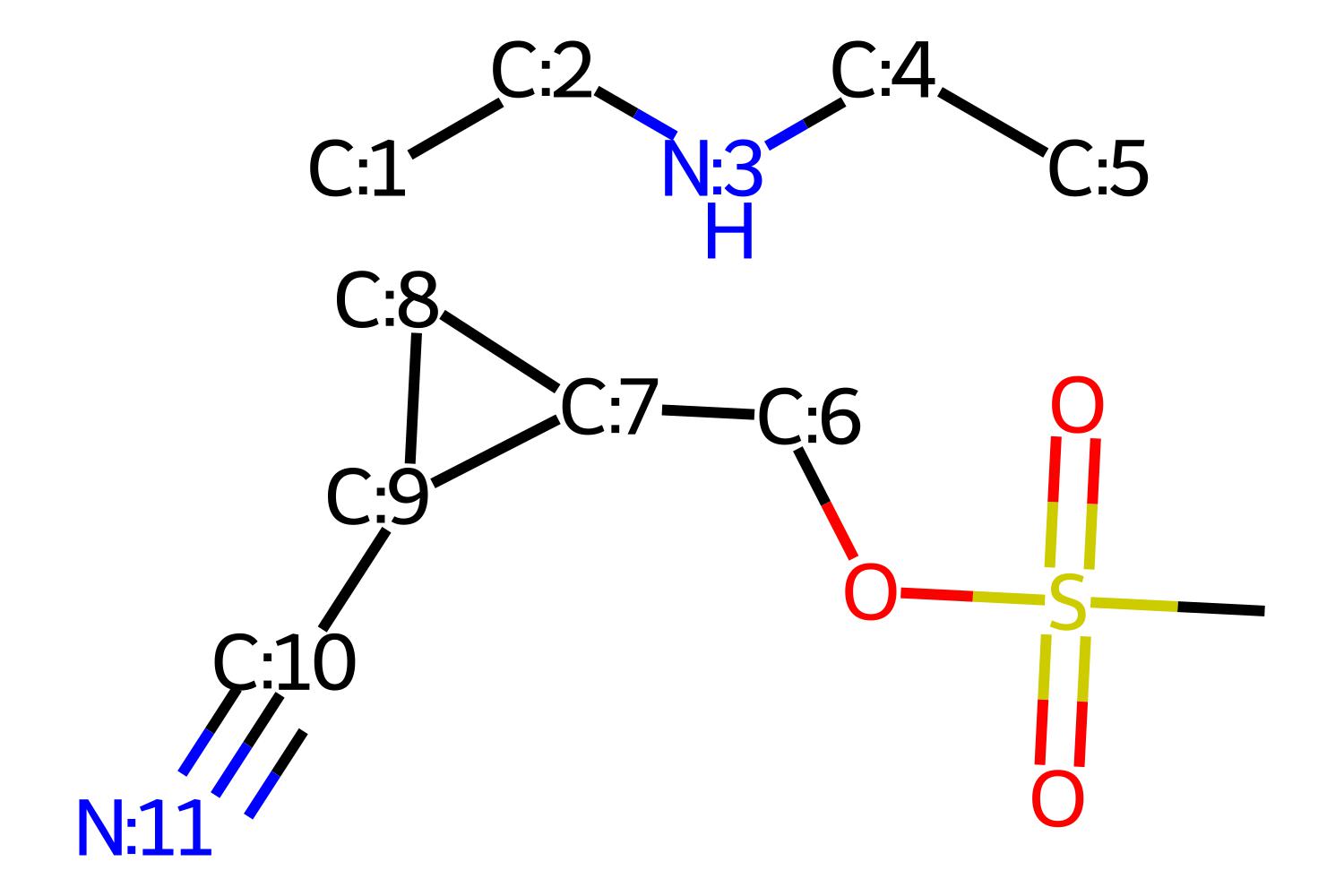}
 \centerline{Ground Truth Reactants}
\end{minipage}
\vspace{3pt}
\begin{minipage}{0.32\linewidth}
 \centering
 \includegraphics[width=0.9\linewidth]{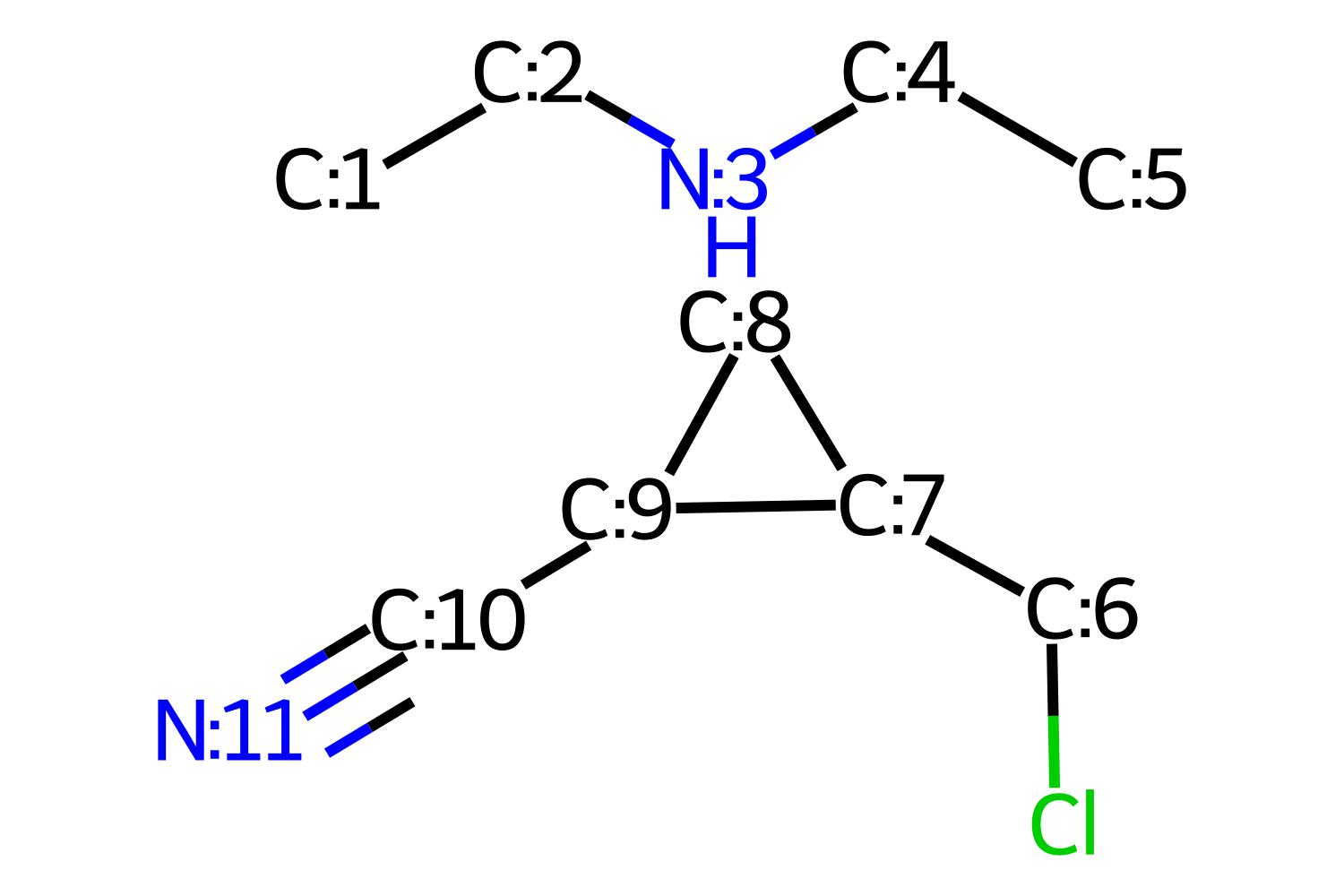}
 \centerline{Top 1 Prediction Reactants}
\end{minipage}
\vspace{3pt}
\begin{minipage}{0.32\linewidth}
 \centering
 \includegraphics[width=0.9\linewidth]{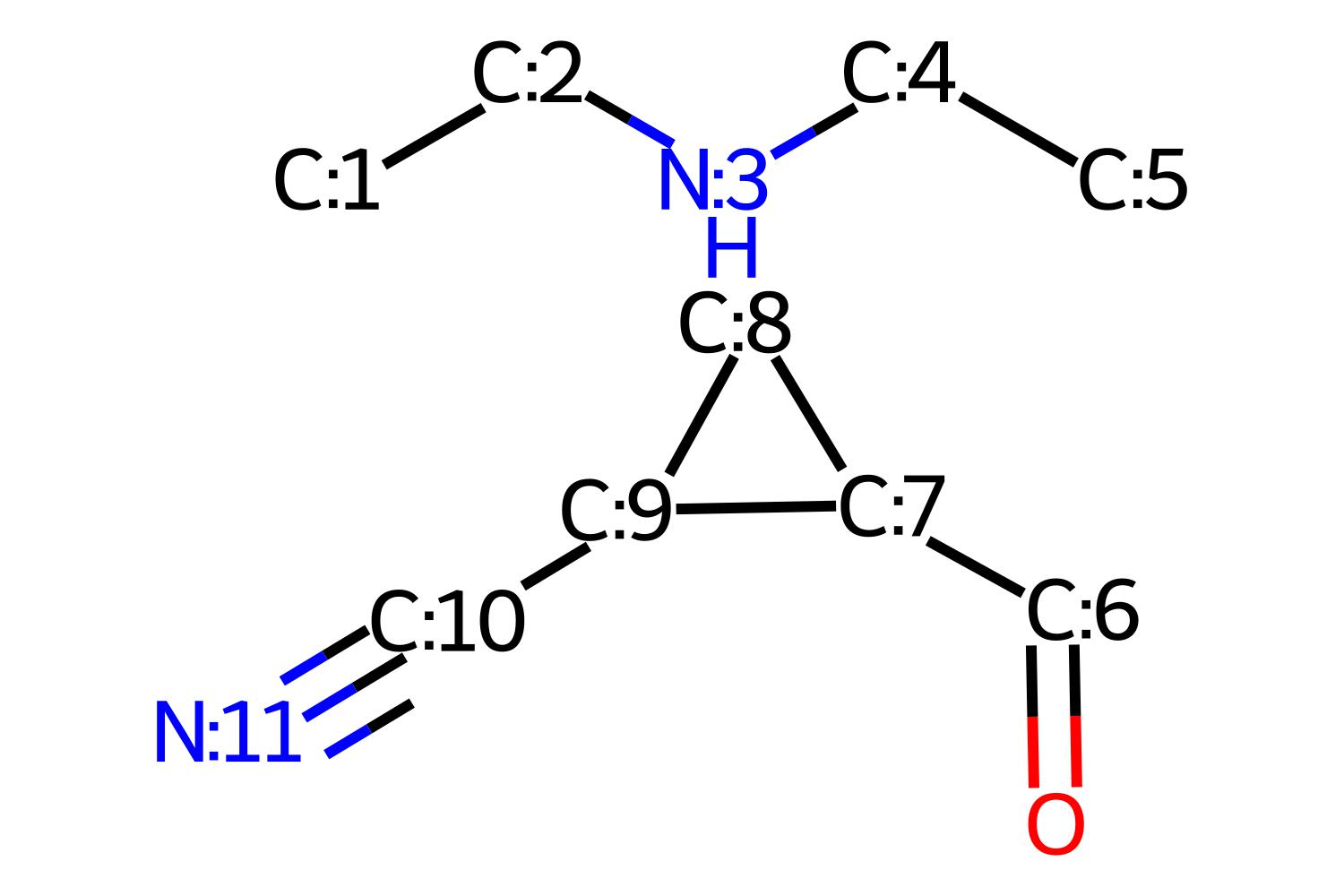}
 \centerline{Top 2 Prediction Reactants}
\end{minipage}
\vspace{3pt}
\begin{minipage}{0.32\linewidth}
 \centering
 \includegraphics[width=0.9\linewidth]{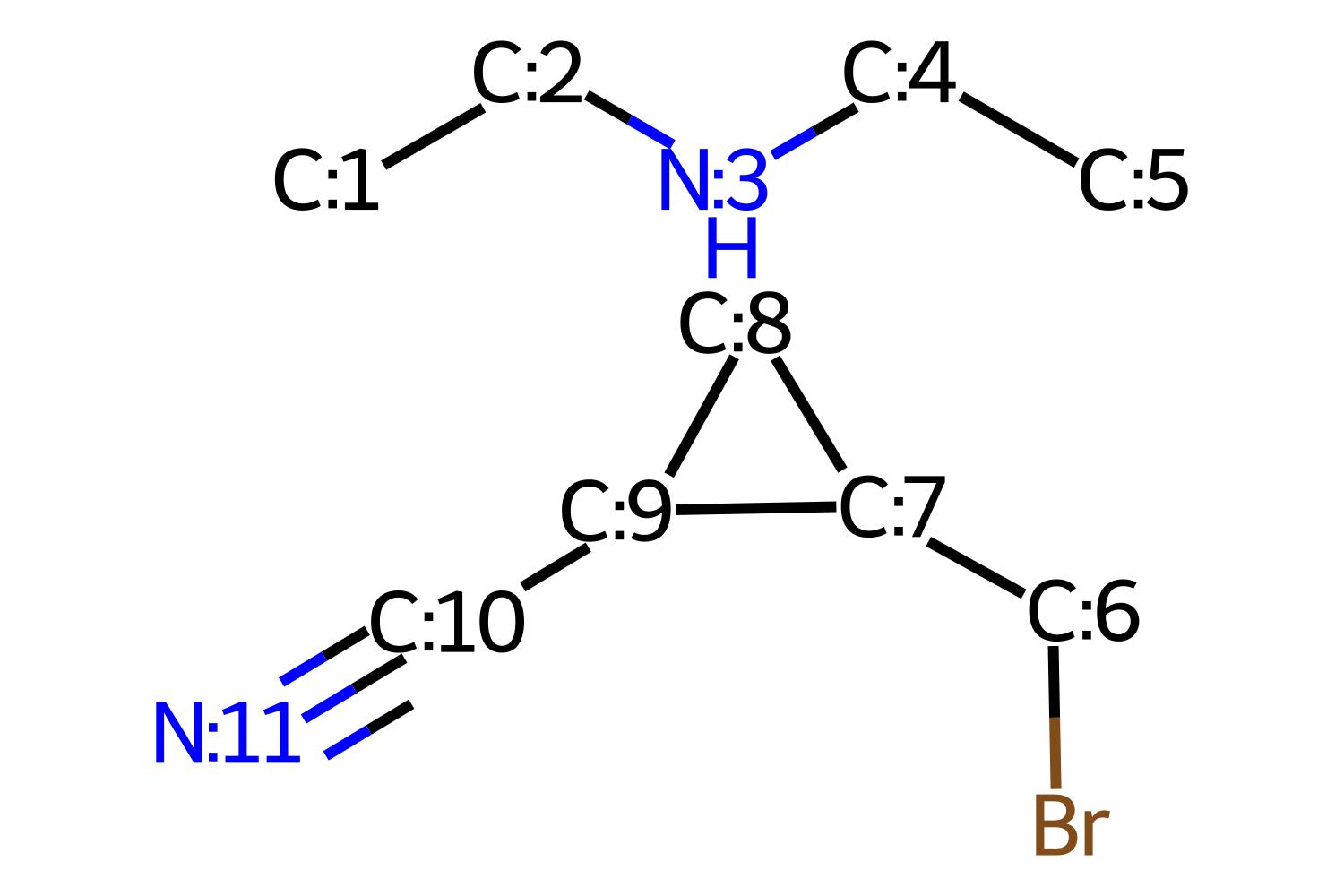}
 \centerline{Top 3 Prediction Reactants}
\end{minipage}
\vspace{3pt}
\begin{minipage}{0.32\linewidth}
 \centering
 \includegraphics[width=0.9\linewidth]{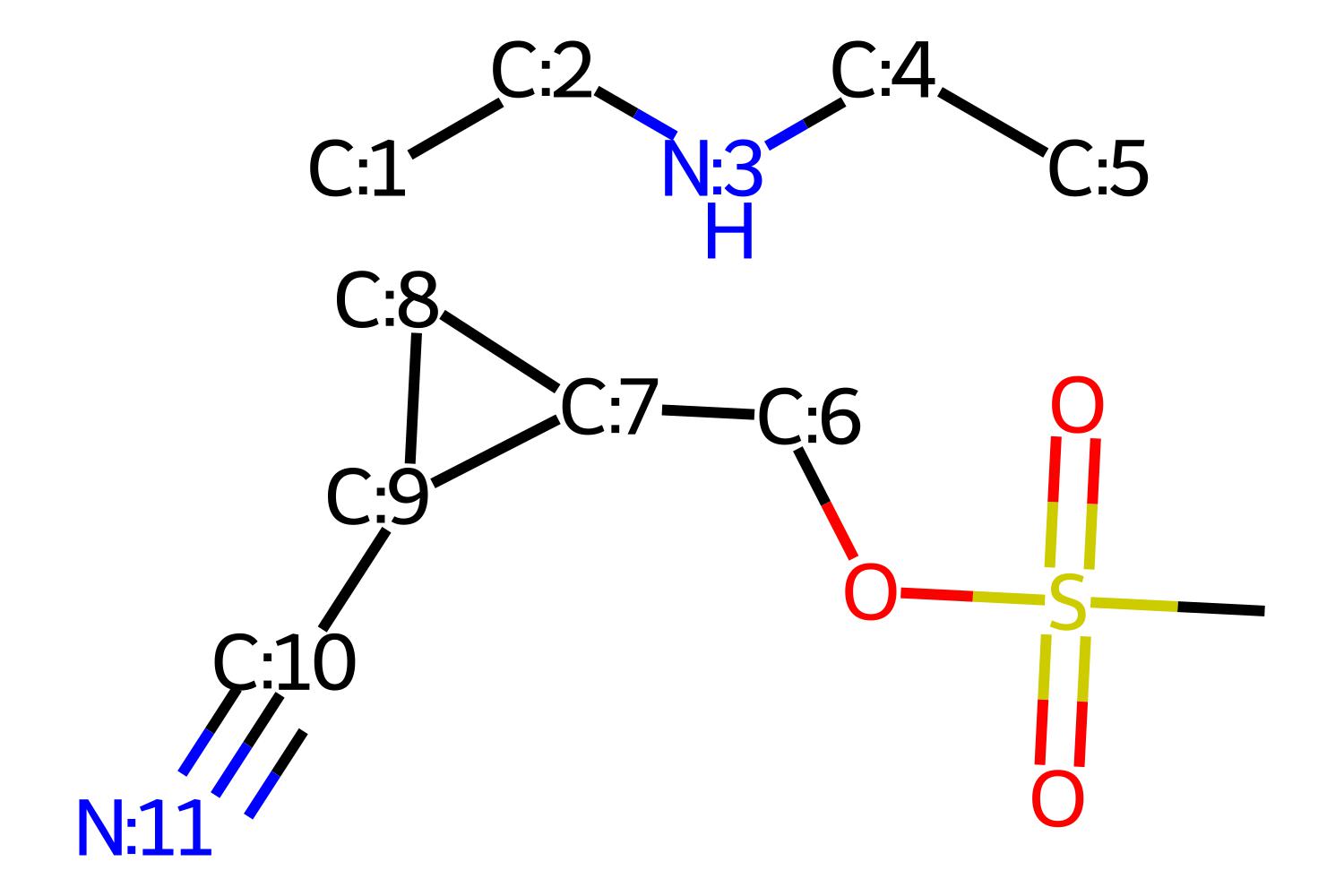}
 \centerline{Top 4 Prediction Reactants}
\end{minipage}
\vspace{3pt}
\begin{minipage}{0.32\linewidth}
 \centering
 \includegraphics[width=0.9\linewidth]{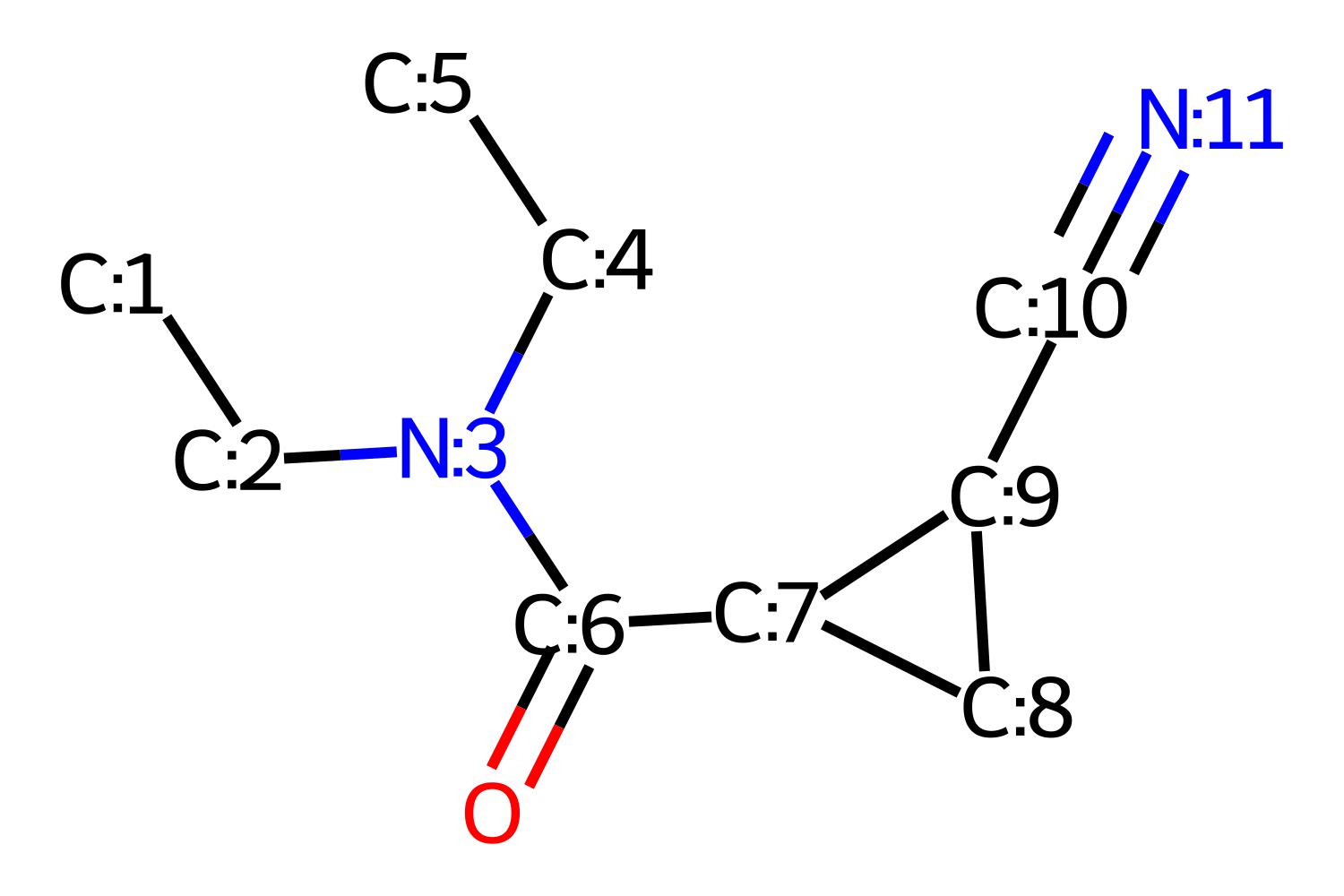}
 \centerline{Top 5 Prediction Reactants} 
\end{minipage}
\vspace{3pt}
\begin{minipage}{0.32\linewidth}
 \centering
 \includegraphics[width=0.9\linewidth]{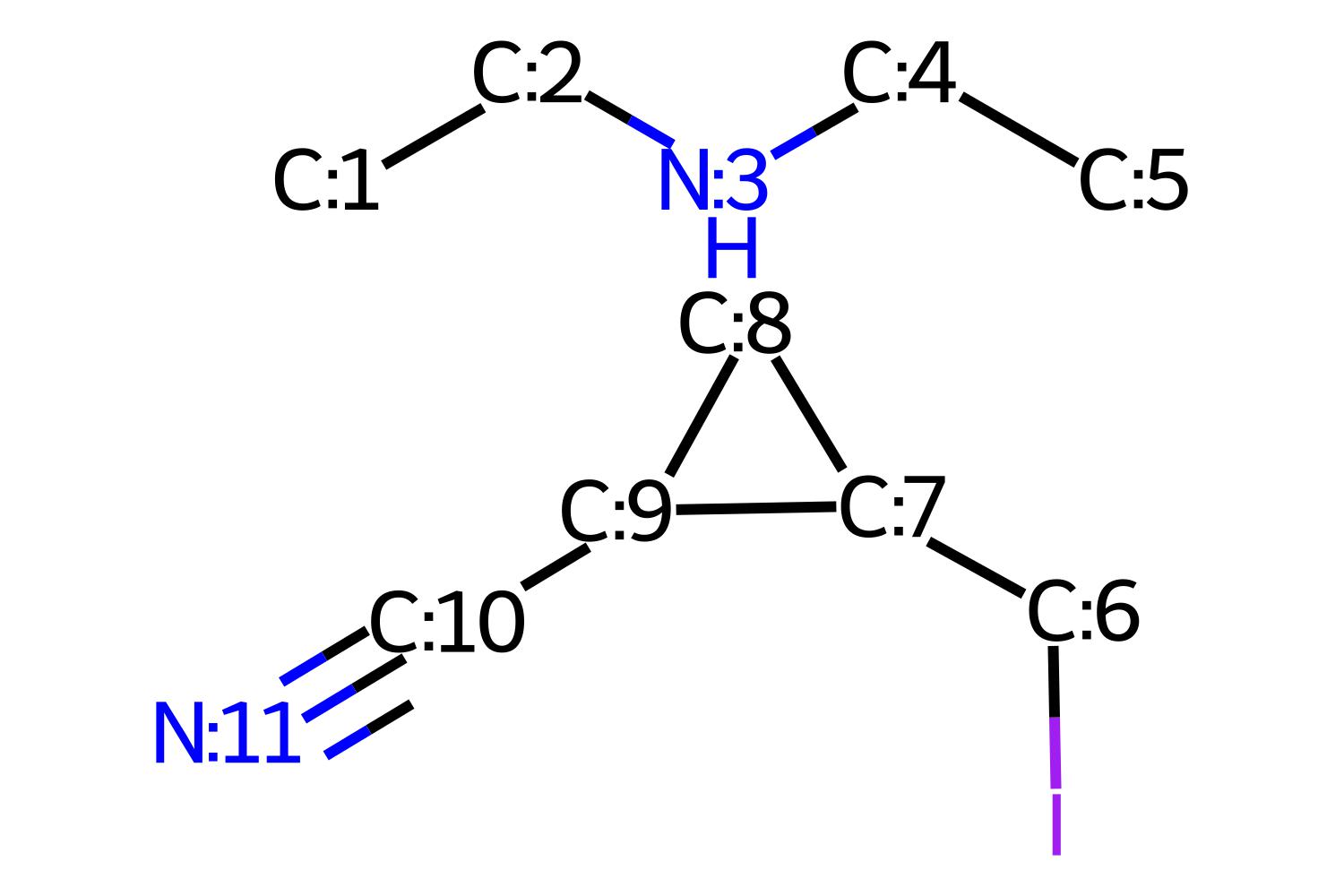}
 \centerline{Top 6 Prediction Reactants} 
\end{minipage}
\vspace{3pt}
\begin{minipage}{0.32\linewidth}
 \centering
 \includegraphics[width=0.9\linewidth]{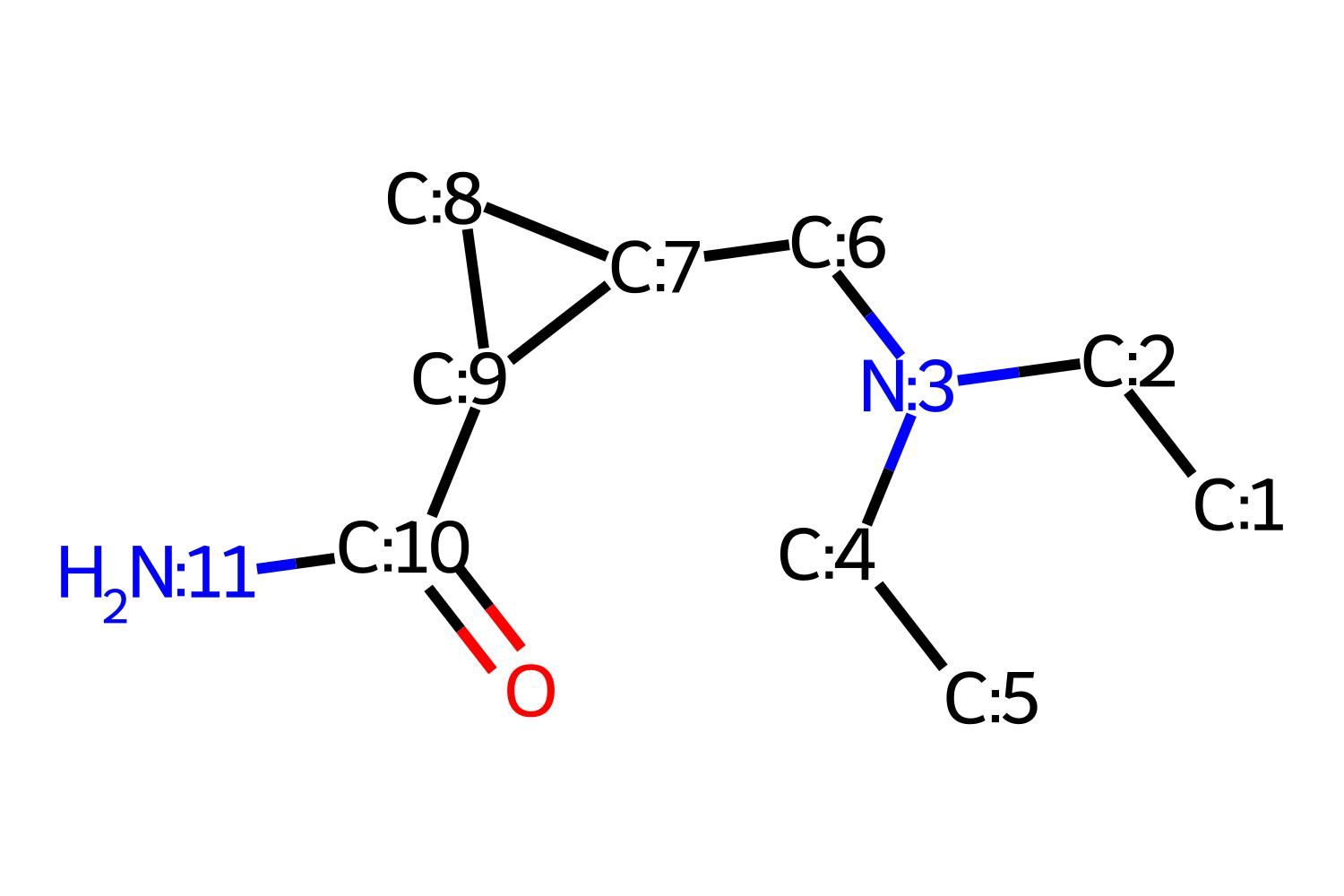}
 \centerline{Top 7 Prediction Reactants} 
\end{minipage}
\vspace{3pt}
\begin{minipage}{0.32\linewidth}
 \centering
 \includegraphics[width=0.9\linewidth]{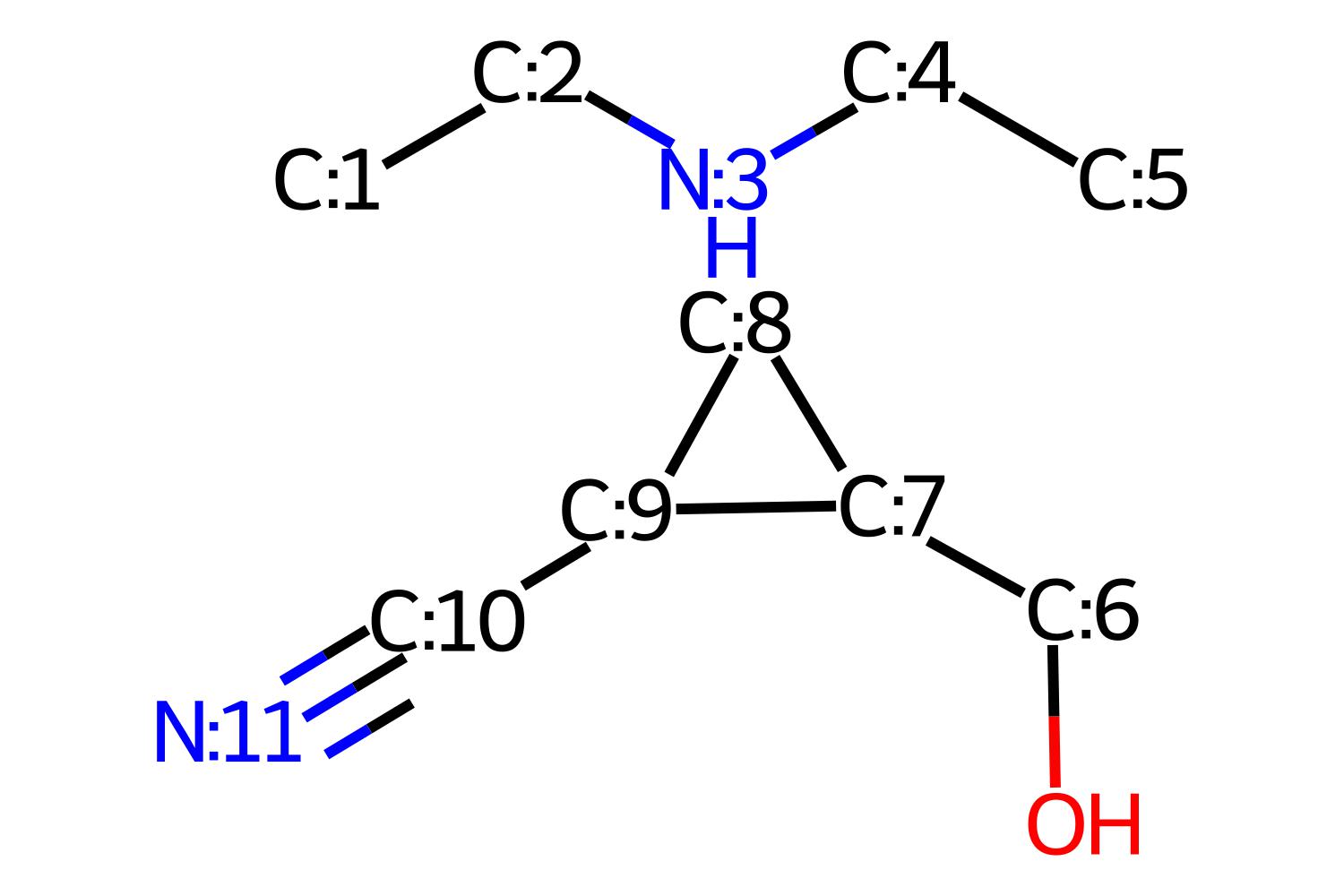}
 \centerline{Top 8 Prediction Reactants} 
\end{minipage}
\vspace{3pt}
\begin{minipage}{0.32\linewidth}
 \centering
 \includegraphics[width=0.9\linewidth]{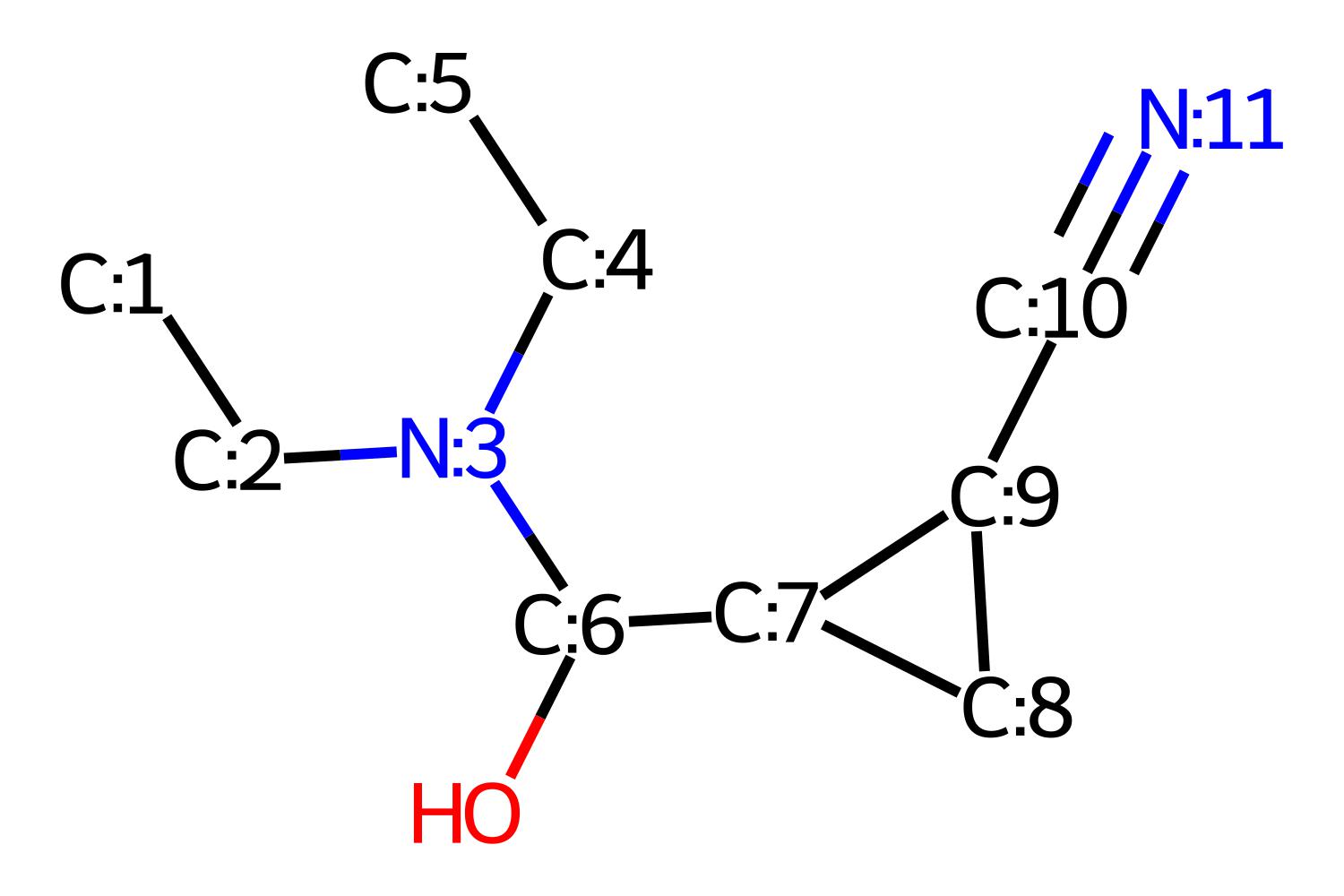 }
 \centerline{Top 9 Prediction Reactants} 
\end{minipage}
\vspace{3pt}
\begin{minipage}{0.32\linewidth}
 \centering
 \includegraphics[width=0.9\linewidth]{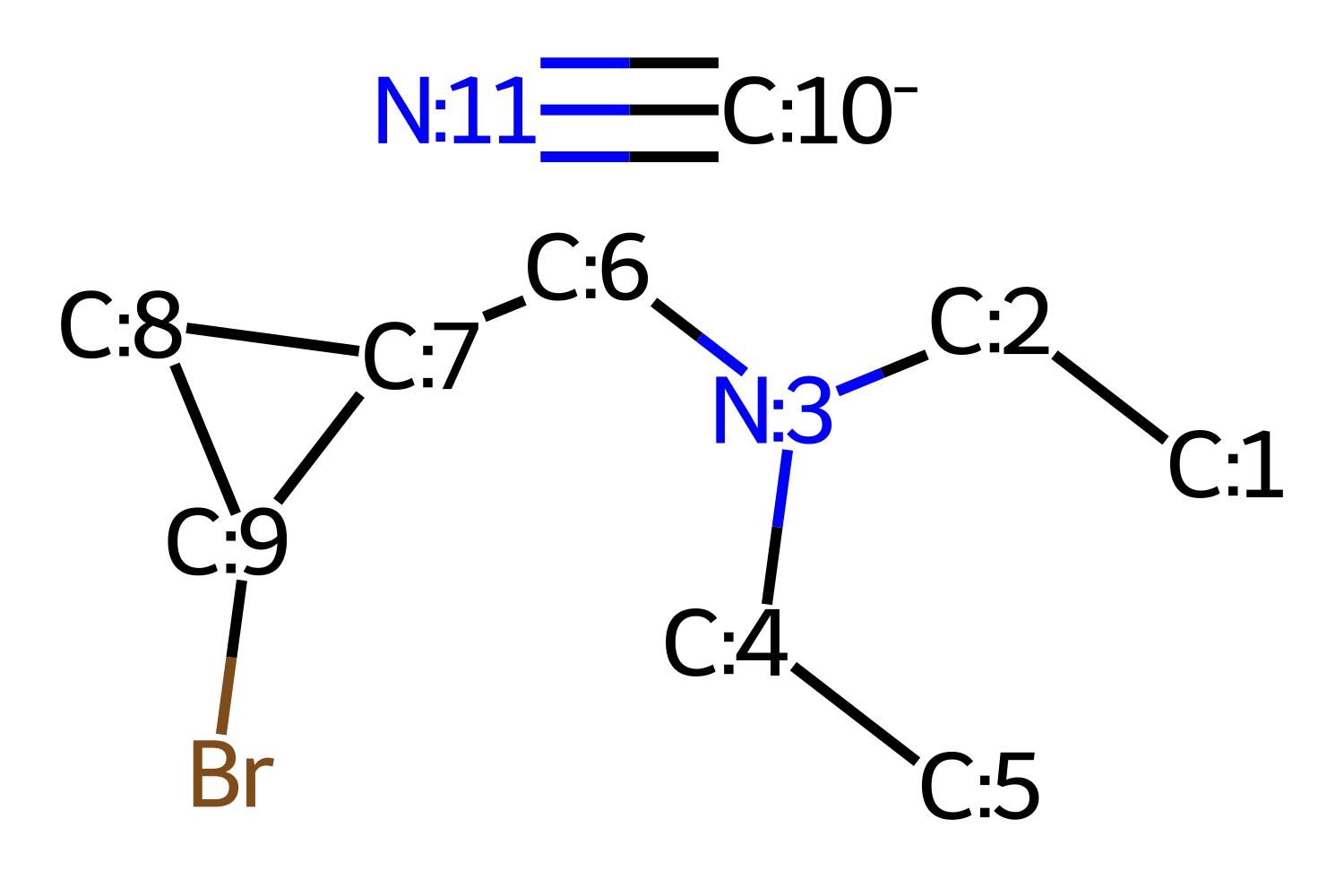}
 \centerline{Top 10 Prediction Reactants} 
\end{minipage}
\caption{This is example 3 of \name{} prediction on the USPTO-50k test dataset with the reaction class unknown. Aside from the eighth reaction, all the other reactions are chemically valid in terms of mechanisms. The fourth predicted reactants exactly match the ground truth. The first, third, fourth, and sixth reactions can be categorized as heteroatom alkylation and arylation reactions, while the second, fifth, and ninth reactions are classified as reduction reactions. The seventh reaction uniquely belongs to the functional group interconversion reaction category, and the tenth reaction is a C-C bond formation reaction. It is worth noting that the eighth prediction can be easily converted to the ground truth by adding just one extra step.}
\label{fig:result_three}
\end{figure}

\onecolumn

\end{suppinfo}

\bibliography{achemso-demo}

\end{document}